\documentclass[manuscript,screen,review=False]{acmart}
\DeclareUnicodeCharacter{2061}{}
\AtBeginDocument{%
  }

\setcopyright{acmlicensed}
\copyrightyear{2024}
\acmYear{2024}
\acmDOI{XXXXXXX.XXXXXXX}

\acmISBN{978-1-4503-XXXX-X/18/06}

\usepackage{enumerate}
\usepackage{algorithm}
\usepackage{multirow}
\usepackage{makecell}

\begin{document}

\title{Emotion-aware Personalized Music Recommendation with a Heterogeneity-aware Deep Bayesian Network}

\author{Erkang Jing}
\email{jingerkang@mail.hfut.edu.cn}
\orcid{0000-0003-3763-6937}
\affiliation{%
  \institution{School of Management, Hefei University of Technology}
  \city{Hefei}
  \state{Anhui}
  \country{China}
}

\author{Yezheng Liu}
\affiliation{%
  \institution{School of Management, Hefei University of Technology and Key Laboratory of Process Optimization and Intelligent Decision-Making, Ministry of Education}
  \city{Hefei}
  \state{Anhui}
  \country{China}}
\email{liuyezheng@hfut.edu.cn}
\orcid{0000-0002-9193-5236}

\author{Yidong Chai}
\authornote{Corresponding author}
\affiliation{%
  \institution{School of Management, Hefei University of Technology and Key Laboratory of Philosophy and Social Sciences for Cyberspace Behaviour and Management}
  \city{Hefei}
  \state{Anhui}
  \country{China}}
\email{chaiyd@hfut.edu.cn}
\orcid{0000-0003-0260-7589}

\author{Shuo Yu}
\affiliation{%
 \institution{Rawls College of Business, Texas Tech University}
 \city{Lubbock}
 \state{Texas}
 \country{U.S.}}
\email{Shuo.Yu@ttu.edu}
\orcid{0000-0002-1885-2813}

\author{Longshun Liu}
\affiliation{%
  \institution{School of Management, Hefei University of Technology}
  \city{Hefei}
  \state{Anhui}
  \country{China}}
\email{longshun.liu@mail.hfut.edu.cn}
\orcid{}

\author{Yuanchun Jiang}
\affiliation{%
  \institution{School of Management, Hefei University of Technology and Key Laboratory of Philosophy and Social Sciences for Cyberspace Behaviour and Management}
  \city{Hefei}
  \state{Anhui}
  \country{China}}
\email{ycjiang@hfut.edu.cn}
\orcid{0000-0003-0886-3647}

\author{Yang Wang}
\affiliation{%
  \institution{School of Computer Science and Information Engineering, Hefei University of Technology}
  \city{Hefei}
  \state{Anhui}
  \country{China}}
\email{yangwang@hfut.edu.cn}
\orcid{0000-0003-1029-9280}

\renewcommand{\shortauthors}{E. Jing et al.}


\begin{abstract}
  Music recommender systems play a critical role in music streaming platforms by providing users with music that they are likely to enjoy. Recent studies have shown that user emotions can influence users’ preferences for music moods. However, existing emotion-aware music recommender systems (EMRSs) explicitly or implicitly assume that users’ actual emotional states expressed through identical emotional words are homogeneous. They also assume that users’ music mood preferences are homogeneous under the same emotional state. In this article, we propose four types of heterogeneity that an EMRS should account for: emotion heterogeneity across users, emotion heterogeneity within a user, music mood preference heterogeneity across users, and music mood preference heterogeneity within a user. We further propose a Heterogeneity-aware Deep Bayesian Network (HDBN) to model these assumptions. The HDBN mimics a user’s decision process of choosing music with four components: personalized prior user emotion distribution modeling, posterior user emotion distribution modeling, user grouping, and Bayesian neural network-based music mood preference prediction. We constructed two datasets, called EmoMusicLJ and EmoMusicLJ-small, to validate our method. Extensive experiments demonstrate that our method significantly outperforms baseline approaches on metrics of HR, Precision, NDCG, and MRR. Ablation studies and case studies further validate the effectiveness of our HDBN. The source code and datasets are available at \url{https://github.com/jingrk/HDBN}.
\end{abstract}




\maketitle

\section{Introduction}
Music plays a vital role in people’s lives, offering enjoyment, emotional fulfillment, self-expression, and social bonding \cite{Liu2023, Fabian2017, Shen2020, Cheng2016}. With advances in information technology, music became one of the first products to be digitalized and sold online. Today, subscription and streaming services are the primary business models for music platforms such as Apple Music, Spotify, and NetEase Cloud. According to the 2024 International Federation of Phonographic Industry Global Music Report \footnote{https://globalmusicreport.ifpi.org}, global revenue from music streaming reached $\$$19.3 billion in 2023, a 10.4$\%$ increase from the previous year. These platforms aim to boost user engagement by introducing new music that users are likely to enjoy. However, with millions of tracks available, users often face information overload when exploring music libraries \cite{Razgallah2024, Wang2023}. To address this, music recommender systems (MRSs) have been widely used to facilitate the matching between users and music pieces \cite{Schedl2015}. MRSs help reduce users’ information overload during music exploration \cite{Peter2019, Christine2017} and assist platforms in improving user loyalty and retention, thereby strengthening their competitiveness in the music streaming market \cite{Schafer1999}.

\begin{figure}[h]
  \centering
  \includegraphics[width=\linewidth]{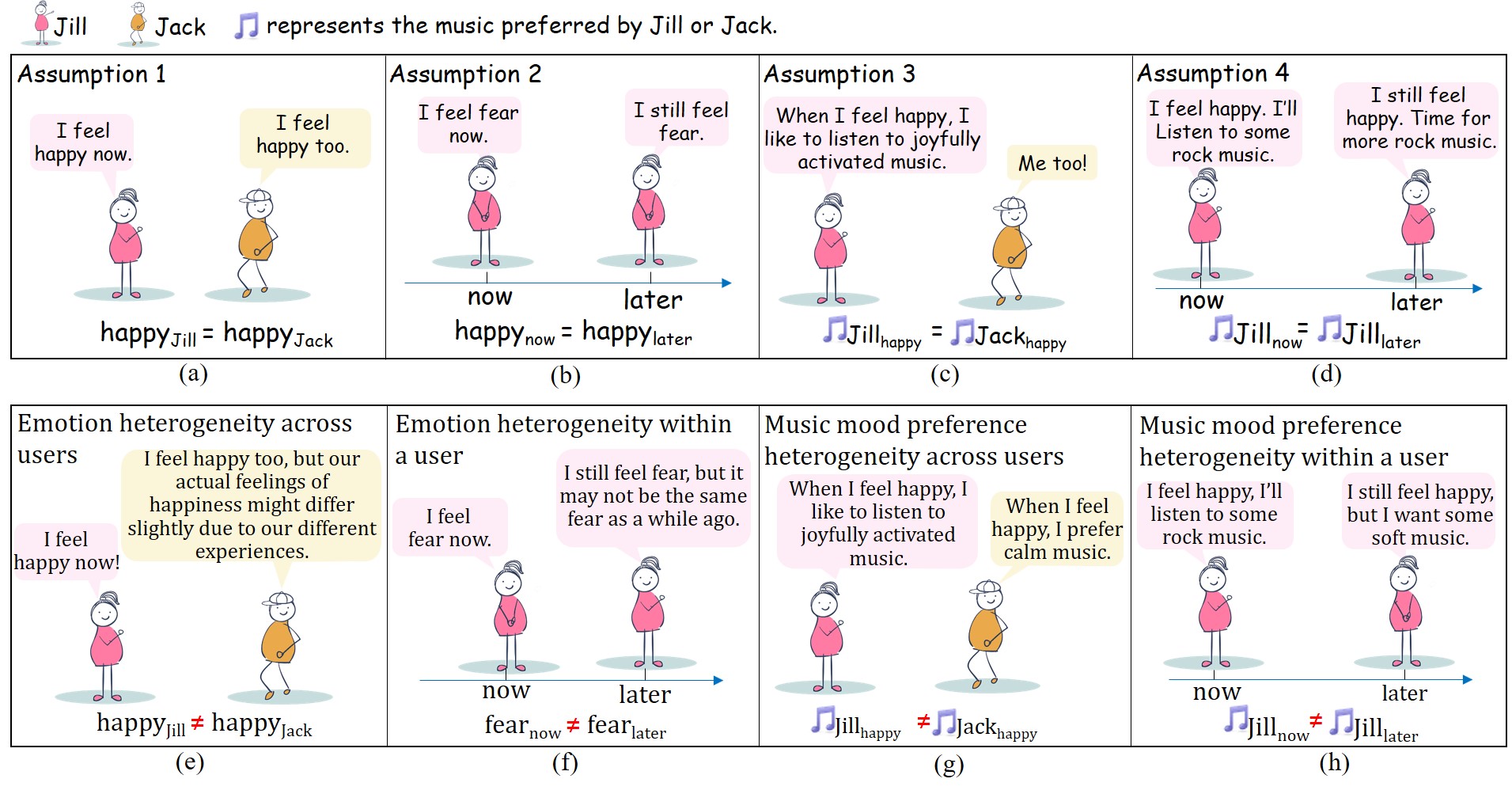}
  \caption{Toy examples of the existing EMRSs assumptions (a, b, c, and d) and the four heterogeneity assumptions proposed in our method (e, f, g, and h).}
  \Description{}
  \label{assumptions}
\end{figure}

Early MRSs followed the design of general recommender systems, focusing on collaborative filtering (user-music interactions, such as listening and rating behaviors), music features (e.g., rhythm), and contextual information (e.g., weather and location) \cite{Song2012, Peter2013, Wu2022, wang2021a, wang2021b}. Later, researchers recognized that user emotions may largely influence music preferences \cite{Liang2021, Xue2018, Christa2015}. For example, when a person is happy, they may tend to listen to music with positive moods, such as joyful activation \cite{Deng2015, Patrik2011}. As a result, emotion-aware music recommender systems (EMRSs) have been proposed to integrate emotional information into music recommendations \cite{Liang2021, Guo2017}. Emotional information comprises two parts: \textit{user emotion}, such as happy, sad, or angry feelings of users, and \textit{music mood}, such as joyful activation, nostalgia, or tension contained in the music. EMRSs enhance users’ experience and their willingness to engage with the music streaming platform \cite{Ivana2019, Nicolas2008}. Existing EMRSs either incorporated emotional information as additional features into the recommendation process \cite{Shen2020, Guo2017, Cheng2016b} or focused on the matching between user emotions and music moods (e.g., users who are happy or want to be happy are recommended music with joyful activation) \cite{Tran2023, Annam2024} under the following assumptions (explicit or implicit):
\begin{enumerate}
    \item [(1)] {\itshape Across different users}, the actual emotional states they describe using an emotion word (e.g., “happy”) are homogeneous \cite{Bontempelli2022}. For instance, in Figure \ref{assumptions}(a), if both Jack and Jill describe their current emotion as “happy,” their actual emotional states are considered identical.
    \item [(2)] {\itshape Within a user}, his/her actual emotional states described by an emotion word (e.g., “fear”) at different times are homogeneous \cite{Bontempelli2022}. For instance, in Figure \ref{assumptions}(b), if Jill describes her current emotion as “fear” at two different times, her actual emotional states are considered identical at both times.
    \item [(3)] {\itshape Across different users}, their preferences for music moods under a given emotional state are homogeneous \cite{Tran2023, Annam2024}. For instance, in Figure \ref{assumptions}(c), if Jack and Jill are both happy, their preferences for music moods are considered identical (e.g., they both prefer music with joyful activation in a happy state).
    \item [(4)] {\itshape Within a user}, his/her preference for music moods under a given emotional state are homogeneous \cite{Tran2023, Annam2024}. For instance, in Figure \ref{assumptions}(d), if Jill is happy at two different times, her preferences for music moods are identical at both times.
\end{enumerate}

However, psychological research and common sense suggest that the above assumptions may not always hold \cite{Liu2023, Xue2018, Barrett2006, Barrett2009, Barrett2022, Ronald2012, Liu2014, Patrik2008}.

\begin{enumerate}
    \item [(1)] Regarding Assumption 1, studies \cite{Barrett2006, Barrett2009} have shown that individuals’ perception, interpretation, and classification of emotions are influenced by the conceptual knowledge of emotions learned from everyday language, social activities, culture, and other factors. This results in differences in emotion cognition and expression across individuals. We refer to such phenomena as {\itshape emotion heterogeneity across users} (see Figure \ref{assumptions}(e)).
    \item [(2)] Regarding Assumption 2, we often use the same emotion word to describe different emotional states. For example, the fear of facing an exam differs from the fear of facing a bee sting \cite{Barrett2022}. We refer to such phenomena as {\itshape emotion heterogeneity within a user} (see Figure \ref{assumptions}(f)).
    \item [(3)] Regarding Assumption 3, many studies have indicated that not everyone tends to select music with moods that match their emotions and that different users may prefer different music moods even if they are in the same emotional state \cite{Xue2018, Ronald2012, Liu2014}. We refer to such phenomena as {\itshape music mood preference heterogeneity across users} (see Figure \ref{assumptions}(g)).
    \item [(4)] Regarding Assumption 4, a user’s music mood preference may change over time even if she/he reports no change in her/his emotional state \cite{Liu2023, Patrik2008}. We refer to such phenomena as {\itshape music mood preference heterogeneity within a user} (see Figure \ref{assumptions}(h)).
\end{enumerate}
To the best of our knowledge, no existing EMRSs have explicitly incorporated the four types of heterogeneity described above as underlying assumptions in their design, which may result in suboptimal performances in personalized music recommendation.

\begin{figure}[h]
  \centering
  \includegraphics[width=0.9\linewidth]{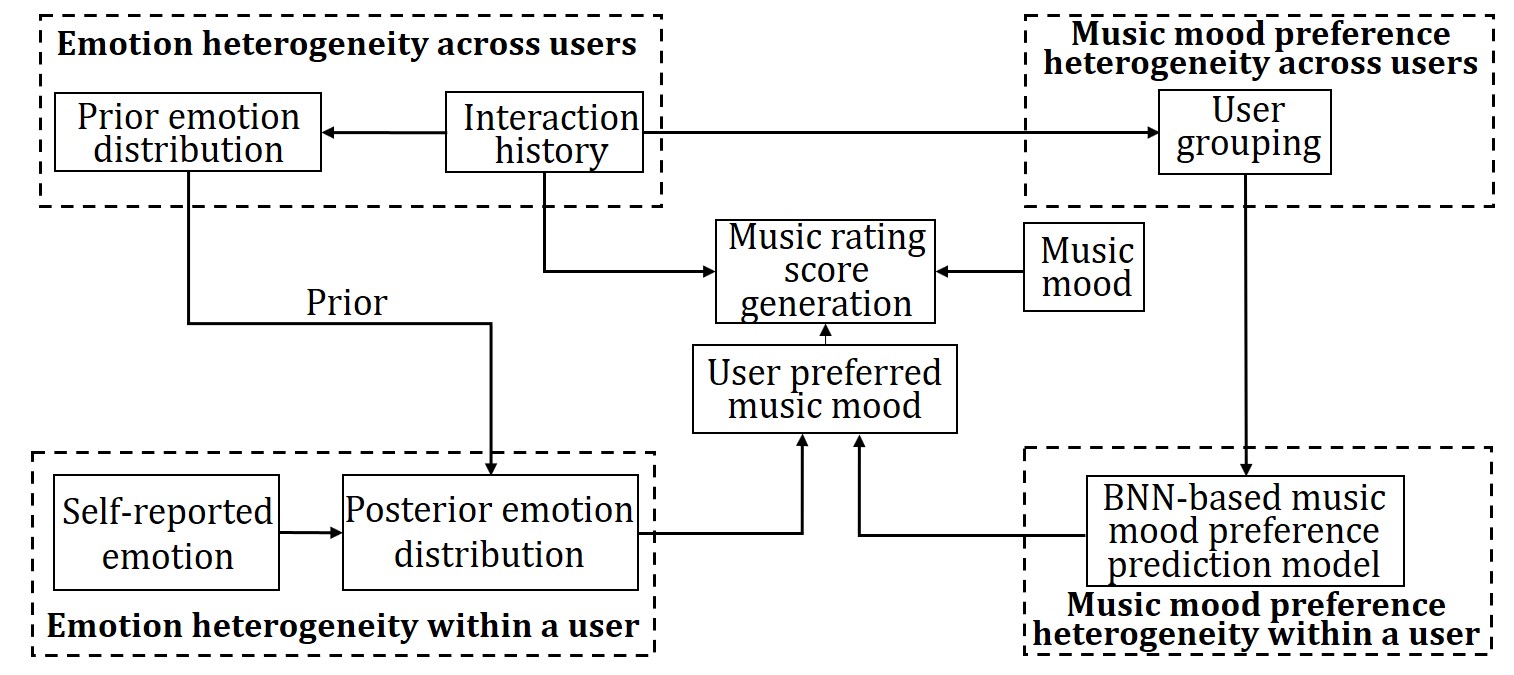}
  \caption{Conceptual framework of HDBN. The contents in the dotted boxes correspond to our approaches to address the four types of heterogeneity. After obtaining the predicted user’s preferred music mood, combined with the user interaction history and the true music mood label, the user’s rating score of the music can be generated.}
  \Description{}
  \label{framework}
\end{figure}

To address these gaps, we propose a generative model called Heterogeneity-aware Deep Bayesian Network (HDBN) that specifically models the four types of heterogeneity for personalized music recommendation. Research has shown that generative models can flexibly and intuitively integrate psychological and behavioral theories into the process of modeling user decision-making, more precisely model the conditional dependencies between variables and the joint probability distribution of users and items, and also improve model interpretability \cite{Wei2023, Ye2012, chen2022}. Consequently, our HDBN adopts the generative approach. The HDBN models the joint distribution of users, music, and latent emotional variables by mimicking the decision process in users’ music selection. Figure \ref{framework} shows the conceptual framework of our HDBN.

First, to address emotion heterogeneity across users, we design an inference network that learns a personalized prior emotion distribution for each user from the user’s music listening history. Users’ different personal experiences and cultural backgrounds lead to varied understandings of emotions \cite{Barrett2006}, which can be reflected in users’ behaviors like music listening. The inference network can capture personalized prior emotion distributions from users’ music listening behavior, addressing the emotion heterogeneity across users.

Second, to address emotion heterogeneity within a user, we design another inference network to infer the posterior emotion distribution from the user’s self-reported emotion tag before listening to music. The inference process is guided by the user’s prior emotion distribution. Studies show that a user’s self-reported emotion tag, such as happy, represents a range rather than a specific emotional value \cite{Barrett2009}. Representing the user’s self-reported emotions with the posterior emotion distribution increases flexibility, allowing us to extract varied emotion representations even when the same emotion tag is reported. Thus, we solved the issue of emotion heterogeneity within a user.

Third, to address music mood preference heterogeneity across users, we cluster users into groups based on their music listening history and design a Bayesian neural network (BNN) for each group to model users’ preferred music moods. User grouping provides sufficient data for each group to train a music mood preference prediction model. Users within a group share similar preferences, while differences exist between groups. Thus, training separate models for each group addresses the music mood preference heterogeneity across users.

Fourth, BNNs employ distributions instead of fixed values as model weights, which means that we get a different neural network each time we apply the model to predict music mood preferences for a user. This approach samples a set of specific model weights from the weight distributions each time a music mood preference prediction is made, allowing for more diverse predictions of music mood preferences. Thus, it addresses the problem of mood preference heterogeneity within a user. Meanwhile, this also addresses the issue of capturing subtle preference differences among users within the same user group.

To validate the effectiveness of our proposed model, we strived to seek a dataset that includes users’ self-reported emotion words preceding their music choices. Specifically, the dataset should ensure that users’ emotional states influence their music choices, not the music they listened to that influences their emotional states. Unfortunately, to the best of our knowledge, no existing dataset mets this requirement. As a result, we constructed a dataset called EmoMusicLJ based on LJ2M \cite{Liu2014} where users give self-reported emotions before selecting the music to listen to. LJ2M includes the users’ IDs, and the music tracks’ IDs, titles, and artists. Based on LJ2M, we crawl additional meta information such as music audio from\footnote{https://us.7digital.com/}, music genre, and released year from Musicovery\footnote{https://musicoveryb2b.mystrikinglu.com}. We then tagged the music with mood labels based on music audio and Emotify \cite{Anna2016}. Each music listening event includes a user ID, user self-reported emotion tag, music ID, as well as genre, release year, artist, and mood of music. We also derived a dataset with relative lower sparsity from EmoMusicLJ, called EmoMusicLJ-small, to validate the effectiveness of our method under different data sparsity levels.

The main contributions of our work are summarized as follows:
\begin{itemize}
    \item First, we propose the HDBN model, which explicitly accounts for the four types of heterogeneity for personalized music recommendation. Our model consists of four components: 1) an inference network to learn personalized prior emotion distributions for emotion heterogeneity across users, 2) an inference network to learn posterior emotion distributions for emotion heterogeneity within a user, 3) user grouping and establishing a Bayesian neural network (BNN) for each group for music mood preference heterogeneity across users, and 4) BNN-based music mood preference prediction for music mood preference heterogeneity within a user.
    \item Second, we constructed two datasets, EmoMusicLJ and EmoMusicLJ-small. EmoMusicLJ contains 154,742 interactions from 15,233 users and 6,180 music tracks, with a sparsity of 99.84\%. EmoMusicLJ-small contains 21,990 interactions from 3,473 users and 217 music tracks, with a sparsity of 97.08\%. Both datasets have been open-sourced and can serve as benchmark datasets for further studies on EMRSs.
    \item Third, we conducted extensive experiments comparing our model with state-of-the-art baseline models using metrics such as Hit Rate (HR), Precision, Normalized Discounted Cumulative Gain (NDCG), and Mean Reciprocal Rank (MRR). Experimental results demonstrated that our model significantly outperforms baseline models. Ablation studies demonstrated the effectiveness of the four components. Additionally, we conducted a human evaluation experiment and found that recommendations generated by our method were more widely accepted by users.
\end{itemize}

The remainder of this article is organized as follows. Section 2 provides a summary of related works and identified research gaps. Section 3 presents a detailed description of our model. Section 4 presents extensive experiments conducted against baseline models to validate the effectiveness of our method. Section 5 concludes the paper.

\section{Related Works}
In this section, we review the literature of music recommendation and emotion-aware music recommendation to form the basis of our research.

\subsection{Music recommendation}
Existing music recommendation methods can be categorized into four types: collaborative filtering (CF) methods, content-based methods, hybrid methods, and context-aware methods \cite{Wang2023, wang2021a, wang2021b}. Collaborative filtering consists of user-based CF and item-based CF. User-based CF methods recommend music based on the neighbor users who have similar music preferences (e.g., reflected by their listening history) to the target user, while item-based CF methods recommend music by measuring the similarity between music tracks based on their playing records \cite{Wang2023}. However, CF methods suffer from cold-start issues and also fail to leverage the rich, domain-specific features contained in music content \cite{Deldjoo2024}.

Content-based methods recommend music based on users’ preferences for music content, such as acoustic features, genre, lyrics, and instruments \cite{Wang2023, Deldjoo2024}. For example, Yao et al. \cite{Yao2024} used graph neural networks to learn the relationships between tag content and user, and music track. However, content-based methods may suggest tracks that do not fully align with users’ personal preferences \cite{Cheng2020}. To overcome these limitations, hybrid methods combine various approaches. For example, Magron et al. \cite{Margron2022} proposed a neural content-aware collaborative filtering framework (NCACF), which extracts content information from music acoustics features and models user and music embeddings from listening history using deep learning.

As music recommendation systems integrate various types of information, context-aware methods have been proposed to incorporate users' contextual factors, such as time, location, activity, weather, and emotion \cite{Wang2023, wang2021a, wang2021b}. For example, Wang et al. \cite{wang2021b} introduced a content- and context-aware method based on heterogeneous information networks and graph neural networks, which combines music content information with user behavior context, including listening history, play sequences, and listening sessions. User emotion, a crucial contextual factor influencing music listening decisions, has been shown to improve music recommendation performance and enhance user experience \cite{Chaitali2023, Revathy2023}. Music recommender systems that explicitly incorporate user emotion and music mood are referred to emotion-aware music recommender systems \cite{Deng2015, Marco2021}, which have attracted considerable interest from scholars \cite{Rushabh2023, Priyanka2023}. We reviewed existing studies on emotion-aware music recommendation in the next section.  

\subsection{Emotion-aware music recommendation}
Emotion is a crucial factor in music-listening decisions \cite{Marco2021}. With advances in technologies enabling emotion recognition from facial images, speech, social media footprints, and other sources, many studies have incorporated user emotion and music mood into music recommender systems to improve recommendation services. Table \ref{table1} summarizes the main studies on emotion-aware music recommendation.

    \begin{table}[htbp]
        \centering
        \captionof{table}{Main Research on Emotion-aware Music Recommender Systems}
        \label{table1}
        \renewcommand{\arraystretch}{1.25}
        \resizebox{\textwidth}{!}{
        \begin{tabular}{lp{0.09\textwidth}p{0.11\textwidth}p{0.57\textwidth}p{0.18\textwidth}} \toprule
        \textbf{Year} & \textbf{Author}  & \textbf{Emotion source} & \textbf{Method description} & \textbf{Method category in terms of emotion usage} \\ \midrule
    2024 & Han et al. \cite{Han2024} & Physiological indicators & Mapped user emotion and music mood to 3-D space (tension-arousal, energy-arousal, and valence). Recommended music based on the similarity between emotion and music mood. & Emotion-matching\\
    2023 & Wang et al. \cite{Wang2023} & Emotion tag & Incorporated user emotion and music mood into heterogeneous music graph. Used GNN to learn user interests and music content representations. & Emotion-as-a-feature\\
    2023 & Tran et al. \cite{Tran2023} & Facial image & Used DRViT and InvNet50 to predict user’s valance and arousal, and generated top-5 closet songs for user. & Emotion-matching\\
    2023 & Annam et al. \cite{Annam2024} & Facial image & Used VGG-16 to predict user facial emotion and recommend music based on a predefined relationship between music mood and user emotion. & Emotion-matching\\
    2022 & Bontempelli et al. \cite{Bontempelli2022} & Self-report & Predicted music mood with binary classifiers. Recommended music based on similarity score between music mood and user emotion. & Emotion-matching\\
    2022 & Li et al. \cite{Li2022} & Smart bracelets & Regarded user emotion prediction based on smart bracelets data as sub-task of music rating. Predicted user’s rating of music based on the predicted user emotion. & Emotion-as-a-feature\\
    2021 & Polignano et al. \cite{Marco2021} & Footprints on social media & Recommended music based on the coherence score between affective user profile (emotion) and not-affective music features (genre, lyrics). & Emotion-matching\\
    2020 & Shen et al. \cite{Shen2020} & Posts on WeChat & Modeled the interactions between user emotion and personality via hierarchical attention mechanism. & Emotion-as-a-feature\\
    2020 & Moscato et al. \cite{Moscato2020} & Online social network logs & Initialized user emotion from the big five traits and updated it based on initialized and recent emotion. Recommended music based on the distance between music mood and user emotion. & Emotion-matching\\
    2019 & Andjelkovic et al. \cite{Ivana2019} & Listening history &Recommended music based on similarity between user emotion and music mood first, and then based on music content similarity.& Emotion-matching\\
    2019 & Kang and Seo \cite{Kang2019} & Text on smart phone &Predicted user emotion from message. Recommended music based on similarity between user emotion and music mood.& Emotion-matching\\
    2017 & Iyer et al. \cite{Iyer2017} & Face image &Recommended music that gradually increases in happy mood.& Emotion-matching\\
    2015 & Deng et al. \cite{Deng2015} & Microblogs & Detected user emotion from microblogs. Recommended music by collaborative filtering method based on similarity between user emotion and music mood. & Emotion-matching\\
    \bottomrule
        \end{tabular}}
    \end{table}

According to the way to use emotion, existing methods can be grouped into two types: emotion-matching methods and emotion-as-a-feature methods. Emotion-matching methods infer user emotion from his/her behavioral data (e.g., facial images, online footprints) or self-reports and detect music moods with a recognition model. Then, recommendations are generated based on the similarity between user emotion and music mood. For example, Moscato et al. \cite{Moscato2020} initialized user emotions by encoding their personality traits into the pleasure-arousal-dominance (PAD) space and updated them based on the moods of the most recently listened to music represented in the same PAD space. Subsequently, they conducted music retrieval based on the Euclidean distance between user emotions and music moods.

In emotion-as-a-feature methods, user emotions and music moods are integrated into a music rating score prediction model (e.g., neural network), along with other information such as music meta-features. Recommendations are then generated based on the predicted scores. For example, Shen et al. \cite{Shen2020} proposed a Personality and Emotion Integrated Attentive (PEIA) model, which uses hierarchical attention to capture both the long-term and short-term effects of personality and emotion on music preferences. Recent studies have used heterogeneous graph to better aggregate contextual information and music content features \cite{Wang2023, Yu2024}. For example, Wang et al. \cite{Wang2023} built a heterogeneous music graph (HMG) integrating user listening behaviors and music content. A multi-view enhanced graph attention network is then used to learn representations of user interests and music features from different user views and music views. User emotion and music mood can also be effectively integrated into the HMG.

While valuable, both types of methods have four limitations.
\begin{itemize}
    \item First, they fail to consider the problem of emotion heterogeneity across users, which refers to the fact that the prior emotion distributions vary across different users. For instance, some individuals are more likely to feel irritated than others \cite{John2000}. Existing methods use the same emotion tags or high-dimensional vectors to represent the same emotions among users.
    \item Second, they fail to consider the problem of emotion heterogeneity within a user, which refers to the variation in a user’s emotions expressed using the same emotion tag at different times. As highlighted by Barrett’s Conceptual Act model \cite{Barrett2009}, the emotions expressed by a user using the same emotion tag may vary depending on different combinations of conceptual instances in different contexts and experiences. Similarly, existing methods fail to consider this problem since they model user emotion using the same tag or vectors.
    \item Third, they fail to consider the problem of music mood preference heterogeneity across users, which refers to the fact that different users under the same emotion may prefer different music moods. This has been validated by many studies. For example, Shifriss et al. \cite{Shifriss2015} found that when people were in a bad emotion (e.g., sad, depressed), they chose happy music or sad music differently depending on the purpose they set out. However, existing methods typically use the same music mood preference model for different users, whether in emotion matching or emotion-as-a-feature approaches, and thus neglecting this type of heterogeneity.
    \item Fourth, they fail to consider the problem of music mood preference heterogeneity within a user, which refers to the fact that a user under a given emotion at different times may prefer different music moods. Studies have shown that factors such as context can influence users’ music selections, regardless of their inherent music preferences, leading to this heterogeneity \cite{Patrik2008, Li2022}. Existing methods rely on a fixed music mood preference prediction model to represent users’ inherent music preference patterns, thus overlooking the music mood preference heterogeneity within a user.
\end{itemize}

In summary, although many emotion-aware recommendation methods have been proposed to leverage user emotions and music moods, they fail to consider 1) emotion heterogeneity across users, 2) emotion heterogeneity within a user, 3) music mood preference heterogeneity across users, and 4) music mood preference heterogeneity within a user. Given these limitations, this study aims to design a novel emotion-aware recommendation method that addresses them, thereby improving recommendation results.

\section{THE PROPOSED HETEROGENEITY-AWARE DEEP BAYESIAN NETWORK (HDBN)}
We denote the user set as $\mathbb{U}=\{u_1,u_2,\ldots,u_U\}$, where $U$ represents the number of users, and $u_i\in \mathbb{U}$ denotes a user. We denote the music set as $\mathbb{V}=\{v_1,v_2,\ldots,v_V\}$, where $V$ represents the number of music, and $v_i \in \mathbb{V}$ denotes a music track. We denote the users’ emotion tag set as $\vmathbb{e}=\{e_1,e_2,\ldots,e_M\}$, where $M$ represents the number of emotion tags, and $e_i \in \vmathbb{e}$ denotes an emotion tag. There are a series of user-emotion-music triplets, which form the dataset $\mathcal{D}=\{(u,e,v)_i\}_{i=1}^N$. $N$ denotes the size of the dataset. Each triplet means a user $u$, under an emotion $e$, chooses to listen to music $v$. In this study, the personalized music recommendation aims to predict a user’s preference for music based on his or her emotions and then generate a top-$T$ recommendation list. Next, we describe our proposed method.

\subsection{Method}
\subsubsection{An overview of HDBN}
\begin{figure}[h]
  \centering
  \includegraphics[width=0.9\linewidth]{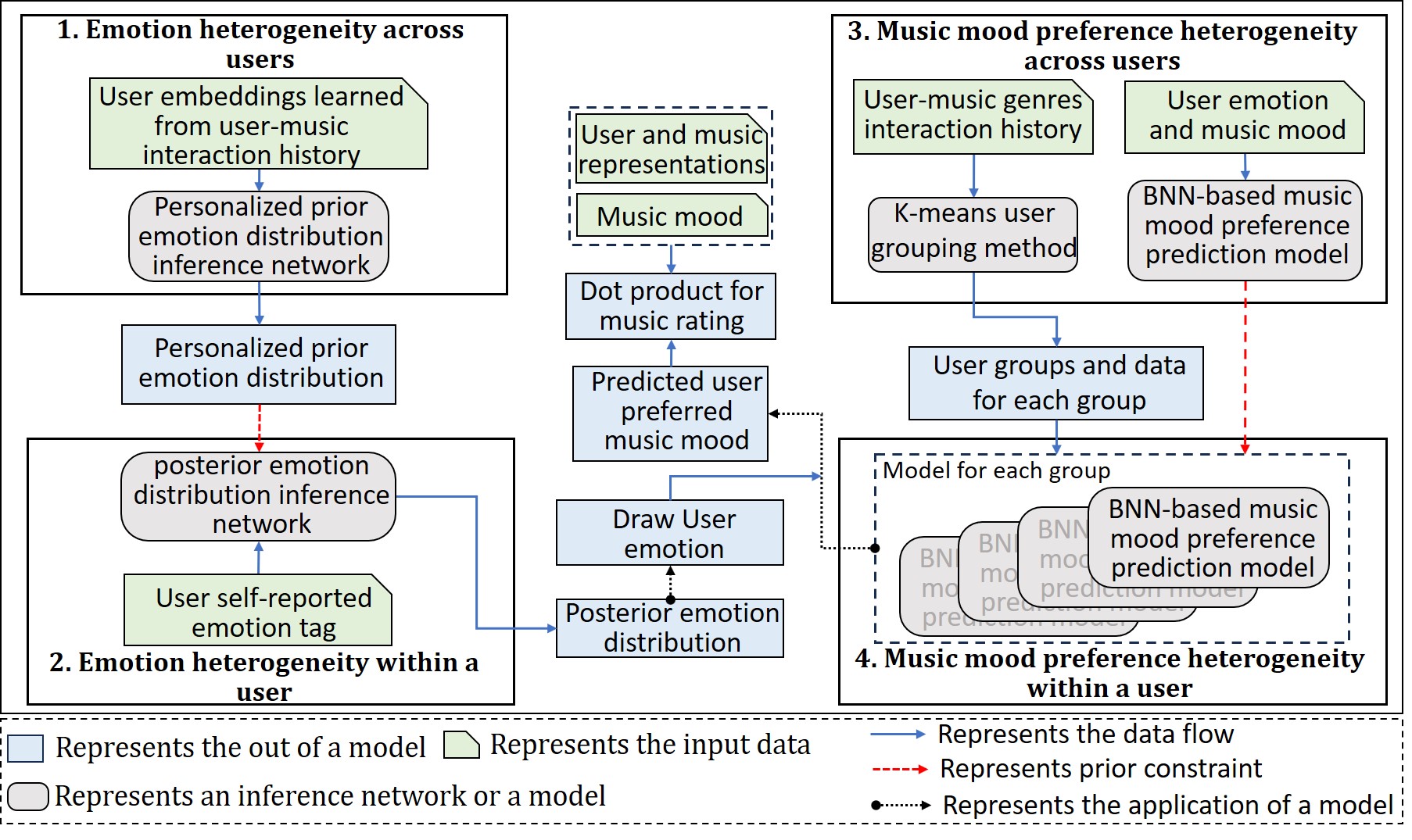}
  \caption{An overview of the HDBN model flow.}
  \Description{the four component of HDBN and its data flow.}
  \label{overview}
\end{figure}

HDBN is a generative model including four key components to address the four limitations that we have identified in the literature review. First, we design a personalized prior emotion distribution for each user to address emotion heterogeneity across users; second, we design a posterior emotion distribution for a specific user’s self-reported emotion tag (e.g., happy) to address emotion heterogeneity within a user; third, we cluster users into different groups and design a BNN for each group to address the music mood preference heterogeneity across users; fourth, BNN exhibits diverse music mood preference predictions to address the music mood preference heterogeneity within a user. Figure \ref{overview} presents an overview of the model flow. Next, we detail the generative process of HDBN.

\subsubsection{The generative process}
In this section, we describe the generative process of HDBN, which addresses the four types of heterogeneity problems.

\noindent \textit{a) General latent emotion distribution for all users}

According to Conceptual Act Model \cite{Barrett2009}, user emotions are not atomic concepts described by emotion words, such as happy and fearful, but are composed of more fundamental primitive psychological elements, thus forming a shape (e.g., roughly circular or circumplex shape) in a psychological space \cite{Barrett2006, Barrett2006b}. Hence, we assume that the observed user self-reported emotion is from a latent emotion distribution (LED), which reflects how the self-reported emotions are distributed \cite{John2000, Huang2017, Zeng2021}. Influenced by cultural and social attributes, commonalities exist in the cognition of emotion across different users \cite{Barrett2006}. Hence, we assume that there is a general LED for all users, denoted as $\boldsymbol{\mathcal{S}}$. As suggested by Zeng et al. \cite{Zeng2021}, we treat $\boldsymbol{\mathcal{S}}$ as a standard Gaussian distribution, i.e., $\boldsymbol{\mathcal{S}}=\mathcal{N}(\boldsymbol{\mathcal{S}};\boldsymbol{\mu},\boldsymbol{\sigma}^2)$ where $\boldsymbol{\mu}=\boldsymbol{0}$ and $\boldsymbol{\sigma}=\boldsymbol{1}$.

\noindent \textit{b) Personalized prior LED to model emotion heterogeneity across users}

As aforementioned, different people have different emotion patterns. For instance, extraverted users are more likely to become happy while neurotic ones are more likely to be worried \cite{Hughes2020}. Hence, LED should be personalized. We call the personalized LED of user $u$ as $u$'s prior LED, denoted as $\boldsymbol{\mathcal{S}}_u$ which also follows a Gaussian distribution. The general LED $\boldsymbol{\mathcal{S}}$ is the prior for $\boldsymbol{\mathcal{S}}_u$. Formally, $\boldsymbol{\mathcal{S}}_u=\mathcal{N}(\boldsymbol{\mathcal{S}}_u;\boldsymbol{\mu}_u,\boldsymbol{\sigma}_1^2 )$ where $\boldsymbol{\mu}_u$ is from the distribution $\boldsymbol{\mathcal{S}}$, i.e., $\boldsymbol{\mu}_u\sim\boldsymbol{\mathcal{S}}_u$. By allowing each user to have their own prior LED $\boldsymbol{\mathcal{S}}_u$, we address the problem of emotion heterogeneity across users.

\noindent \textit{c) Posterior LED to model emotion heterogeneity within a user}

A user's emotion varies from time to time, depending on the specific experiences she/he has. Hence, the true meanings of a user’s self-reported emotion when she/he chooses to listen to different music also change. Hence, we treat $u$'s prior LED (i.e., $\boldsymbol{\mathcal{S}}_u$) as the prior to generate a specific LED for each of a user’s self-reported emotion when listening to music $v$, called posterior LED and denoted as $\boldsymbol{\mathcal{S}}_{u,v}$. Formally, $\boldsymbol{\mathcal{S}}_{u,v}=\mathcal{N}(\boldsymbol{\mathcal{S}}_{u,v};\boldsymbol{\mu}_{u,v},\boldsymbol{\sigma}_2^2 )$ where $\boldsymbol{\mu}_{u,v}$ is from the distribution $\boldsymbol{\mathcal{S}}_u$, i.e., $\boldsymbol{\mathcal{\mu}}_{u,v}\sim \boldsymbol{\mathcal{S}}_u$. Since a user’s self-reported emotion drawn from the corresponding posterior LED varies at different times, we address the problem of emotion heterogeneity within a user.

A user's self-reported emotion, denoted as $\boldsymbol{s}_{u,v}$, is regarded as an observed sample drawn from the posterior LED $\boldsymbol{\mathcal{S}}_{u,v}$, i.e., $\boldsymbol{s}_{u,v} \sim \boldsymbol{\mathcal{S}}_{u,v}$. This study adopts the reparameterization trick to generate samples of $\boldsymbol{\mathcal{s}}_{u,v}$ \cite{Diederik2014, Chen2019}. Formally, we first draw a random variable $\boldsymbol{\epsilon}$ from the standard Gaussian distribution $\boldsymbol{\epsilon}\sim\mathcal{N}(\boldsymbol{0},\boldsymbol{1})$, and then generate a sample of $\boldsymbol{s}_{u,v}$ based on $\boldsymbol{s}_{u,v}=\boldsymbol{\mu}_{u,v}+\boldsymbol{\sigma}_2\cdot\boldsymbol{\epsilon}$.

\noindent \textit{d) User grouping and separate modeling to model music mood preference heterogeneity across users}

As mentioned, even under the same emotion (i.e., under the same posterior LED), different users may prefer music with different moods, which we refer to as music mood preference heterogeneity across users. For example, users who scored high on openness and agreeableness are more likely to choose music with higher activating moods (e.g., high loudness), while highly neurotic users prefer music with lower activating moods  \cite{Alessandro2020}. Hence, we first cluster users into different groups. Specifically, assuming there are $G$ groups of users and the group ID of user $u$ is $g=\textit{g}(u)$. $\textit{g}(u)$ can be inferred from the user’s listening history with clustering algorithms such as K-means. Then, we create a separate music mood prediction model for each group. Across groups, the prediction model varies significantly to account for the music mood preference heterogeneity across users (groups).

\noindent \textit{e) BNN-based music mood preference prediction model to model music mood preference heterogeneity within a user}

A user may prefer different music moods under the same self-report emotion at different times, which we refer to as music mood preference heterogeneity within a user. Hence, the music mood preference prediction model should be flexible to accommodate the heterogeneity. Motivated by this, we model the music mood preference prediction of each group using Bayesian neural networks (BNN). The last layer of the BNN is a softmax layer, ensuring the sum of the probability for each music mood equals one. As BNN assumes that the model parameters are distributions rather than fixed values, the prediction results vary even given the same emotion input. In this way, we account for music mood preference heterogeneity within a user in the music selection process.

The music mood preference prediction model for group $g$ is a function parameterized by $\boldsymbol{\psi}_g$, denoted as $\mathcal{F}_{\boldsymbol{\psi}_g}$. Although music mood preferences vary across user groups, universal preference patterns exist. For example, Park et al. \cite{Minsu2019} analyzed 765 million music-listening events from users across 51 countries and found that people tend to prefer more relaxing and less intense music late at night. Hence, we assume each $\boldsymbol{\psi}_g$ is generated from a shared BNN, parameterized by $\boldsymbol{\psi}$. Following \cite{Charles2015, Andrew2020}, we assume $\boldsymbol{\psi}$ follows a standard Gaussian distribution, i.e., $\boldsymbol{\psi}\sim\boldsymbol{\mathcal{N}}(\boldsymbol{\psi};\boldsymbol{\mu}_{\boldsymbol{\psi}},\boldsymbol{\sigma}_{\boldsymbol{\psi}}^2)$ where $\boldsymbol{\mu}_{\boldsymbol{\psi}}=\boldsymbol{0}$ and $\boldsymbol{\sigma}_{\boldsymbol{\psi}}=\boldsymbol{1}$. Similarly, $\boldsymbol{\psi}_g$ follows a Gaussian distribution with $\boldsymbol{\psi}$ as the expectation, i.e., $\boldsymbol{\psi}_g\sim \mathcal{N}(\boldsymbol{\psi}_g;\boldsymbol{\psi},\boldsymbol{\sigma}_3^2 )$. Then the user preferred music mood (denoted as $\boldsymbol{l}_{u,v}$) is generated based on the observed emotion $\boldsymbol{s}_{u,v}$, i.e., $\boldsymbol{l}_{u,v}=\mathcal{F}_{\boldsymbol{\psi}_\textit{g}(u)} (\boldsymbol{s}_{u,v})$.

\noindent \textit{f) Generating music preferred by the user}

A music track is more likely to be selected if its actual music mood $\boldsymbol{o}_v$ is close to the user-preferred music mood $\boldsymbol{l}_{u,v}$. Additionally, a user may prefer a certain type of music such as a certain genre or from a certain singer. Such factors can be reflected in the user’s listening history which consists of user-music interactions. We employ a neural network (i.e., embedding layer) to learn the interaction-based user representation, $\boldsymbol{r}_u$, and music representation, $\boldsymbol{r}_v$, from the rich user-music interaction data. Then, the matching score $m_{u,v}$ is generated by a dot product, i.e., $m_{u,v}=[\boldsymbol{l}_{u,v},\boldsymbol{r}_u ] [\boldsymbol{o}_v,\boldsymbol{r}_v ]^T$.

\begin{figure}[h]
  \centering
  \includegraphics[width=0.98\linewidth]{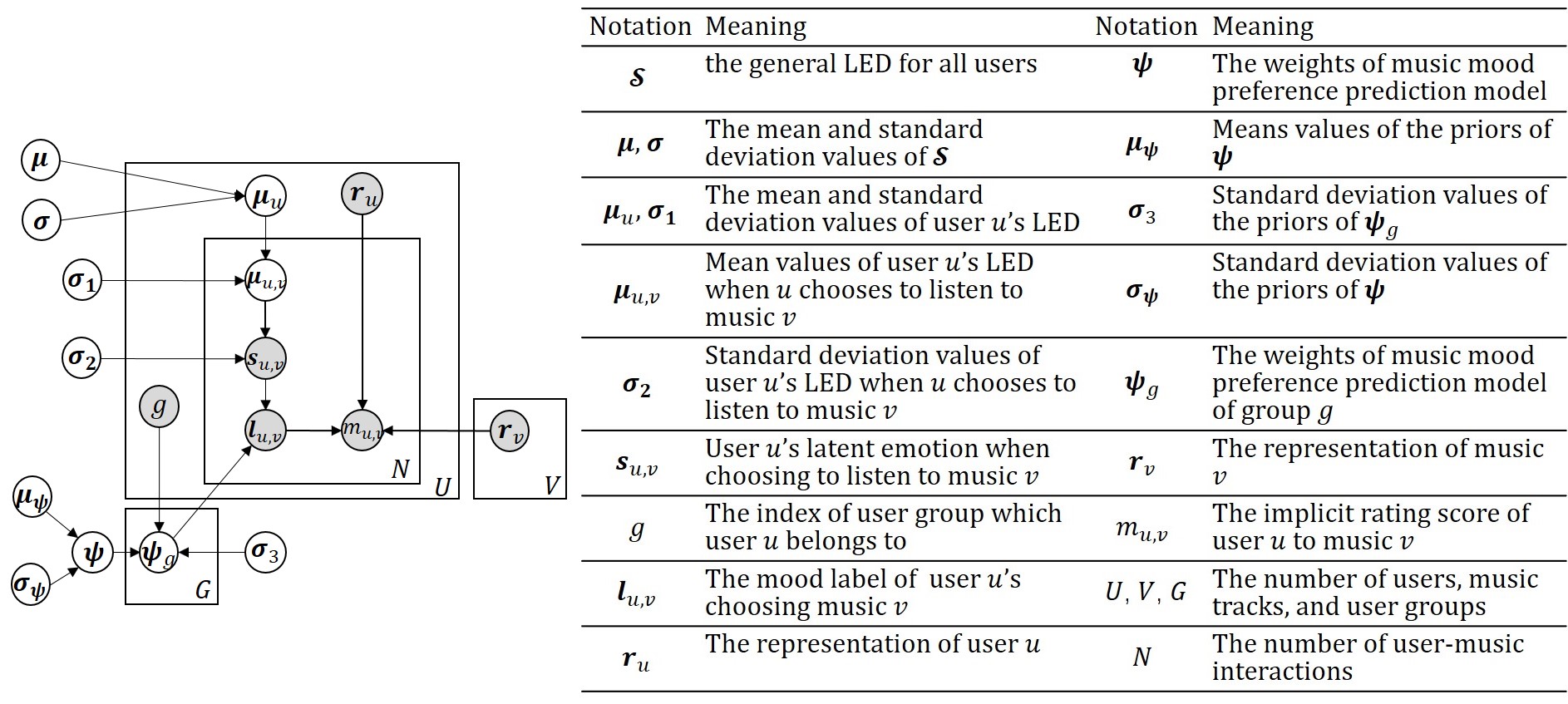}
  \caption{Graphical representation and notation description of the generative process.}
  \Description{the generation flow of our model.}
  \label{generative_process}
\end{figure}

The generative process is outlined in Algorithm \ref{algorithm1} and visualized in Figure \ref{generative_process}. The shaded circles represent observable variables from the data, while the other circles represent hidden variables. Note that music representation $\boldsymbol{r}_v$ and user representation $\boldsymbol{r}_u$ were extracted by an external existing neural network (i.e., the embedding layer) based on user-music interactions and hence their values are observable in the generative process. The group ID $g$ of a user is obtained with the K-means, so $g$ is also observable.

\begin{algorithm}[h]
    \caption{The Generative Process}
    \label{algorithm1}
    \begin{flushleft}
    for a user $u\in\mathbb{U}$:\\
    \hspace{5mm} \textcolor{gray}{// look up user representation from the embedding matrix}\\
    \hspace{5mm}Get user representation $\boldsymbol{r}_u=NN_u(u)$\\
    \hspace{5mm} \textcolor{gray}{// generate user-specific LED based on the general prior of LED $\boldsymbol{\mathcal{S}}(\boldsymbol{0},\boldsymbol{1})$}\\
    \hspace{5mm}Get user $u$'s personalized prior latent emotion distribution $\boldsymbol{\mathcal{S}}_u=\mathcal{N}(\boldsymbol{\mathcal{S}}_u;\boldsymbol{\mu}_u,\boldsymbol{\sigma}_1^2), \boldsymbol{\mu}_u\sim\boldsymbol{\mathcal{S}}(\boldsymbol{0},\boldsymbol{1})$\\
    \hspace{5mm} \textcolor{gray}{//generate self-reported emotion-specific LED based on $\boldsymbol{\mathcal{S}}_u$}\\
    \hspace{5mm} Get user $u$’s specific posterior latent emotion distribution $\boldsymbol{\mathcal{S}}_{u,v}=\mathcal{N}(\boldsymbol{\mathcal{S}}_{u,v};\boldsymbol{\mu}_{u,v},\boldsymbol{\sigma}_2^2 ), \boldsymbol{\mu}_{u,v}\sim\boldsymbol{\mathcal{S}}_u$\\
    \hspace{5mm} \textcolor{gray}{//sample user $u$’s LED from his/her posterior LED}\\
    \hspace{5mm}Draw user-specific emotion $\boldsymbol{s}_{u,v}\sim\boldsymbol{\mathcal{S}}_{u,v}$\\
    \hspace{5mm} \textcolor{gray}{//look up user’s group index}\\
    \hspace{5mm}Get user $u$’s group $g=\textit{g}(u)$\\
    \hspace{5mm} \textcolor{gray}{// reparameterization trick}\\
    \hspace{5mm}Draw music mood preference prediction model for $u$ $\boldsymbol{\psi}_g\sim\mathcal{N}(\boldsymbol{\psi}_g;\boldsymbol{\psi},\boldsymbol{\sigma}_3^2)$\\
    \hspace{5mm} \textcolor{gray}{// predict user preferred music mood with group-specific model $\mathcal{F}_{\boldsymbol{\psi}_g}(\boldsymbol{s}_{u,v})$}\\
    \hspace{5mm}Get user $u$’s preferred music mood $\boldsymbol{l}_{u,v}=\mathcal{F}_{\boldsymbol{\psi}_g}(\boldsymbol{s}_{u,v})$\\
    \hspace{5mm}for a music $v\in\mathbb{V}$:\\
    \hspace{10mm} \textcolor{gray}{// look up music representation from the embedding matrix}\\
    \hspace{10mm}Get music representation $\boldsymbol{r}_v=NN_v(v)$\\
    \hspace{10mm} \textcolor{gray}{// look up music mood from the dataset}\\
    \hspace{10mm}Get music mood $\boldsymbol{o}_v$\\
    \hspace{10mm}for $u$’s preference score to $v$:\\
    \hspace{15mm} \textcolor{gray}{// compute the user $u$’s implicit rating score to music $v$}\\
    \hspace{15mm}$m_{u,v}=[\boldsymbol{l}_{u,v},\boldsymbol{r}_u][\boldsymbol{o}_v,\boldsymbol{r}_v ]^T$\\
    \end{flushleft}
\end{algorithm}

\subsubsection{Variational inference}In this section, we provide the inference methods for all variational posterior distributions in the generation process, as well as the learning approach for all inference network parameters.

\noindent \textit{a) Derivation and decomposition of the optimization objective}

The generative process can be reached by maximizing the logarithmic likelihood of the observed $m_{u,v}$ and $\boldsymbol{s}_{u,v}$, $\forall u\in\mathbb{U}, v\in \mathbb{V}$, which is given by (See Appendix \ref{appA11} for details),
\begin{equation}\label{equa1}
    \begin{split}
        \log p(\{m_{u,v}\},\{\boldsymbol{s}_{u,v}\})&=\mathbb{E}_q\left[\log p(\{m_{u,v}\},\{\boldsymbol{s}_{u,v}\},\{\boldsymbol{\psi}_g\},\boldsymbol{\psi},\{\boldsymbol{\mu}_{u,v}\},\{\boldsymbol{\mu}_{u}\},\{\boldsymbol{l}_{u,v}\})-\log q(\{\boldsymbol{\psi}_g\},\boldsymbol{\psi},\{\boldsymbol{\mu}_{u,v}\},\{\boldsymbol{\mu}_{u}\},\{\boldsymbol{l}_{u,v}\})\right]\\
        &\quad+\mathrm{KL}\left(q(\{\boldsymbol{\psi}_g\},\boldsymbol{\psi},\{\boldsymbol{\mu}_{u,v}\},\{\boldsymbol{\mu}_{u}\},\{\boldsymbol{l}_{u,v}\})\parallel p(\{\boldsymbol{\psi}_g\},\boldsymbol{\psi},\{\boldsymbol{\mu}_{u,v}\},\{\boldsymbol{\mu}_{u}\},\{\boldsymbol{l}_{u,v}\}|\{m_{u,v}\},\{\boldsymbol{s}_{u,v}\})\right)\\
    \end{split}
\end{equation}
where $\{m_{u,v}\}$ is the collection of all $m_{u,v}$, and we use similar notations for the other variables. $p(\{m_{u,v}\},\{\boldsymbol{s}_{u,v}\},\{\boldsymbol{\psi}_g\},\boldsymbol{\psi},\{\boldsymbol{\mu}_{u,v}\},\{\boldsymbol{\mu}_{u}\},\{\boldsymbol{l}_{u,v}\})$ and $q(\{\boldsymbol{\psi}_g\},\boldsymbol{\psi},\{\boldsymbol{\mu}_{u,v}\},\{\boldsymbol{\mu}_{u}\},\{\boldsymbol{l}_{u,v}\})$ are the posterior and variational distributions of $\{\boldsymbol{\psi}_g\},\boldsymbol{\psi},\{\boldsymbol{\mu}_{u,v}\},\{\boldsymbol{\mu}_{u}\},\{\boldsymbol{l}_{u,v}\}$, respectively. Since the Kullback-Leibler (KL) divergence is non-negative, we have:
\begin{equation}\label{equa2}
    \begin{split}
        \log p(\{m_{u,v}\},\{\boldsymbol{s}_{u,v}\})\geq\mathbb{E}_q\left[\log p(\{m_{u,v}\},\{\boldsymbol{s}_{u,v}\},\{\boldsymbol{\psi}_g\},\boldsymbol{\psi},\{\boldsymbol{\mu}_{u,v}\},\{\boldsymbol{\mu}_{u}\},\{\boldsymbol{l}_{u,v}\})-\log q(\{\boldsymbol{\psi}_g\},\boldsymbol{\psi},\{\boldsymbol{\mu}_{u,v}\},\{\boldsymbol{\mu}_{u}\},\{\boldsymbol{l}_{u,v}\})\right]
    \end{split}
\end{equation}

Hence, we obtain a lower bound for $\log p(\{m_{u,v}\},\{\boldsymbol{s}_{u,v}\})$. This is called evidential lower bound (ELBO), and we denote it as ELBO($q$). ELBO($q$) can be further transformed as (details in Appendix \ref{appA12}):
\begin{equation}\label{equa3}
    \begin{split}
        \mathrm{ELBO}(q)=&\mathbb{E}_q\left[\log p(\{m_{u,v}\},\{\boldsymbol{s}_{u,v}\}\mid \{\boldsymbol{\psi}_g\},\boldsymbol{\psi},\{\boldsymbol{\mu}_{u,v}\},\{\boldsymbol{\mu}_{u}\},\{\boldsymbol{l}_{u,v}\})\right ]\\ &-\mathrm{KL}\left(q(\{\boldsymbol{\psi}_g\},\boldsymbol{\psi},\{\boldsymbol{\mu}_{u,v}\},\{\boldsymbol{\mu}_{u}\},\{\boldsymbol{l}_{u,v}\})\parallel p(\{\boldsymbol{\psi}_g\},\boldsymbol{\psi},\{\boldsymbol{\mu}_{u,v}\},\{\boldsymbol{\mu}_{u}\},\{\boldsymbol{l}_{u,v}\})\right)\\
    \end{split}
\end{equation}
where $p(\{\boldsymbol{\psi}_g\},\boldsymbol{\psi},\{\boldsymbol{\mu}_{u,v}\},\{\boldsymbol{\mu}_{u}\},\{\boldsymbol{l}_{u,v}\})$ is the prior distribution for $\{\boldsymbol{\psi}_g\},\boldsymbol{\psi},\{\boldsymbol{\mu}_{u,v}\},\{\boldsymbol{\mu}_{u}\},\{\boldsymbol{l}_{u,v}\}$. According to the conditional independence relationship in the generative process, $p(\{\boldsymbol{\psi}_g\},\boldsymbol{\psi},\{\boldsymbol{\mu}_{u,v}\},\{\boldsymbol{\mu}_{u}\},\{\boldsymbol{l}_{u,v}\})$ can be factorized as,
\begin{equation}\label{equa4}
    \begin{split}
        p(\{\boldsymbol{\psi}_g\},\boldsymbol{\psi},\{\boldsymbol{\mu}_{u,v}\},\{\boldsymbol{\mu}_{u}\},\{\boldsymbol{l}_{u,v}\})=p(\boldsymbol{\psi})\prod_g{p(\boldsymbol{\psi}_g\mid \boldsymbol{\psi})}\prod_u\left(p(\boldsymbol{\mu}_{u}\mid \boldsymbol{\mu})\prod_vp(\boldsymbol{\mu}_{u,v}\mid \boldsymbol{\mu}_u)\right)\prod_u\prod_vp\left(\boldsymbol{l}_{u,v}\mid \mathcal{F}_{\boldsymbol{\psi}_{\textit{g}(u)}}(\boldsymbol{s}_{u,v})\right)
    \end{split}
\end{equation}

As described in the generative process, $p(\boldsymbol{\psi})$, $p(\boldsymbol{\psi}_g\mid\boldsymbol{\psi})$, $p(\boldsymbol{\mu}_{u}\mid \boldsymbol{\mu})$, and $p(\boldsymbol{\mu}_{u,v}\mid \boldsymbol{\mu}_u)$ are all Gaussian distributions that have been introduced. $p(\boldsymbol{l}_{u,v}\mid \mathcal{F}_{\boldsymbol{\psi}_{\textit{g}(u)}}(\boldsymbol{s}_{u,v}))$ follows the distribution in line with $\boldsymbol{o}_v$. We assume that $q(\{\boldsymbol{\psi}_g\},\boldsymbol{\psi},\{\boldsymbol{\mu}_{u,v}\},\{\boldsymbol{\mu}_{u}\},\{\boldsymbol{l}_{u,v}\})$ has the same factorization property as the prior distribution in Equation (4), i.e.,
\begin{equation}\label{equa5}
    \begin{split}
        q(\{\boldsymbol{\psi}_g\},\boldsymbol{\psi},\{\boldsymbol{\mu}_{u,v}\},\{\boldsymbol{\mu}_{u}\},\{\boldsymbol{l}_{u,v}\})=q(\boldsymbol{\psi})\prod_g{q(\boldsymbol{\psi}_g\mid \boldsymbol{\psi})}\prod_u\left(q(\boldsymbol{\mu}_{u}\mid \boldsymbol{\mu})\prod_vq(\boldsymbol{\mu}_{u,v}\mid \boldsymbol{\mu}_u)\right)\prod_u\prod_vq\left(\boldsymbol{l}_{u,v}\mid \mathcal{F}_{\boldsymbol{\psi}_{\textit{g}(u)}}(\boldsymbol{s}_{u,v})\right)
    \end{split}
\end{equation}

For $q(\boldsymbol{\psi})$, the goal is to obtain BNN parameters, $\boldsymbol{\psi}$. Hence, the inference process is the same as learning the parameters of the BNNs. Hence, we can adopt existing methods such as Bayes by Backprop \cite{Charles2015} to learn $\boldsymbol{\psi}$. For $q(\boldsymbol{\psi}_g\mid\boldsymbol{\psi})$, $\boldsymbol{\psi}_g$ is generated based on $\boldsymbol{\psi}$. Hence, we initialize $\boldsymbol{\psi}_g$ as $\boldsymbol{\psi}$ and then update $\boldsymbol{\psi}_g$. The update process is the same as learning BNN parameters. For $q(\boldsymbol{\mu}_{u}\mid \boldsymbol{\mu})$, $q(\boldsymbol{\mu}_{u,v}\mid \boldsymbol{\mu}_u)$ and $q\left(\boldsymbol{l}_{u,v}\mid \mathcal{F}_{\boldsymbol{\psi}_{\textit{g}(u)}}(\boldsymbol{s}_{u,v})\right)$, we adopt neural networks for inference. Next, we describe the inference networks in our study.

\noindent \textit{b) Inference networks for} $\boldsymbol{\mu}_u$ \textit{and} $\boldsymbol{\mu}_{u,v}$

\begin{figure}[h]
  \centering
  \includegraphics[width=0.5\linewidth]{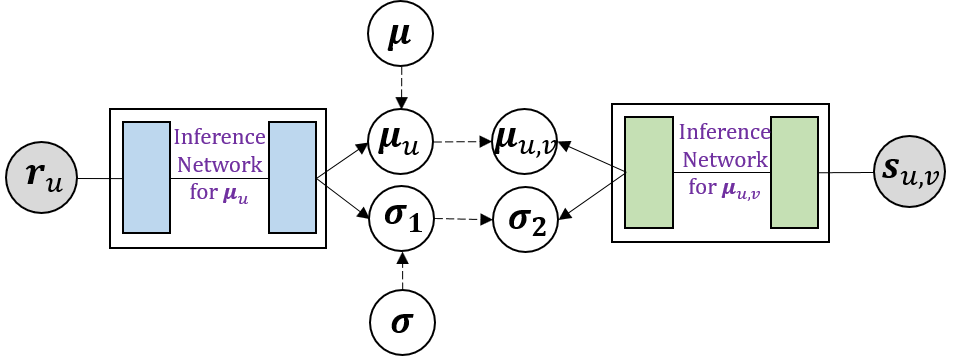}
  \caption{The inference networks for $\boldsymbol{\mu}_u$ and $\boldsymbol{\mu}_{u,v}$. The dashed arrows represent the regulations from the prior distribution.}
  \Description{Four limitations of existing emotion-aware music recommendation and the corresponding solutions of our method.}
  \label{SVAE}
\end{figure}

Since the user self-reported emotion $\boldsymbol{s}_{u,v}$ contains rich cues about $\boldsymbol{\mu}_{u,v}$, we adopt an inference network to infer $\boldsymbol{\mu}_{u,v}$ based on $\boldsymbol{s}_{u,v}$. Similarly, since user representation $\boldsymbol{r}_u$ is closely related to $\boldsymbol{\mu}_u$, we infer $\boldsymbol{\mu}_u$ based on user representation $\boldsymbol{r}_u$. Since $\boldsymbol{\mu}_{u,v}$ is generated from $\boldsymbol{\mu}_u$, we simultaneously infer $\boldsymbol{\mu}_{u,v}$ and $\boldsymbol{\mu}_u$ with a structured process. The first network is used to infer $\boldsymbol{\mu}_u$ while the second one is used to infer $\boldsymbol{\mu}_{u,v}$ based on $\boldsymbol{\mu}_u$. Figure \ref{SVAE} visually represents the inference networks.

As mentioned above, $\boldsymbol{\mathcal{S}}_u$ follows a standard Gaussian distribution. Hence, there are two parameters, i.e. the mean vector $\boldsymbol{\mu}_u$ and the standard deviation vector $\boldsymbol{\sigma}_1$, to be inferred. We encode the variational distribution of $\boldsymbol{\mathcal{S}}_u$ from $\boldsymbol{r}_u$ using a shared multi-layer neural network with parameters of $\boldsymbol{\theta}_1$. This process is denoted as:
\begin{equation}\label{equa6}
    \boldsymbol{\mu}_u=NN_{\boldsymbol{\theta}_1}(\boldsymbol{r}_u)
\end{equation}
\begin{equation}\label{equa7}
    \boldsymbol{\sigma}_1=NN_{\boldsymbol{\theta}_1}(\boldsymbol{r}_u)
\end{equation}
\begin{equation}\label{equa8}
    q(\boldsymbol{\mu}_u\mid \boldsymbol{\mu})=\mathcal{N}(\boldsymbol{\mu}_u,\boldsymbol{\sigma}_1^2)
\end{equation}

Similarly, we encode the variational distribution of $\boldsymbol{\mathcal{S}}_{u,v}$ which consists of the mean vector $\boldsymbol{\mu}_{u,v}$ and standard deviation vector $\boldsymbol{\sigma}_2$ from $\boldsymbol{s}_{u,v}$ using a shared multi-layer neural network with parameters of $\boldsymbol{\theta}_2$. This process is denoted as:
\begin{equation}\label{equa9}
    \boldsymbol{\mu}_{u,v}=NN_{\boldsymbol{\theta}_2}(\boldsymbol{s}_{u,v})
\end{equation}
\begin{equation}\label{equa10}
    \boldsymbol{\sigma}_2=NN_{\boldsymbol{\theta}_2}(\boldsymbol{s}_{u,v})
\end{equation}
\begin{equation}\label{equa11}
    q(\boldsymbol{\mu}_{u,v}\mid \boldsymbol{\mu}_u)=\mathcal{N}(\boldsymbol{\mu}_{u,v},\boldsymbol{\sigma}_2^2)
\end{equation}

\noindent \textit{c) Inference networks for} $\boldsymbol{l}_{u,v}$

Since the generation process for $\boldsymbol{l}_{u,v}$ is deterministic, it can be determined given $\boldsymbol{s}_{u,v}$ and the BNN parameterized by $\boldsymbol{\psi}_g$, i.e., $\boldsymbol{l}_{u,v}=\mathcal{F}_{\boldsymbol{\psi}_g}(\boldsymbol{s}_{u,v})$.

After obtaining the value of hidden variables including $\{\boldsymbol{\psi}_g\},\boldsymbol{\psi},\{\boldsymbol{\mu}_{u,v}\},\{\boldsymbol{\mu}_{u}\}$, and $\{\boldsymbol{l}_{u,v}\}$, we can compute the probability $p(\{m_{u,v}\},\{\boldsymbol{s}_{u,v}\}\mid\{\boldsymbol{\psi}_g\},\boldsymbol{\psi},\{\boldsymbol{\mu}_{u,v}\},\{\boldsymbol{\mu}_{u}\},\{\boldsymbol{l}_{u,v}\})$. Since $m_{u,v}$ and $\boldsymbol{s}_{u,v}$ are conditionally independent given the hidden variables, we can factor the probability and calculated each. For $\{m_{u,v}\}$,
\begin{equation}\label{equa12}
    \begin{split}
        p(\{m_{u,v}\}\mid \{\boldsymbol{\psi}_g\},\boldsymbol{\psi},\{\boldsymbol{\mu}_{u,v}\},\{\boldsymbol{\mu}_{u}\},\{\boldsymbol{l}_{u,v}\})&=\log \left(\prod_u\prod_vp(m_{u,v}\mid \boldsymbol{\psi}_{\textit{g}(u)},\boldsymbol{\psi},\boldsymbol{\mu}_{u,v},\boldsymbol{\mu}_u,\boldsymbol{l}_{u,v})\right)\\ &=\sum_u\sum_v\log p(m_{u,v}\mid \boldsymbol{\psi}_{\textit{g}(u)},\boldsymbol{\psi},\boldsymbol{\mu}_{u,v},\boldsymbol{\mu}_u,\boldsymbol{l}_{u,v})
    \end{split}
\end{equation}

For $\{\boldsymbol{s}_{u,v}\}$,
\begin{equation}\label{13}
    \begin{split}
        p(\{\boldsymbol{s}_{u,v}\}\mid\{\boldsymbol{\psi}_g\},\boldsymbol{\psi},\{\boldsymbol{\mu}_{u,v}\},\{\boldsymbol{\mu}_{u}\},\{\boldsymbol{l}_{u,v}\})&=\log \left(\prod_u\prod_vp(\boldsymbol{s}_{u,v}\mid \boldsymbol{\psi}_{\textit{g}(u)},\boldsymbol{\psi},\boldsymbol{\mu}_{u,v},\boldsymbol{\mu}_u,\boldsymbol{l}_{u,v})\right)\\ &=\sum_u\sum_v\log p(\boldsymbol{s}_{u,v}\mid \boldsymbol{\psi}_{\textit{g}(u)},\boldsymbol{\psi},\boldsymbol{\mu}_{u,v},\boldsymbol{\mu}_u,\boldsymbol{l}_{u,v})
    \end{split}
\end{equation}

\subsubsection{Parameter learning}After obtaining the value of ELBO, we can update parameters to maximize ELBO. Three parts of the parameters need to be updated: 1) $\boldsymbol{\psi}$, 2) ${\boldsymbol{\psi}_g}$, 3) $\boldsymbol{\theta}_1$ and $\boldsymbol{\theta}_2$.

\noindent \textit{a) Learning} $\boldsymbol{\psi}$

As mentioned above, $\boldsymbol{\psi}$ is the BNN parameter. Holding other parameters fixed and removing the irrelevant items, Equation (\ref{equa3}) can be simplified as:
\begin{equation}\label{equa14}
    \mathrm{ELBO}_1=\mathbb{E}[\log p(\{m_{u,v}\}\mid \boldsymbol{\psi})]-\mathrm{KL}(q(\boldsymbol{\psi})\parallel p(\boldsymbol{\psi}))
\end{equation}
This is equivalent to learning BNN parameters that are supposed to perform well on maximizing the prediction of $m_{u,v}$ while also regulated by the prior distribution (i.e., the second term). Since $m_{u,v}=\boldsymbol{l}_{u,v}\cdot \boldsymbol{o}_v^T+\boldsymbol{r}_v\cdot \boldsymbol{r}_u^T$, maximizing the first term is equivalent to making the $\boldsymbol{l}_{u,v}$ close to $\boldsymbol{o}_{v}$ as much as possible. To this end, we compute their KL divergence to minimize the KL divergence $\mathrm{KL}(\boldsymbol{o}_v\parallel \boldsymbol{l}_{u,v})$. Maximizing the second term, $-\mathrm{KL}(q(\boldsymbol{\psi})\parallel p(\boldsymbol{\psi}))$, is equivalent to minimizing $\mathrm{KL}(q(\boldsymbol{\psi})\parallel p(\boldsymbol{\psi}))$. Hence, maximizing the $\mathrm{ELBO}_1$ is equivalent to minimizing the $-\mathrm{ELBO}_1$, which is called the cost function and denoted as:
\begin{equation}\label{equa15}
     \mathcal{L}_1=\mathbb{E}[\mathrm{KL}(\boldsymbol{o}_v\parallel \boldsymbol{l}_{u,v})]+\alpha \mathrm{KL}(q(\boldsymbol{\psi})\parallel p(\boldsymbol{\psi}))
\end{equation}
where $\alpha$ is a hyperparameter that represents the trade-off between the generation of user-preferred music mood and the weight prior constraint.

The first term in equation (\ref{equa15}) can be calculated as:
\begin{equation}\label{equa16}
    \mathbb{E}[\{\mathrm{KL}(\boldsymbol{o}_v\parallel \boldsymbol{l}_{u,v})\}]=\frac{1}{N}\sum_{n=1}^N\sum_{i=1}^d\left(o_{v,i}\frac{\log o_{v,i}}{l_{u,v,i}}\right)
\end{equation}
where $N$ is the data size, $d$ is the dimensional size of $\boldsymbol{o}_v$.

For the second term in equation (\ref{equa15}), since $q(\boldsymbol{\psi})=\mathcal{N}(\boldsymbol{\mu}_{\boldsymbol{\psi}},\boldsymbol{\sigma}_{\boldsymbol{\psi}})$ and $p(\boldsymbol{\psi})=\mathcal{N}(\boldsymbol{0},\boldsymbol{1})$, it can be calculated as,
\begin{equation}\label{equa17}
    \begin{split}
        \mathrm{KL}(q(\boldsymbol{\psi})\parallel p(\boldsymbol{\psi}))&=\sum_{\psi\in \boldsymbol{\psi}}\mathrm{KL}\left(q(\psi)\parallel p(\psi)\right)\\ &=\frac{1}{2}\sum_{\psi\in \boldsymbol{\psi}}\left(\mu_\psi^2+\sigma_\psi^2-\log (\sigma_\psi^2)-1\right)
    \end{split}
\end{equation}

Suggested by prior studies \cite{Charles2015}, we draw a sample of weights (denoted as $\widetilde{\boldsymbol{\psi}}$) from the distribution, i.e., $\widetilde{\boldsymbol{\psi}}\sim \mathcal{N}(\boldsymbol{\mu}_{\boldsymbol{\psi}},\boldsymbol{\sigma}_{\boldsymbol{\psi}})$. Since the sampling process is not derivable, we introduce the re-parameterization trick. We first sample the random variable $\boldsymbol{\epsilon}$ from the standard Gaussian, i.e., $\boldsymbol{\epsilon}\sim \mathcal{N}(\boldsymbol{0},\boldsymbol{1})$, and then get $\widetilde{\boldsymbol{\psi}}$ by shifting $\boldsymbol{\epsilon}$ by $\boldsymbol{\mu}_{\boldsymbol{\psi}}$ and scaling it by $\boldsymbol{\sigma}_{\boldsymbol{\psi}}$, i.e., $\widetilde{\boldsymbol{\psi}}=\boldsymbol{\mu}_{\boldsymbol{\psi}}+\boldsymbol{\sigma}_{\boldsymbol{\psi}}\circ \boldsymbol{\epsilon}$, where $\circ$ denotes the point-wise multiplication operation. Then, based on $\widetilde{\boldsymbol{\psi}}$, we can predict $\boldsymbol{l}_{u,v}$ and calculate the cost function to update the variational parameters $\boldsymbol{\mu}_{\boldsymbol{\psi}}$ and $\boldsymbol{\sigma}_{\boldsymbol{\psi}}$ with gradient descent methods.

\noindent \textit{b) Learning} $\boldsymbol{\psi}_g$

The parameter learning process is similar to the above except that the prior distribution is the learned $\boldsymbol{\psi}$. Formally, the KL divergence becomes $\mathrm{KL}(q(\boldsymbol{\psi}_g\mid \boldsymbol{\psi})\parallel p(\boldsymbol{\psi}_g))$ where $p(\boldsymbol{\psi}_g)$ is the Gaussian distribution with $\boldsymbol{\psi}$ as the expectation. Similarly, the simplified version of $\mathrm{ELBO}(q)$ is:
\begin{equation}\label{equa18}
    \mathrm{ELBO}_2=\mathbb{E}_q[\log p(\{m_{u,v}\}\mid \boldsymbol{\psi}_g)]-\mathrm{KL}\left(q(\boldsymbol{\psi}_g\mid \boldsymbol{\psi})\parallel p(\boldsymbol{\psi}_g)\right)
\end{equation}

For $q(\boldsymbol{\psi}_g\mid \boldsymbol{\psi})$, we treat its fine-tuning as the inference process. We input $\boldsymbol{\psi}$ as an initialization to obtain $\boldsymbol{\psi}_g$ via fine-tuning, i.e., the distribution of $\boldsymbol{\psi}_g$ is first initialized as $\mathcal{N}(\boldsymbol{\mu}_{\boldsymbol{\psi}},\boldsymbol{\sigma}_{\boldsymbol{\psi}})$ and then updated to maximize the $\mathrm{ELBO}_2$ on data of the $g$-th group. The output is the posterior distribution of $\boldsymbol{\psi}_g$, which is also a Gaussian distribution, denoted as $\mathcal{N}(\boldsymbol{\mu}_{\boldsymbol{\psi}_g},\boldsymbol{\sigma}_{\boldsymbol{\psi}_g})$. Similarly, maximizing the first term is equivalent to minimizing the KL divergence between $\boldsymbol{o}_v$ and $\boldsymbol{l}_{u,v}$ for users in the $g$-th group, denoted as $\mathrm{KL}(\boldsymbol{o}_v\parallel \boldsymbol{l}_{u,v})$. Maximizing $\mathrm{ELBO}_2$ is equivalent to minimizing $-\mathrm{ELBO}_2$, denoted as,
\begin{equation}\label{equa19}
    \mathcal{L}_2=\mathbb{E}_q[\{\mathrm{KL}(\boldsymbol{o}_v\parallel \boldsymbol{l}_{u,v})\}]+\alpha\mathrm{KL}\left(q(\boldsymbol{\psi}_g\mid \boldsymbol{\psi})\parallel q(\boldsymbol{\psi})\right)
\end{equation}

Similarly, the first term and the second term can be calculated as:
\begin{equation}\label{equa20}
    \mathbb{E}_q[\{\mathrm{KL}(\boldsymbol{o}_v\parallel \boldsymbol{l}_{u,v})\}]=\frac{1}{N_g}\sum_{n=1}^{N_g}\sum_{i=1}^{d}\left(o_{v,i}\frac{\log o_{u,v}}{l_{u,v,i}}\right)
\end{equation}
\begin{equation}\label{equa21}
    \mathrm{KL}\left(q(\boldsymbol{\psi}_g\mid \boldsymbol{\psi})\parallel q(\boldsymbol{\psi})\right)=\frac{1}{2}\sum_{\psi_g\in \boldsymbol{\psi}_g}\left(\log \frac{\sigma_\psi^2}{\sigma_{\psi_g}^2}+\frac{\sigma_{\psi_g}^2}{\sigma_\psi^2}+\frac{(\mu_{\psi_g}-\mu_\psi)^2}{\sigma_\psi^2}-1\right)
\end{equation}
where $N_g$ is the data size in the $g$-th group. With a similar sampling process, the parameters are also updated with gradient descent.

\noindent \textit{c) Learning parameters} $\boldsymbol{\theta}_1$ \textit{and} $\boldsymbol{\theta}_2$

As mentioned above, we adopt inference networks parametrized by $\boldsymbol{\theta}_1$ and $\boldsymbol{\theta}_2$ to infer the personalized prior LED and the posterior LED of user self-report emotion based on the observed $\boldsymbol{r}_u$ and $\boldsymbol{s}_{u,v}$. Holding other parameters fixed and removing the irrelevant items, $\mathrm{ELBO}(q)$ is simplified as:
\begin{equation}\label{equa22}
    \mathrm{ELBO}_3=\mathbb{E}_q[\log p(\{\boldsymbol{s}_{u,v}\}\mid \{\boldsymbol{\mu}_{u,v}\},\{\boldsymbol{\mu}_u\})]-\mathrm{KL}\left(q(\{\boldsymbol{\mu}_{u,v}\},\{\boldsymbol{\mu}_u\})\parallel p(\{\boldsymbol{\mu}_{u,v}\},\{\boldsymbol{\mu}_u\})\right)
\end{equation}

For the first term $\mathbb{E}_q[\log p(\{\boldsymbol{s}_{u,v}\}\mid \{\boldsymbol{\mu}_{u,v}\},\{\boldsymbol{\mu}_u\})]$, we introduce a reconstruction network based on the posterior distribution in Equation (\ref{equa11}), and then draw a sample of $\boldsymbol{s}_{u,v}^{'}$ from the reconstruction network. Formally, assuming the parameters of the reconstruction network is $\boldsymbol{\phi}_2$, and then $\boldsymbol{s}_{u,v}^{'}$ is obtained by $\boldsymbol{s}_{u,v}^{'}=NN_{\boldsymbol{\phi}_2}(\boldsymbol{z}_{u,v})$, $\boldsymbol{z}_{u,v}\sim \mathcal{N}(\boldsymbol{\mu}_{u,v},\boldsymbol{\sigma}_2)$. Then, maximizing the $\mathbb{E}_q[\log p(\{\boldsymbol{s}_{u,v}\}\mid \{\boldsymbol{\mu}_{u,v}\},\{\boldsymbol{\mu}_u\})]$ equals to minimizing the mean square error (MSE) between the ground truth $\boldsymbol{s}_{u,v}$ and the reconstructed $\boldsymbol{s}_{u,v}^{'}$. We denote the MSE as follows:
\begin{equation}\label{equa23}
    \mathcal{L}_\mathrm{MSE1}=\mathbb{E}_q\left[\mathrm{MSE}(\boldsymbol{s}_{u,v},\boldsymbol{s}_{u,v}^{'})\right]=\frac{1}{N}\sum_{n=1}^{N}\sum_{j=1}{J}(s_{u,v,j}-s_{u,v,j}^{'})
\end{equation}
where $J$ is the dimensional size of $\boldsymbol{s}_{u,v}$; $s_{u,v,j}$ and $s_{u,v,j}^{'}$ are the $j$-th values of $\boldsymbol{s}_{u,v}$ and $\boldsymbol{s}_{u,v}^{'}$, respectively.

The second term $\mathrm{KL}(q(\{\boldsymbol{\mu}_{u,v}\},\{\boldsymbol{\mu}_u\})\parallel p(\{\boldsymbol{\mu}_{u,v}\},\{\boldsymbol{\mu}_u\}))$ is denoted as $\mathcal{L}_{\mathrm{KL}}$ and can be transformed to (details in Appendix \ref{appA13}):
\begin{equation}\label{equa24}
    \begin{split}
        \mathcal{L}_{\mathrm{KL}}=&\frac{1}{2}\sum_{\boldsymbol{\mu}_u\in \{\boldsymbol{\mu}_u\}}\sum_{j=1}^{J}\left(\mu_{u,j}^2+\sigma_{1,j}^2-\log (\sigma_{1,j}^2)-1\right)\\
        &+\frac{1}{2}\sum_{\boldsymbol{\mu}_{u,v}\in \{\boldsymbol{\mu}_{u,v}\}}\sum_{j=1}^{J}\left(\log \frac{\sigma_{1,j}^2}{\sigma_{2,j}^2}+\frac{\sigma_{2,j}^2}{\sigma_{1,j}^2}+\frac{(\mu_{u,v,j}-\mu_{u,j})^2}{\sigma_{1,j}^2}-1\right)
    \end{split}
\end{equation}

Since $\boldsymbol{\mu}_{u,v}$ is also hidden and its value is inferred by the neural network that takes $\boldsymbol{\mu}_u$ as its input (see above), we hope $\boldsymbol{\mu}_u$ to be high-quality (e.g., keeping rich information about $\boldsymbol{\mathcal{S}}_u$ as much as possible). Hence, we introduce a reconstruction network parameterized by $\boldsymbol{\phi}_1$ that takes $\boldsymbol{\mu}_u$ as input and hope it can reconstruct $\boldsymbol{r}_u$, which contains the user’s personalized information. Formally, we denote the reconstructed sample as $\boldsymbol{r}_u^{'}$, then, $\boldsymbol{r}_u^{'}=NN_{\boldsymbol{\phi}_1}(\boldsymbol{z}_u)$, $\boldsymbol{z}_u\sim \mathcal{N}(\boldsymbol{\mu}_u,\boldsymbol{\sigma}_1)$. We minimize the MSE between $\boldsymbol{r}_u^{'}$ and $\boldsymbol{r}_u$ through the following,
\begin{equation}\label{equa25}
    \mathcal{L}_{\mathrm{MSE2}}=\mathbb{E}_q(\mathrm{MSE}[\boldsymbol{r}_u,\boldsymbol{r}_u^{'})]=\frac{1}{U}\sum_{u=1}^U\sum_{k=1}^K(r_{u,k}-r_{u,k}^{'})^2
\end{equation}
where $K$ is the dimensional size of $\boldsymbol{r}_u$. $r_{u,k}$ and $r_{u,k}^{'}$ are the $k$-th values of $\boldsymbol{r}_u$ and $\boldsymbol{r}_u^{'}$, respectively.

We update the parameters including $\boldsymbol{\theta}_1$, $\boldsymbol{\theta}_2$, $\boldsymbol{\phi}_1$, and $\boldsymbol{\phi}_2$ by minimizing the sum of $\mathcal{L}_\mathrm{MSE1}$, $\mathcal{L}_\mathrm{MSE2}$ and $\mathcal{L}_\mathrm{KL}$ with gradient descent.

\noindent \textit{d) Iterative learning}

In addition to maximizing the likelihood of the observation whose lower bound is given by Equation (\ref{equa3}), we also adopt the Bayesian Personalized Ranking (BPR) loss \cite{Steffen2009} to train our model. BPR loss is commonly used for recommendation tasks, assuming that users prefer an item (music in our case) that they have chosen over an item that they have not. Formally, for a user $u\in \mathbb{U}$ chooses a music $v\in \mathbb{V}$ under emotion $e\in \vmathbb{e}$. We randomly sample a music $\neg v\in \mathbb{V}\setminus v$ that $u$ has not chosen, where $\mathbb{V}\setminus v$ represents the set of all music except $v$. We get a pairwise data record $(u,e,v,\neg v)$ and $u$’s preference score $m_{u,\neg v}=\boldsymbol{l}_{u,\neg v}\cdot \boldsymbol{o}_{\neg v}^T+\boldsymbol{r}_{\neg v}\cdot \boldsymbol{r}_u^T$ for music $\neg v$. As aforementioned, user $u$ prefers music $v$ over $\neg v$, i.e., $m_{u,v}$ is expected to be higher than $m_{u,\neg v}$. Thus, we have the cost function for recommendation, denoted as $\mathcal{L}_\mathrm{rec}$:
\begin{equation}\label{equa26}
    \mathcal{L}_\mathrm{rec}=\frac{1}{\mid{\mathcal{D}_S}\mid}\sum_{(u,e,v,\neg v)\in \mathcal{D}_S}-\ln \left(\frac{1}{1+\mathrm{e}^{-(m_{u,v}-m_{u,\neg v})}}\right)
\end{equation}
where $\mathcal{D}_S$ is the data set ${(u,e,v,\neg v)}$ constructed by random sampling, and $\mid{\mathcal{D}_S}\mid$ is the size of $\mathcal{D}_S$.

We combine Equations (\ref{equa15}), (\ref{equa19}), (\ref{equa23}), (\ref{equa24}), (\ref{equa26}), and (\ref{equa26}) to get the final cost function, denoted as $\mathcal{L}_3$:
\begin{equation}\label{equa27}
    \mathcal{L}_3=\mathcal{L}_\mathrm{rec}+\lambda_1\mathcal{L}_\mathrm{KL1}+\lambda_2\mathcal{L}_\mathrm{KL2}+\lambda_3\mathcal{L}_\mathrm{MSE2}+\lambda_4\mathcal{L}_\mathrm{MSE1}+\lambda_5\mathcal{L}_1+\lambda_6\mathcal{L}_2
\end{equation}
where $\lambda_1$, $\lambda_2$, $\lambda_3$, $\lambda_4$, $\lambda_5$, and $\lambda_6$ are hyperparameters. $\mathcal{L}_\mathrm{KL1}$ and $\mathcal{L}_\mathrm{KL2}$ are the first and second terms in $\mathcal{L}_\mathrm{KL}$, respectively. By minimizing the cost function $\mathcal{L}_3$, we can update the parameters of $\boldsymbol{\psi}$, $\boldsymbol{\psi}_g$, $\boldsymbol{\theta}_1$, $\boldsymbol{\theta}_2$, $\boldsymbol{\phi}_1$, $\boldsymbol{\phi}_2$, $\boldsymbol{r}_u$, and $\boldsymbol{r}_v$ by backpropagation.

Since there are a large number of parameters to learn, considering the efficiency and stability of model training, we combine a two-phase approach with the Expectation-Maximization (EM) algorithm \cite{Arthur1977} to learn these parameters. In the first phase, we learn the user music mood preference prediction models $\boldsymbol{\psi}$ and $\boldsymbol{\psi}_g$. In the second phase, we learn the parameters of $\boldsymbol{\theta}_1$, $\boldsymbol{\theta}_2$, $\boldsymbol{\phi}_1$, $\boldsymbol{\phi}_2$, $\boldsymbol{r}_u$, and $\boldsymbol{r}_v$ following the EM algorithm. In the E-step, we fix the inference networks and obtain the hidden variables, including $\boldsymbol{\mu}_u$, $\boldsymbol{\mu}_{u,v}$, and $\boldsymbol{l}_{u,v}$. In the M-step, we fix the generative networks and compute the cost function $\mathcal{L}_3$, updating the parameters of inference networks by maximizing $-\mathcal{L}_3$. We iterate the E- and M-steps until convergence. Algorithm \ref{algorithm2} provides the learning process.

\begin{algorithm}[h]
    \caption{Learning Process of HDBN}
    \label{algorithm2}
    \begin{flushleft}
    \textbf{Input}: User set $\mathbb{U}$, music set $\mathbb{V}$, user emotion set $\vmathbb{e}$, user-music interaction history $\{(u,e,v)\}$, and music mood $\{\boldsymbol{o}_v\}$. Hyper-parameters: $\alpha$, $\lambda_1$, $\lambda_2$, $\lambda_3$, $\lambda_4$, $\lambda_5$, and $\lambda_6$;\\
    \textbf{Output}: $\boldsymbol{\psi}$, $\{\boldsymbol{\psi}_g\}_{g=1}^G$, $\boldsymbol{\theta}_1$, $\boldsymbol{\phi}_1$, $\boldsymbol{\theta}_2$, $\boldsymbol{\phi}_2$, $\{\boldsymbol{r}_u\}$, and $\{\boldsymbol{r}_v\}$;\\
    \textbf{Phase $\mathrm{I}$}:\\
    \hspace{5mm}Initialize $\boldsymbol{\psi}$;\\
    \hspace{5mm} \textcolor{gray}{// pre-train the music mood preference prediction model on the full dataset}\\
    \hspace{5mm}\textbf{for} epoch in range(Epochs) \textbf{do}\\
    \hspace{10mm}\textbf{for} each batch input in $\{(u,e,v)\}$ \textbf{do}\\
    \hspace{15mm}Compute the preferred music mood: $\boldsymbol{l}_{u,v}=\mathrm{softmax}(\mathcal{F}_{\boldsymbol{\psi}}(\boldsymbol{s}_{u,v}))$;\\
    \hspace{15mm}Compute $\mathcal{L}_1$ and update $\boldsymbol{\psi}$ by gradient descent;\\
    \hspace{5mm} \textcolor{gray}{//fine-tune the music mood preference prediction model on the data of each group}\\
    \hspace{5mm}\textbf{for} g in range($G$) \textbf{do}\\
    \hspace{10mm}Initialize $\boldsymbol{\psi}_g$ by $\boldsymbol{\psi}$;\\
    \hspace{10mm}\textbf{for} epoch in range(Epochs) \textbf{do}\\
    \hspace{15mm}\textbf{for} each batch input in $\{(u,e,v)\}_g$ \textbf{do}\\
    \hspace{20mm}Compute the user-preferred music mood: $\boldsymbol{l}_{u,v}=\mathrm{softmax}(\mathcal{F}_{\boldsymbol{\psi}_g}(\boldsymbol{s}_{u,v})$;\\
    \hspace{20mm}Compute $\mathcal{L}_2$ and update $\boldsymbol{\psi}_g$ by gradient descent;\\
    \textbf{Phase $\mathrm{II}$}:\\
    \hspace{5mm}Initialize $\boldsymbol{\theta}_1$, $\boldsymbol{\phi}_1$, $\boldsymbol{\theta}_2$, $\boldsymbol{\phi}_2$, $\{\boldsymbol{r}_u\}$, and $\{\boldsymbol{r}_v\}$;\\
    \hspace{5mm}\textbf{while} \textit{not convergence} \textbf{do}\\
    \hspace{10mm}\textbf{for all} $u\in \mathbb{U}$ \textbf{do}\\
    \hspace{15mm}\textbf{for all} $v \in \{(u,e,v)\}_g$ \textbf{do}\\
    \hspace{20mm} \textcolor{gray}{// Sample negative music tracks from a collection of music $\mathbb{V}\setminus v$ that removes user-selected music}\\
    \hspace{20mm}Conduct negative sampling to build the training set $\mathcal{D}_{train}$;\\
    \hspace{10mm}\textbf{E-step}:\\
    \hspace{15mm}\textbf{for} each batch input $\{(u,e,v,\neg v)\}\in \mathcal{D}_{train}$ \textbf{do}\\
    \hspace{20mm}Generate the user’s prior LED $q(\boldsymbol{\mu}_u\mid \boldsymbol{\mu})=\mathcal{N}(\boldsymbol{\mu}_u,\boldsymbol{\sigma}_1)$\\
    \hspace{20mm}Generate the user’s posterior LED $q(\boldsymbol{\mu}_{u,v}\mid \boldsymbol{\mu}_u)=\mathcal{N}(\boldsymbol{\mu}_{u,v},\boldsymbol{\sigma}_2)$;\\
    \hspace{20mm}Draw user latent emotion $\boldsymbol{z}_{u,v}~\mathcal{N}(\boldsymbol{\mu}_{u,v},\boldsymbol{\sigma}_2)$;\\
    \hspace{20mm}Obtain user-preferred music mood $\boldsymbol{l}_{u,v}(\boldsymbol{l}_{u,\neg v})=\mathrm{softmax}(\mathcal{F}_{\boldsymbol{\psi}_g}(\boldsymbol{z}_{u,v}))$;\\
    \hspace{10mm}\textbf{M-step}:\\
    \hspace{15mm}Calculate $m_{u,v}=\boldsymbol{l}_{u,v}\cdot \boldsymbol{o}_{v}^T+\boldsymbol{r}_{v}\cdot \boldsymbol{r}_u^T$ and $m_{u,\neg v}=\boldsymbol{l}_{u,\neg v}\cdot \boldsymbol{o}_{\neg v}^T+\boldsymbol{r}_{\neg v}\cdot \boldsymbol{r}_u^T$;\\
    \hspace{15mm}Calculate $\mathcal{L}_3$ and update $\boldsymbol{\theta}_1$, $\boldsymbol{\phi}_1$, $\boldsymbol{\theta}_2$, $\boldsymbol{\phi}_2$, $\{\boldsymbol{r}_u\}$, and $\{\boldsymbol{r}_v\}$ by gradient descent; \textcolor{gray}{//$\lambda_5=0$ and $\lambda_6=0$}
    \end{flushleft}
\end{algorithm}

\section{EXPERIMENTS AND RESULT ANALYSIS}
In this section, we conduct extensive experiments to evaluate our proposed method. We are primarily concerned with answering the following research questions:

\begin{itemize}
    \item \textbf{RQ1}. What is the performance of our method compared to existing music recommendation methods?
    \item \textbf{RQ2}. Does the incorporation of emotion heterogeneity across users and within a user and music mood preference heterogeneity across users and within a user improve recommendation performance?
    \item \textbf{RQ3}. How do key hyperparameters influence the performance of our method?
    \item \textbf{RQ4}. Can our method learn meaningful representations of emotion distributions?
    \item \textbf{RQ5}. How do real users evaluate the music recommendation list provided by our method?
\end{itemize}

\subsection{Dataset}
The data used to evaluate the effectiveness of our method should meet the following requirement: users’ emotions should be acquired before they listen to music such that the selections of music are influenced by users’ emotions, rather than music influencing users’ emotions. However, at present, there is no publicly available dataset that conforms to the above requirement. Hence, we constructed a dataset named EmoMusicLJ to evaluate our method and support future related research.

\begin{figure}[h]
  \centering
  \includegraphics[width=0.5\linewidth]{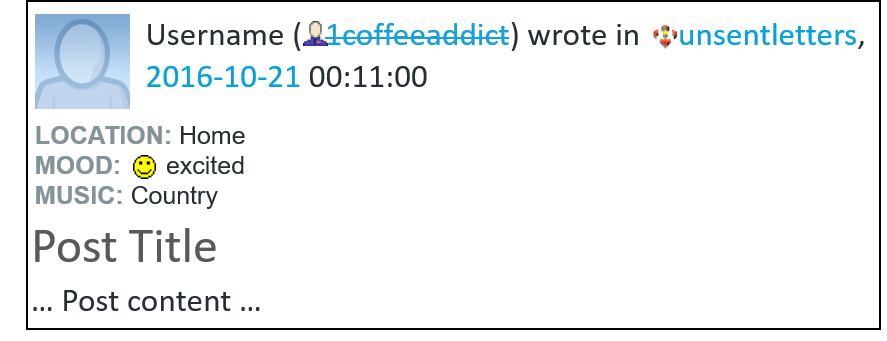}
  \caption{An example of a post on LiveJournal. The username and post content in this example have been hidden for privacy reasons.}
  \Description{Four limitations of existing emotion-aware music recommendation and the corresponding solutions of our method.}
  \label{livejournal}
\end{figure}

The construction process of EmoMusicLJ is as follows. \textbf{Step (1)}: User filtering. We take the LiveJournal two-million post (LJ2M) dataset collected from LiveJournal\footnote{https://www.livejournal.com/} by Liu et al. \cite{Liu2014} as the basis for constructing EmoMusicLJ. We also crawled some recent data records from LiveJournal. LiveJournal allows users to share their experiences and emotions through emotion tags\footnote{https://www.livejournal.com/moodlist.bml} and then to provide the music they like at that moment. Figure \ref{livejournal} shows an example of a post on LiveJournal. From LJ2M, we extracted valuable information, including user IDs, user self-reported emotions, and the titles, artists, and corresponding 7Digital IDs of music tracks. To ensure an adequate amount of data for behavior modeling, we retained users who posted at least ten entries. As a result, we obtained 12,557 users. \textbf{Step (2)}: Music alignment. Since some music IDs correspond to multiple titles, we split the different titles into separate music tracks and assigned them new IDs. Then, we filtered out the tracks that have been listened to less than three times. \textbf{Step (3)}: Music meta-information supplement. We collected meta-information for music, including genre, year of release, and audio files. For genre and release year, we crawled them from the well-known commercial web service Musicovery based on the music title and artist name. For audio files, we retrieved candidate files from 7Digital based on the music title and artist name. Then we matched the IDs of the candidate files to the target music. If a match was found, we saved the corresponding audio file. Otherwise, we saved the first candidate audio file with the same title and artist as the target music. \textbf{Step (4)}: Music mood labeling. To get music mood, we trained a Bi-LSTM-based music mood recognition model on the Emotify dataset \cite{Anna2016}. Emotify comprises 400 one-minute-long music pieces. In total, 1,778 annotators were asked to assign up to three of nine mood tags (including amazement, solemnity, tenderness, nostalgia, calmness, power, joyful activation, tension, and sadness) to a music track. On average, each music track received approximately 40 annotations. The music mood labeling is a label distribution learning (LDL) task, where the learning objective is to minimize the KL divergence between the ground truth and the predicted music mood distributions \cite{Jia2023, Xu2021}. During model training and mood annotation, we utilized 35 audio features related to five music aspects: energy, rhythm, temporal, spectrum, and melody \cite{Yang2011}. The performance of the Bi-LSTM model was measured by the KL divergence, cosine similarity (CR), and the mean squared error (MSE) between the ground truth and predicted music mood distributions, with the final losses of KL=0.2579, CR=0.8564, and MSE=0.0523 upon model convergence.

We also construct a dataset named EmoMusicLJ-small from EmoMusicLJ with relative low level of data sparsity. Table \ref{dataCharacter} provides the statistical characteristics of EmoMusicLJ and EmoMusicLJ-small. Following a previous study \cite{Zheng2020}, EmoMusicLJ was randomly split into training, validation, and test sets in a ratio of 8:1:1.

\begin{table}[htbp]
        \centering
        \captionof{table}{The statistical Characteristics of EmoMusicLJ and EmoMusicLJ-small}
        \label{dataCharacter}
        \renewcommand{\arraystretch}{1.25}
        \resizebox{\textwidth}{!}{
            \begin{tabular}{lp{0.1\textwidth}p{0.1\textwidth}p{0.15\textwidth}p{0.1\textwidth}p{0.1\textwidth}p{0.1\textwidth}p{0.1\textwidth}p{0.1\textwidth}} \toprule
             Dataset & \#Users &  \#Music & \#User emotion & \#Genre & \#Year & \#Artist & \#Interaction & Sparsity\\ \midrule
            EmoMUsicLJ
            &\makecell[l]{15,233}
            &\makecell[l]{6,180}
            &\makecell[l]{132}
            &\makecell[l]{18}
            &\makecell[l]{57}
            &\makecell[l]{1,349}
            &\makecell[l]{157,472}
            &\makecell[l]{99.84\%}\\
            EmoMusicLJ-small
            &\makecell[l]{3,473}
            &\makecell[l]{217}
            &\makecell[l]{132}
            &\makecell[l]{11}
            &\makecell[l]{9}
            &\makecell[l]{107}
            &\makecell[l]{21,990}
            &\makecell[l]{97.08\%}\\
            \bottomrule
        \end{tabular}}
    \end{table}

\subsection{Baselines}
We compare our method against three types of baseline methods: 1) emotion-free methods, 2) emotion-matching methods, and 3) emotion-as-a-feature methods.
\begin{enumerate}
    \item [1)] Emotion-free methods: These methods are widely employed in recommendation systems and can be adopted for music recommendation, but they do not consider emotional information.
        \begin{itemize}
            \item \textbf{Random} and \textbf{Pop}: In the Random method, music recommendations are made by randomly selecting tracks from the database. Pop recommends music based on its overall popularity.
            \item \textbf{UCF} \cite{Charu2016}: The user-based collaborative filtering (UCF) method generates recommendations for a target user by considering items interacted with by users who share similar music preferences to the target user.
            \item \textbf{ICF} \cite{Charu2016}: The item-based collaborative filtering (ICF) generates recommendations by assessing the similarity between the target item and the items in the target user’s interaction history.
            \item \textbf{NCF} and \textbf{NMF} \cite{He2017}: Neural collaborative filtering (NCF) leverages neural networks to capture the implicit interactions between users and items, replacing the traditional inner product of user and item vectors. Neural matrix factorization (NMF) integrates NCF with generalized matrix factorization to learn more adaptable and nonlinear interactions between user and item representations.
            \item \textbf{MF-BPR} \cite{Steffen2009}: BPR-MF is a traditional and renowned method for modeling item recommendation based on implicit user-item interactions. It learns a matrix factorization model using Bayesian personalized ranking.
        \end{itemize}
    \item [2)] Emotion matching methods: These methods recommend music by computing the similarity score between users’ emotions or mood similarity between music.
        \begin{itemize}
            \item \textbf{MoodSim} \cite{Priyanka2023}: MoodSim trains a neural network to predict a user’s music mood preference based on the user’s emotion in her/his history behavior data. Music mood similarity scores between the candidate tracks and the predicted user preferred music mood are computed to support the recommendation.
            \item \textbf{UCFE} \cite{Deng2015}: User-based collaborative filtering with emotion (UCFE) calculates user similarity by considering users’ emotional contexts. (See Appendix \ref{appA21} for details)
            \item \textbf{ICFE} \cite{Deng2015}: Similar to UCFE, item-based collaborative filtering with emotion (ICFE) computes the similarity between music pieces by considering users’ emotional contexts. (See Appendix \ref{appA22} for details)
            \item \textbf{UCF+E}: Different from UCFE, which considers only the user emotions when calculating user similarity, in combination with traditional UCF, UCF+E incorporates user emotions as auxiliary information for computing user similarity.
            \item \textbf{ICF+E}: ICF+E combines user emotions as auxiliary information with the traditional ICF when computing the similarity between music pieces.
        \end{itemize}
    \item [3)] Emotion-as-a-feature methods: These methods are deep learning-based techniques that incorporate user emotion and music mood as features.
        \begin{itemize}
            \item \textbf{PEIA} \cite{Shen2020}: The personality and emotion integrated attentive (PEIA) model uses hierarchical attention to learn the influence of users’ personalities and emotions on their music preferences. Due to the lack of user personality-related data, we remove the personality component of PEIA while retaining all other modules.
            \item \textbf{Wide\&Deep} \cite{Cheng2016}: An approach for jointly training wide linear models and deep neural networks. Wide\&Deep combines the ability of generalized linear models to handle sparse features with the advantages of deep neural networks to learn generalized feature combinations.
            \item \textbf{DeepFM} \cite{Guo2017}: An approach that can simultaneously learn low-order and high-order feature interactions from the input raw features.
            \item \textbf{TagGCN} \cite{Yao2024}: User emotions are regarded as tags and used along with user-music interactions to construct user-emotion bipartite graph, emotion-music bipartite graph, and user-music bipartite graph. Graph neural networks are used to learn the relationships between users, emotion tags, and music tracks.
            \item \textbf{MEGAN} \cite{Wang2023}: User's emotion and music mood along with music content features including genre, release year, and artist are used to construct heterogeneous music graph. Multi-view enhanced graph attention network (MEGAN) separately learns user representation from user’s profile view, behavior view, and emotion view, and music representation from attribute view, interaction view, and emotion view.
        \end{itemize}
\end{enumerate}

\subsection{Evaluation metrics and experimental setting}
We adopt the following widely used metrics in recommendation systems \cite{Shen2020, He2017}: Hit Rate (HR), Precision, Normalized Discounted Cumulative Gain (NDCG), and Mean Reciprocal Rank (MRR).

Given a music $v$, its predicted rating is calculated alongside all other music that the user has not listened to. The top-$T$ recommendation list is obtained by sorting the predicted ratings in descending order and selecting the first $T$ music. Note that there is only one ground-truth music track for one record in the test set. HR@$T$ and Precision@$T$ are calculated as follows:
\begin{equation}\label{equa28}
    \mathrm{HR}@T=\begin{cases}
         1, \hspace{5mm} \mathrm{if}~v~\mathrm{in~the~top-}T\mathrm{~list} \\
         0, \hspace{5mm} \mathrm{otherwise}
    \end{cases}
\end{equation}

\begin{equation}\label{equa29}
    \mathrm{Precision}@T=\begin{cases}
         \frac{1}{T}, \hspace{5mm} \mathrm{if}~v~\mathrm{in~the~top-}T\mathrm{~list} \\
         0, \hspace{5mm} \mathrm{otherwise}
    \end{cases}
\end{equation}

While HR@$T$ and Precision@$T$ solely assess whether music $v$ is present in the top-$T$ list, NDCG@$T$ and MRR@$T$ consider the relative position of $v$ within the top-$T$ list. NDCG@$T$ and MRR@$T$ are calculated as follows:
\begin{equation}\label{equa30}
    \mathrm{NDCG}@T=\frac{2^{\mathbb{I}(v)}}{\log_2(t+1)}
\end{equation}
\begin{equation}\label{equa31}
    \mathrm{MRR}@\mathrm{T}=\begin{cases}
         \frac{1}{t}, \hspace{5mm} \mathrm{if}~v~\mathrm{in~the~top-}T\mathrm{~list} \\
         0, \hspace{5mm} \mathrm{otherwise}
    \end{cases}
\end{equation}
where $t$ denotes the position of $v$ in the top-$T$ list, $t\in\{1,2,…,T\}$. $\mathbb{I}(v)$ acts as an indicator, equaling 1 if $v$ is in the top-$T$ list, and 0 otherwise. Higher values of both HR@$T$, Precision@$T$, NDCG@$T$, and MRR@$T$ indicate better performance. We report their average values across all test samples. $T$ is set to 5, 10, 15, and 20.

\subsection{Experiment setup}
All experiments were conducted on a workstation equipped with an Intel(R) Xeon(R) CPU E5-2620 v4 @2.10 GHz and a NVIDIA Titan X graphic card, and a PC featuring an Intel(R) Core(TM) i7-7700 CPU @3.60GHz. We implemented our method based on TensorFlow 1.15.0 with Python 3.7.2. Grid search was employed to optimize hyperparameters for our approach and baseline methods. For baseline methods, if the baseline method had open-sourced its code, we used the official code and modified the input and evaluation parts to adapt it to our datasets. Otherwise, we reproduced the baseline method based on the descriptions of the method in the original paper.

For our approach, in music mood annotation experiment, we set the learning rate to 0.01, batch size to 100, hidden size of BiLSTM to 128, and dropout rate of fully connected layers to 0.2. The number of user groups was set to 50 and 10 for EmoMusicLJ and EmoMusicLJ-small, respectively. For the music mood preference prediction model, a two-layer BNN with 64 neurons in each layer was used on both datasets. The learning rate was set to 0.01 during the pre-training phase and 0.001 during the fine-tuning phase. The batch size was set to 1024 and 512 during the pre-training phase on EmoMusicLJ and EmoMusicLJ-small, respectively. The batch size was set to 64 during the fine-tuning phase on both datasets. The value of $\alpha$ was set to 1e-5 during both the pre-training and tine-tuning phases on both datasets. In the recommendation experiment, the size of the latent emotion distribution was set to 16 according to \cite{John2000}. The embedding dimensions for both users and music was set to 64 for both datasets. The number of negative samples was set to 10 and 7 for EmoMusicLJ and EmoMusicLJ-small, respectively. The batch size and learning rate were set to 512 and 0.05 on both datasets. On EmoMusicLJ, the coefficients of $\lambda_1$, $\lambda_2$, $\lambda_3$, and $\lambda_4$ were set to 0.01, 0.05, 1e-6, and 1e-4, respectively. On EmoMusicLJ-small, the coefficients of $\lambda_1$, $\lambda_2$, $\lambda_3$, and $\lambda_4$ were set to 0.005, 0.005, 5e-6, and 5e-5, respectively.

\subsection{Overall performance (RQ1)}
In this section, we outline the recommendation performances of our HDBN and baseline methods. Table \ref{overall_result1} and Table \ref{overall_result2} summarize the overall performances on metrics of HR@T, Precision@T, NDCG@T, and MRR@T on EmoMusicLJ and EmoMusicLJ-small, respectively.

    \begin{table}[htbp]
        \centering
        \captionof{table}{Overall Performance Comparison between HDBN and Baseline Methods on EmoMusicLJ}
        \label{overall_result1}
        \renewcommand{\arraystretch}{1.25}
        \resizebox{\textwidth}{!}{
            \begin{tabular}{lp{0.1\textwidth}p{0.1\textwidth}p{0.1\textwidth}p{0.1\textwidth}p{0.1\textwidth}p{0.1\textwidth}p{0.1\textwidth}p{0.1\textwidth}p{0.1\textwidth}} \toprule
             \multirow{2}*{Method type} & \multirow{2}*{Method} & \multicolumn{4}{c}{HR} &  \multicolumn{4}{c}{Precision} \\ \cmidrule(r){3-6} \cmidrule(r){7-10}
            & & @5 & @10 & @15 & @20 & @5 & @10 & @15 & @20\\ \midrule
            \multirow{7}*{Emotion-free}&Random&0.0010&0.0014&0.0026&0.0034&0.0002&0.0001&0.0002&0.0002\\
            &Pop&0.0230&0.0376&0.0529&0.0646&0.0046&0.0038&0.0035&0.0032\\
            &UCF&0.0863&0.1269&0.1526&0.1705&0.0173&0.0127&0.0102&0.0085\\
            &ICF&0.0274&0.0439&0.0565&0.0669&0.0055&0.0044&0.0038&0.0033\\
            &NCF&0.0701&0.0998&0.1196&0.1332&0.0140&0.0100&0.0080&0.0067\\
            &NMF&0.0749&0.1025&0.1185&0.1324&0.0150&0.0102&0.0079&0.0066\\
            &MF-BPR&0.1212&\underline{0.1567}&\underline{0.1786}&\underline{0.1960}&0.0242&\underline{0.0157}&\underline{0.0119}&\underline{0.0098}\\ \midrule
            \multirow{5}*{Emotion-matching}&MoodSim&0.0003&0.0012&0.0020&0.0027&0.0001&0.0001&0.0001&0.0001\\
            &UCFE&0.0319&0.0401&0.0471&0.0533&0.0064&0.0040&0.0031&0.0027\\
            &ICFE&0.0062&0.0098&0.0115&0.0150&0.0012&0.0010&0.0008&0.0008\\
            &UCF+E&0.0850&0.1219&0.1447&0.1617&0.0170&0.0122&0.0096&0.0081\\
            &ICF+E&0.0282&0.0447&0.0581&0.0679&0.0056&0.0045&0.0039&0.0034\\ \midrule
            \multirow{5}*{Emotion-as-a-feature}&PEIA&0.0835&0.1170&0.1395&0.1557&0.0167&0.0117&0.0093&0.0078\\
            &Wide\&Deep&0.0528&0.0809&0.1036&0.1243&0.0106&0.0081&0.0069&0.0062\\
            &DeepFM&0.0993&0.1300&0.1495&0.1647&0.0199&0.0130&0.0100&0.0082\\
            &TagGCN&0.1039&0.1243&0.1375&0.1457&0.0208&0.0124&0.0092&0.0073\\
            &MEGAN&\underline{0.1289}&0.1392&0.1477&0.1556&\underline{0.0258}&0.0139&0.0098&0.0078\\
            \midrule
            our&HDBN&\textbf{0.1379}&\textbf{0.1692}&\textbf{0.1899}&\textbf{0.2068}&\textbf{0.0276}&\textbf{0.0169}&\textbf{0.0127}&\textbf{0.0103}\\ \midrule
            \multicolumn{2}{c}{Imprv}&$6.98\%$&$7.98\%$&$6.33\%$&$5.51\%$&$6.98\%$&$7.64\%$&$6.72\%$&$5.10\%$\\ \midrule \midrule
            
            \multirow{2}*{Method type} & \multirow{2}*{Method} & \multicolumn{4}{c}{NDCG} &  \multicolumn{4}{c}{MRR} \\ \cmidrule(r){3-6} \cmidrule(r){7-10}
            & & @5 & @10 & @15 & @20 & @5 & @10 & @15 & @20\\ \midrule
            \multirow{7}*{Emotion-free}&Random&0.0005&0.0006&0.0010&0.0013&0.0003&0.0004&0.0006&0.0007\\
            &Pop&0.0146&0.0193&0.0233&0.0261&0.0119&0.0138&0.0150&0.0156\\
            &UCF&0.0529&0.0683&0.0751&0.0794&0.0451&0.0505&0.0525&0.0535\\
            &ICF&0.0174&0.0227&0.0260&0.0285&0.0142&0.0163&0.0173&0.0179\\
            &NCF&0.0498&0.0594&0.0647&0.0679&0.0432&0.0472&0.0487&0.0495\\
            &NMF&0.0537&0.0627&0.0669&0.0702&0.0467&0.0505&0.0517&0.0525\\
            &MF-BPR&0.0941&0.1056&0.1114&0.1155&0.0851&0.0899&0.0916&0.0926\\ \midrule
            \multirow{5}*{Emotion-matching}&MoodSim&0.0001&0.0004&0.0006&0.0008&0.0001&0.0002&0.0003&0.0003\\
            &UCFE&0.0243&0.0270&0.0288&0.0303&0.0218&0.0230&0.0235&0.0238\\
            &ICFE&0.0037&0.0048&0.0053&0.0061&0.0028&0.0033&0.0034&0.0036\\
            &UCF+E&0.0556&0.0676&0.0736&0.0777&0.0460&0.0510&0.0527&0.0537\\
            &ICF+E&0.0181&0.0233&0.0269&0.0292&0.0148&0.0169&0.0179&0.0185\\ \midrule
            \multirow{5}*{Emotion-as-a-feature}&PEIA&0.0587&0.0694&0.0755&0.0793&0.0506&0.0550&0.0568&0.0577\\
            &Wide\&Deep&0.0361&0.0451&0.0511&0.0560&0.0306&0.0343&0.0361&0.0372\\
            &DeepFM&0.0730&0.0829&0.0880&0.0916&0.0643&0.0684&0.0699&0.0707\\
            &TagGCN&0.0812&0.0878&0.0913&0.0932&0.0737&0.0764&0.0775&0.0779\\
            &MEGAN&\textbf{0.1186}&\textbf{0.1219}&\underline{0.1241}&\underline{0.1260}&\textbf{0.1152}&\textbf{0.1165}&\textbf{0.1172}&\textbf{0.1176}\\
            \midrule
            our&HDBN&0.1097&0.1198&\textbf{0.1252}&\textbf{0.1292}&\underline{0.1003}&\underline{0.1045}&\underline{0.1061}&\underline{0.1071}\\ \midrule
            \multicolumn{2}{c}{Imprv}&$-7.50\%$&$-1.72\%$&$0.89\%$&$2.54\%$&$-12.93\%$&$-10.30\%$&$-9.47\%$&$-8.93\%$\\
            \bottomrule
        \end{tabular}}
        \footnotesize{Note: The best results are highlighted in bold, and the second-best results are underlined. “Imprv” denotes the improvement of the HDBN over the best baseline method.}
    \end{table}

    \begin{table}[htbp]
        \centering
        \captionof{table}{Overall Performance Comparison between HDBN and Baseline Methods on EmoMusicLJ-small}
        \label{overall_result2}
        \renewcommand{\arraystretch}{1.25}
        \resizebox{\textwidth}{!}{
            \begin{tabular}{lp{0.1\textwidth}p{0.1\textwidth}p{0.1\textwidth}p{0.1\textwidth}p{0.1\textwidth}p{0.1\textwidth}p{0.1\textwidth}p{0.1\textwidth}p{0.1\textwidth}} \toprule
             \multirow{2}*{Method type} & \multirow{2}*{Method} & \multicolumn{4}{c}{HR} &  \multicolumn{4}{c}{Precision} \\ \cmidrule(r){3-6} \cmidrule(r){7-10}
            & & @5 & @10 & @15 & @20 & @5 & @10 & @15 & @20\\ \midrule
            \multirow{7}*{Emotion-free}&Random&0.0196&0.0514&0.0664&0.0964&0.0039&0.0051&0.0044&0.0048\\
            &Pop&0.1069&0.1792&0.2365&0.2797&0.0214&0.0179&0.0158&0.0140\\
            &UCF&0.2342&0.3347&0.3993&0.4438&0.0468&0.0335&0.0266&0.0222\\
            &ICF&0.1214&0.2005&0.2588&0.3024&0.0243&0.0214&0.0173&0.0151\\
            &NCF&0.2638&0.3060&0.3502&0.3824&0.0528&0.0306&0.0233&0.0191\\
            &NMF&0.2456&0.2788&0.3170&0.3456&0.0491&0.0279&0.0211&0.0173\\
            &MF-BPR&\underline{0.3046}&\underline{0.3706}&\underline{0.4156}&\underline{0.4570}&\underline{0.0609}&\underline{0.0370}&\underline{0.0277}&\underline{0.0228}\\ \midrule
            \multirow{5}*{Emotion-matching}&MoodSim&0.0186&0.0355&0.0532&0.0705&0.0037&0.0035&0.0035&0.0035\\
            &UCFE&0.1128&0.1669&0.2060&0.2415&0.0226&0.0167&0.0137&0.0121\\
            &ICFE&0.0277&0.0509&0.1019&0.1378&0.0055&0.0051&0.0068&0.0068\\
            &UCF+E&0.2310&0.3320&0.3938&0.4434&0.0462&0.0332&0.0263&0.0222\\
            &ICF+E&0.1169&0.1928&0.2551&0.2974&0.0234&0.0193&0.0170&0.0149\\ \midrule
            \multirow{5}*{Emotion-as-a-feature}&PEIA&0.2442&0.2874&0.3170&0.3520&0.0488&0.0287&0.0211&0.0176\\
            &Wide\&Deep&0.2588&0.3060&0.3383&0.3706&0.0518&0.0306&0.0226&0.0185\\
            &DeepFM&0.2660&0.3047&0.3556&0.3847&0.0532&0.0320&0.0237&0.0192\\
            &TagGCN&0.2756&0.3215&0.3602&0.3942&0.0551&0.0322&0.0240&0.0197\\
            &MEGAN&0.2920&0.3461&0.3893&0.4424&0.0584&0.0347&0.0260&0.0221\\
            \midrule
            our&HDBN&\textbf{0.3251}&\textbf{0.3938}&\textbf{0.4548}&\textbf{0.4975}&\textbf{0.0650}&\textbf{0.0394}&\textbf{0.0303}&\textbf{0.0249}\\ \midrule
            \multicolumn{2}{c}{Imprv}&$6.73\%$&$6.26\%$&$9.43\%$&$8.86\%$&$6.73\%$&$6.49\%$&$9.39\%$&$9.21\%$\\ \midrule \midrule
            
            \multirow{2}*{Method type} & \multirow{2}*{Method} & \multicolumn{4}{c}{NDCG} &  \multicolumn{4}{c}{MRR} \\ \cmidrule(r){3-6} \cmidrule(r){7-10}
            & & @5 & @10 & @15 & @20 & @5 & @10 & @15 & @20\\ \midrule
            \multirow{7}*{Emotion-free}&Random&0.0120&0.0231&0.0243&0.0332&0.0095&0.0147&0.0126&0.0162\\
            &Pop&0.0648&0.0880&0.1032&0.1134&0.0510&0.0605&0.0620&0.0674\\
            &UCF&0.1509&0.1836&0.2005&0.2110&0.1237&0.1372&0.1422&0.1447\\
            &ICF&0.0799&0.1051&0.1205&0.1308&0.0663&0.0765&0.0810&0.0835\\
            &NCF&0.2203&0.2338&0.2455&0.2531&0.2058&0.2113&0.2148&0.2166\\
            &NMF&0.2279&0.2385&0.2485&0.2553&0.2222&0.2265&0.2294&0.2310\\
            &MF-BPR&\underline{0.2495}&\underline{0.2709}&\underline{0.2828}&\underline{0.2925}&\underline{0.2313}&\underline{0.2402}&\underline{0.2437}&\underline{0.2460}\\ \midrule
            \multirow{5}*{Emotion-matching}&MoodSim&0.0107&0.0160&0.0207&0.0247&0.0081&0.0102&0.0116&0.0126\\
            &UCFE&0.0837&0.1010&0.1113&0.1197&0.0742&0.0812&0.0842&0.0862\\
            &ICFE&0.0170&0.0243&0.0375&0.0460&0.0135&0.0164&0.0202&0.0222\\
            &UCF+E&0.1492&0.1819&0.1920&0.2100&0.1224&0.1359&0.1407&0.1435\\
            &ICF+E&0.0775&0.1019&0.1183&0.1282&0.0646&0.0746&0.0794&0.0818\\ \midrule
            \multirow{5}*{Emotion-as-a-feature}&PEIA&0.2156&0.2294&0.2372&0.2455&0.2061&0.2117&0.2141&0.2160\\
            &Wide\&Deep&0.2341&0.2489&0.2575&0.2651&0.2260&0.2319&0.2345&0.2363\\
            &DeepFM&0.2390&0.2562&0.2657&0.2725&0.2280&0.2341&0.2372&0.2390\\
            &TagGCN&0.2361&0.2510&0.2612&0.2693&0.2232&0.2294&0.2324&0.2343\\
            &MEGAN&0.2418&0.2592&0.2706&0.2831&0.2252&0.2323&0.2357&0.2387\\
            \midrule
            our&HDBN&\textbf{0.2724}&\textbf{0.2944}&\textbf{0.3104}&\textbf{0.3205}&\textbf{0.2551}&\textbf{0.2640}&\textbf{0.2688}&\textbf{0.2712}\\ \midrule
            \multicolumn{2}{c}{Imprv}&$9.18\%$&$8.67\%$&$9.76\%$&$9.57\%$&$10.29\%$&$9.91\%$&$10.30\%$&$10.24\%$\\
            \bottomrule
        \end{tabular}}
        \footnotesize{Note: The best results are highlighted in bold, and the second-best results are underlined. “Imprv” denotes the improvement of the HDBN over the best baseline method.}
    \end{table}

Based on Table \ref{overall_result1} and Table \ref{overall_result2}, the following observations can be made:
\begin{enumerate}
    \item [(1)] HDBN achieves the best performance on most metrics on the EmoMusicLJ dataset and performs best across all metrics on the EmoMusicLJ-small dataset. Although HDBN performs worse than baseline method on some metrics in the EmoMusicLJ dataset, it still attains the second-best performances. This underscores the effectiveness of HDBN.
    \item [(2)] In comparing UCF, UCF+E, and UCFE on two datasets EmoMusicLJ and EmoMusicLJ-small, UCF demonstrates the best performance, followed by UCF+E. UCFE performs significantly worse than UCF and UCF+E. UCF+E uses user emotion as the auxiliary information while UCFE only considers user emotion similarity. This suggests that simply assuming similarity in music mood preferences among users with similar emotions may potentially impair recommendation performance. A similar trend is observed when comparing ICF, ICF+E, and ICFE. These observations underscore the importance of accounting for emotion heterogeneity and music mood preference heterogeneity in emotion-aware music recommendation, as UCF+E, UCFE, ICF+E, and ICFE fail to address these critical heterogeneity issues.
    \item [(3)] PEIA, DeepFM, tagGCN, and MEGAN exhibit superior performance compared to UCF+E on most metrics on two datasets, suggesting that deep learning-based emotion-as-a-feature methods can effectively utilize information beyond user emotion similarity. Interestingly, we can observe that deep learning-based methods, such as PEIA, Wide\&Deep, DeepFM, and tagGCN, perform worse than MF-BPR, which aligns with findings by Klingler et al. \cite{Klingler2022}. Klingler et al. noted that matrix factorization methods generally outperform neural network approaches on high-sparsity data. EmoMusicLJ and EmoMusicLJ-small have sparsity levels of 99.84\% and 97.08\%, respectively. We compare the improvement of MF-BPR over these methods on both datasets and find that as the sparsity level decrease, the relative improvement decreases as well. Table \ref{sparsity_result} shows the comparison results.
\end{enumerate}

    \begin{table}[htbp]
        \centering
        \captionof{table}{Relative Improvement of MF-BPR over PEIA, Wide\&Deep, DeepFM, and tagGCN on Different Data Sparsity Levels}
        \label{sparsity_result}
        \renewcommand{\arraystretch}{1.25}
        \resizebox{\textwidth}{!}{
            \begin{tabular}{lp{0.1\textwidth}p{0.1\textwidth}p{0.1\textwidth}p{0.1\textwidth}p{0.1\textwidth}p{0.1\textwidth}p{0.1\textwidth}p{0.1\textwidth}p{0.1\textwidth}} \toprule
             \multirow{2}*{Method} & \multirow{2}*{Sparsity} & \multicolumn{4}{c}{Improvement HR (\%)} &  \multicolumn{4}{c}{Improvement of Precision (\%)} \\ \cmidrule(r){3-6} \cmidrule(r){7-10}
            & & @5 & @10 & @15 & @20 & @5 & @10 & @15 & @20\\ \midrule
            \multirow{2}*{Compared to PEIA}&99.84\%&45.15&33.93&28.03&25.88&44.91&34.19&27.96&25.61\\
            &97.08\%&24.73&28.95&31.10&29.83&24.80&28.92&31.28&29.55\\ \midrule
            \multirow{2}*{Compared to Wide\&Deep}&99.84\%&129.55&93.70&72.39&57.68&128.30&93.83&72.46&58.06\\
            &97.08\%&17.70&21.11&22.85&23.31&17.57&20.92&22.57&23.24\\ \midrule
            \multirow{2}*{Compared to DeepFM}&99.84\%&22.05&20.54&19.46&19.00&21.61&20.77&19.00&19.51\\
            &97.08\%&14.51&21.63&16.87&18.79&14.47&15.63&16.88&18.75\\ \midrule
            \multirow{2}*{Compared to TagGCN}&99.84\%&16.65&26.07&29.89&34.52&16.35&26.61&29.35&34.25\\
            &97.08\%&10.52&15.27&15.38&15.93&10.53&14.91&15.42&15.74\\ \midrule \midrule
            
             \multirow{2}*{Method} & \multirow{2}*{Sparsity} & \multicolumn{4}{c}{Improvement NDCG (\%)} &  \multicolumn{4}{c}{Improvement of MRR (\%)} \\ \cmidrule(r){3-6} \cmidrule(r){7-10}
            & & @5 & @10 & @15 & @20 & @5 & @10 & @15 & @20\\ \midrule
            \multirow{2}*{Compared to PEIA}&99.84\%&60.31&52.16&47.55&45.65&68.18&63.45&61.27&60.49\\
            &97.08\%&15.72&18.09&19.22&19.14&12.23&13.46&13.83&13.89\\ \midrule
            \multirow{2}*{Compared to Wide\&Deep}&99.84\%&160.66&134.15&118.00&106.25&178.10&162.10&153.74&148.92\\
            &97.08\%&6.58&8.84&9.83&10.34&2.35&3.58&3.92&4.10\\ \midrule
            \multirow{2}*{Compared to DeepFM}&99.84\%&28.90&27.38&26.59&26.09&32.35&31.43&31.04&30.98\\
            &97.08\%&4.39&5.74&6.44&7.34&1.45&2.61&2.74&13.64\\ \midrule
            \multirow{2}*{Compared to TagGCN}&99.84\%&15.89&20.27&22.02&23.93&15.47&17.68&18.19&18.87\\
            &97.08\%&5.68&7.93&8.27&8.61&3.63&4.71&4.86&4.99\\
            \bottomrule
        \end{tabular}}
    \end{table}

\subsection{Ablation study (RQ2)}
To validate the effectiveness of the four components in HDBN, we individually ablate each component of HDBN and get the variants.
\begin{itemize}
    \item \textbf{w/o emotion heterogeneity across users (w/o EHAU)}: This variant replaces the prior of posterior LED (i.e., $\boldsymbol{\mathcal{S}}_u$) with standard Gaussian distributions.
    \item \textbf{w/o emotion heterogeneity within a user (w/o EHWU)}: This variant replaces the posterior LED with fixed emotion representation.
    \item \textbf{w/o music mood preference heterogeneity across users (w/o PHAU)}: This variant replaces the group-specific music mood preference prediction model (i.e., $\{\boldsymbol{\psi}_g\}$) with the global model (i.e., $\boldsymbol{\psi}$).
    \item \textbf{w/o music mood preference heterogeneity within a user (w/o PHWU)}: This variant replaces the BNNs (i.e., $\{\boldsymbol{\psi}_g\}$) with traditional neural networks with fixed weights.
\end{itemize}

    \begin{table}[htbp]
        \centering
        \captionof{table}{Recommendation Results of the Ablation Experiment on EmoMusicLJ}
        \label{ablation1}
        \renewcommand{\arraystretch}{1.25}
        \resizebox{\textwidth}{!}{
            \begin{tabular}{lp{0.1\textwidth}p{0.1\textwidth}p{0.1\textwidth}p{0.1\textwidth}p{0.1\textwidth}p{0.1\textwidth}p{0.1\textwidth}p{0.1\textwidth}p{0.1\textwidth}} \toprule
             \multirow{2}*{Method type} & \multirow{2}*{Method} & \multicolumn{4}{c}{HR} &  \multicolumn{4}{c}{Precision} \\ \cmidrule(r){3-6} \cmidrule(r){7-10}
            && @5 & @10 & @15 & @20 & @5 & @10 & @15 & @20\\ \midrule
            \multirow{4}*{Variants} & w/o EHAU
            &\makecell[l]{0.1352\\(2.00\%)}
            &\makecell[l]{0.1662\\(1.81\%)}
            &\makecell[l]{0.1867\\(1.71\%)}
            &\makecell[l]{0.2040\\(1.37\%)}
            &\makecell[l]{0.0270\\(2.22\%)}
            &\makecell[l]{\underline{0.0166}\\(1.81\%)}
            &\makecell[l]{0.0124\\(2.42\%)}
            &\makecell[l]{\underline{0.0102}\\(0.98\%)}\\
            & w/o EHWU
            &\makecell[l]{\underline{0.1363}\\(1.17\%)}
            &\makecell[l]{\underline{0.1663}\\(1.74\%)}
            &\makecell[l]{\underline{0.1869}\\(1.61\%)}
            &\makecell[l]{\underline{0.2044}\\(1.17\%)}
            &\makecell[l]{\underline{0.0273}\\(1.10\%)}
            &\makecell[l]{\underline{0.0166}\\(1.81\%)}
            &\makecell[l]{\underline{0.0125}\\(1.60\%)}
            &\makecell[l]{\underline{0.0102}\\(0.98\%)}\\
            & w/o PHAU
            &\makecell[l]{0.1315\\(4.84\%)}
            &\makecell[l]{0.1633\\(3.61\%)}
            &\makecell[l]{0.1858\\(2.21\%)}
            &\makecell[l]{0.2016\\(2.58\%)}
            &\makecell[l]{0.0263\\(4.94\%)}
            &\makecell[l]{0.0163\\(3.68\%)}
            &\makecell[l]{0.0123\\(3.25\%)}
            &\makecell[l]{0.0100\\(3.00\%)}\\
            & w/o PHWU
            &\makecell[l]{0.1357\\(1.62\%)}
            &\makecell[l]{0.1662\\(1.81\%)}
            &\makecell[l]{0.1864\\(1.88\%)}
            &\makecell[l]{0.2021\\(2.33\%)}
            &\makecell[l]{0.0271\\(1.85\%)}
            &\makecell[l]{\underline{0.0166}\\(1.81\%)}
            &\makecell[l]{0.0124\\(2.42\%)}
            &\makecell[l]{0.0101\\(1.98\%)}\\ \midrule
            our & HDBN&\textbf{0.1379}&\textbf{0.1692}&\textbf{0.1899}&\textbf{0.2068}&\textbf{0.0276}&\textbf{0.0169}&\textbf{0.0127}&\textbf{0.0103}\\ \midrule \midrule

            \multirow{2}*{Method type} & \multirow{2}*{Method} & \multicolumn{4}{c}{NDCG} &  \multicolumn{4}{c}{MRR} \\ \cmidrule(r){3-6} \cmidrule(r){7-10}
            && @5 & @10 & @15 & @20 & @5 & @10 & @15 & @20\\ \midrule
            \multirow{4}*{Variants} & w/o EHAU
            &\makecell[l]{\underline{0.1084}\\(1.20\%)}
            &\makecell[l]{\underline{0.1183}\\(1.27\%)}
            &\makecell[l]{\underline{0.1238}\\(1.13\%)}
            &\makecell[l]{\underline{0.1278}\\(1.10\%)}
            &\makecell[l]{\underline{0.0996}\\(0.70\%)}
            &\makecell[l]{\underline{0.1036}\\(0.87\%)}
            &\makecell[l]{\underline{0.1052}\\(0.86\%)}
            &\makecell[l]{\underline{0.1062}\\(0.85\%)}\\
            & w/o EHWU
            &\makecell[l]{\underline{0.1084}\\(1.20\%)}
            &\makecell[l]{0.1180\\(1.53\%)}
            &\makecell[l]{0.1325\\(1.38\%)}
            &\makecell[l]{0.1276\\(1.25\%)}
            &\makecell[l]{0.0992\\(1.11\%)}
            &\makecell[l]{0.1031\\(1.36\%)}
            &\makecell[l]{0.1047\\(1.34\%)}
            &\makecell[l]{0.1057\\(1.32\%)}\\
            & w/o PHAU
            &\makecell[l]{0.1026\\(6.92\%)}
            &\makecell[l]{0.1128\\(6.21\%)}
            &\makecell[l]{0.1188\\(5.39\%)}
            &\makecell[l]{0.1225\\(5.47\%)}
            &\makecell[l]{0.0931\\(7.73\%)}
            &\makecell[l]{0.0972\\(7.51\%)}
            &\makecell[l]{0.0990\\(7.17\%)}
            &\makecell[l]{0.0999\\(7.21\%)}\\
            & w/o PHWU
            &\makecell[l]{0.1083\\(1.29\%)}
            &\makecell[l]{0.1181\\(1.44\%)}
            &\makecell[l]{0.1235\\(1.38\%)}
            &\makecell[l]{0.1272\\(1.57\%)}
            &\makecell[l]{0.0992\\(1.11\%)}
            &\makecell[l]{0.1033\\(1.16\%)}
            &\makecell[l]{0.1049\\(1.14\%)}
            &\makecell[l]{0.1058\\(1.23\%)}\\ \midrule
            our & HDBN&\textbf{0.1097}&\textbf{0.1198}&\textbf{0.1252}&\textbf{0.1292}&\textbf{0.1003}&\textbf{0.1045}&\textbf{0.1061}&\textbf{0.1071}\\
            \bottomrule
        \end{tabular}}
        \footnotesize{Notes: Best results are highlighted in bold and the second-best results are underlined. Values in parentheses indicate the improvement of HDBN over the variants.}
    \end{table}

    \begin{table}[htbp]
        \centering
        \captionof{table}{Recommendation Results of the Ablation Experiment on EmoMusicLJ-small}
        \label{ablation2}
        \renewcommand{\arraystretch}{1.25}
        \resizebox{\textwidth}{!}{
            \begin{tabular}{lp{0.1\textwidth}p{0.1\textwidth}p{0.1\textwidth}p{0.1\textwidth}p{0.1\textwidth}p{0.1\textwidth}p{0.1\textwidth}p{0.1\textwidth}p{0.1\textwidth}} \toprule
             \multirow{2}*{Method type} & \multirow{2}*{Method} & \multicolumn{4}{c}{HR} &  \multicolumn{4}{c}{Precision} \\ \cmidrule(r){3-6} \cmidrule(r){7-10}
            && @5 & @10 & @15 & @20 & @5 & @10 & @15 & @20\\ \midrule
            \multirow{4}*{Variants} & w/o EHAU
            &\makecell[l]{0.3229\\(0.68\%)}
            &\makecell[l]{0.3929\\(0.23\%)}
            &\makecell[l]{\underline{0.4507}\\(0.91\%)}
            &\makecell[l]{\underline{0.4943}\\(0.65\%)}
            &\makecell[l]{0.0646\\(0.62\%)}
            &\makecell[l]{0.0393\\(0.25\%)}
            &\makecell[l]{0.0300\\(1.00\%)}
            &\makecell[l]{0.0247\\(0.81\%)}\\
            & w/o EHWU
            &\makecell[l]{0.3238\\(0.40\%)}
            &\makecell[l]{\textbf{0.3984}\\(-1.15\%)}
            &\makecell[l]{0.4470\\(1.74\%)}
            &\makecell[l]{0.4879\\(1.97\%)}
            &\makecell[l]{0.0648\\(0.31\%)}
            &\makecell[l]{\textbf{0.0398}\\(-1.01\%)}
            &\makecell[l]{0.0298\\(1.68\%)}
            &\makecell[l]{0.0244\\(2.05\%)}\\
            & w/o PHAU
            &\makecell[l]{0.3238\\(0.40\%)}
            &\makecell[l]{0.3925\\(0.33\%)}
            &\makecell[l]{0.4470\\(1.74\%)}
            &\makecell[l]{0.4898\\(1.57\%)}
            &\makecell[l]{0.0648\\(0.31\%)}
            &\makecell[l]{0.0392\\(0.51\%)}
            &\makecell[l]{0.0298\\(1.68\%)}
            &\makecell[l]{0.0248\\(1.63\%)}\\
            & w/o PHWU
            &\makecell[l]{\textbf{0.3297}\\(-1.40\%)}
            &\makecell[l]{0.3929\\(0.23\%)}
            &\makecell[l]{0.4411\\(3.11\%)}
            &\makecell[l]{0.4902\\(1.49\%)}
            &\makecell[l]{\textbf{0.0659}\\(-1.37\%)}
            &\makecell[l]{0.0393\\(0.25\%)}
            &\makecell[l]{0.0294\\(3.06\%)}
            &\makecell[l]{0.0245\\(1.63\%)}\\ \midrule
            our & HDBN&\underline{0.3251}&\underline{0.3938}&\textbf{0.4548}&\textbf{0.4975}&\underline{0.0650}&\underline{0.0394}&\textbf{0.0303}&\textbf{0.0249}\\ \midrule \midrule

            \multirow{2}*{Method type} & \multirow{2}*{Method} & \multicolumn{4}{c}{NDCG} &  \multicolumn{4}{c}{MRR} \\ \cmidrule(r){3-6} \cmidrule(r){7-10}
            && @5 & @10 & @15 & @20 & @5 & @10 & @15 & @20\\ \midrule
            \multirow{4}*{Variants} & w/o EHAU
            &\makecell[l]{0.2698\\(0.96\%)}
            &\makecell[l]{\underline{0.2923}\\0.72\%)}
            &\makecell[l]{\underline{0.3076}\\(0.91\%)}
            &\makecell[l]{\underline{0.3179}\\(0.82\%)}
            &\makecell[l]{\underline{0.2523}\\(1.11\%)}
            &\makecell[l]{\underline{0.2614}\\(0.99\%)}
            &\makecell[l]{\underline{0.2660}\\(1.05\%)}
            &\makecell[l]{\underline{0.2684}\\(1.04\%)}\\
            & w/o EHWU
            &\makecell[l]{0.2679\\(1.68\%)}
            &\makecell[l]{0.2917\\(0.93\%)}
            &\makecell[l]{0.3046\\(1.90\%)}
            &\makecell[l]{0.3143\\(1.97\%)}
            &\makecell[l]{0.2495\\(2.24\%)}
            &\makecell[l]{0.2592\\(1.85\%)}
            &\makecell[l]{0.2631\\(2.17\%)}
            &\makecell[l]{0.2653\\(2.22\%)}\\
            & w/o PHAU
            &\makecell[l]{0.2686\\(1.41\%)}
            &\makecell[l]{0.2906\\(1.31\%)}
            &\makecell[l]{0.3050\\(1.77\%)}
            &\makecell[l]{0.3151\\(1.71\%)}
            &\makecell[l]{0.2505\\(1.84\%)}
            &\makecell[l]{0.2594\\(1.77\%)}
            &\makecell[l]{0.2636\\(1.94\%)}
            &\makecell[l]{0.2660\\(1.95\%)}\\
            & w/o PHWU
            &\makecell[l]{\underline{0.2711}\\(0.48\%)}
            &\makecell[l]{0.2903\\(1.41\%)}
            &\makecell[l]{0.3030\\(2.44\%)}
            &\makecell[l]{0.3147\\(1.84\%)}
            &\makecell[l]{0.2504\\(1.88\%)}
            &\makecell[l]{0.2587\\(2.05\%)}
            &\makecell[l]{0.2624\\(2.44\%)}
            &\makecell[l]{0.2652\\(2.26\%)}\\ \midrule
            our & HDBN&\textbf{0.2724}&\textbf{0.2944}&\textbf{0.3104}&\textbf{0.3205}&\textbf{0.2551}&\textbf{0.2640}&\textbf{0.2688}&\textbf{0.2712}\\
            \bottomrule
        \end{tabular}}
        \footnotesize{Notes: Best results are highlighted in bold and the second-best results are underlined. Values in parentheses indicate the improvement of HDBN over the variants.}
    \end{table}

Table \ref{ablation1} and Table \ref{ablation2} present the results of the ablation experiments. From Table \ref{ablation1}, we have the following observations:
    \begin{enumerate}
        \item [(1)] HDBN outperforms all variant methods on dataset EmoMusicLJ and achieves the best results on most metrics on dataset EMoMusicLJ-small. The results demonstrate the effectiveness of the four components corresponding to the four types of heterogeneity in HDBN.
        \item [(2)] In Table \ref{ablation1}, the variant method w/o PHAU exhibits the most significant performance decrease. This suggests that considering the differences in music mood preferences across users is crucial in emotion-aware music recommendation. The variant method w/o PHWU exhibits the second highest degree of performance degradation. This indicates that considering the music mood preference heterogeneity within a user under the same emotion contributes to improving recommendation performance.
        \item [(3)] In Table \ref{ablation2}, we observe results that differ from those shown in Table 6. The performance drop for the w/o EHWU and w/o PHWU variant methods is more significant. This suggests that, on the EmoMusicLJ-small dataset, considering emotion heterogeneity and music mood preference heterogeneity within users is more effective. We argue that this may be due to the relatively lower sparsity of EmoMusicLJ-small, where individual user data is more abundant, allowing the model to better capture the heterogeneity within users. Additionally, w/o EHAU has the least performance drop, indicating that the improvement brought by considering emotion heterogeneity across users is not significant. We argue that this may be because the MoMusicLJ-small dataset is much smaller than EmoMusicLJ, and thus it cannot fully represent emotion heterogeneity across users.
    \end{enumerate}

In conclusion, ablation studies demonstrate that the four components of HDBN can effectively improve the performance of emotion-aware recommendations. Furthermore, when individual user data is rich and the dataset size is relatively small, considering emotion heterogeneity and music mood preference heterogeneity within users is more effective. In contrast, when individual user data is relatively sparse, but the overall dataset size is large, considering emotion heterogeneity and music mood preference heterogeneity across users can be more effective.

\subsection{Sensitivity analysis (RQ3)}
In this section, we analyze the influences of key hyperparameters, including the number of user groups $G$, the number of layers in music mood preference prediction model, hyperparameter $\alpha$, the dimensions of user and music embeddings, learning rate and training batch size, the number of negative samples, and hyperparameters $\lambda_1$, $\lambda_2$, $\lambda_3$, and $\lambda_4$.

\subsubsection{The number of user groups $G$}
\begin{figure}[h]
  \centering
  \includegraphics[width=\linewidth]{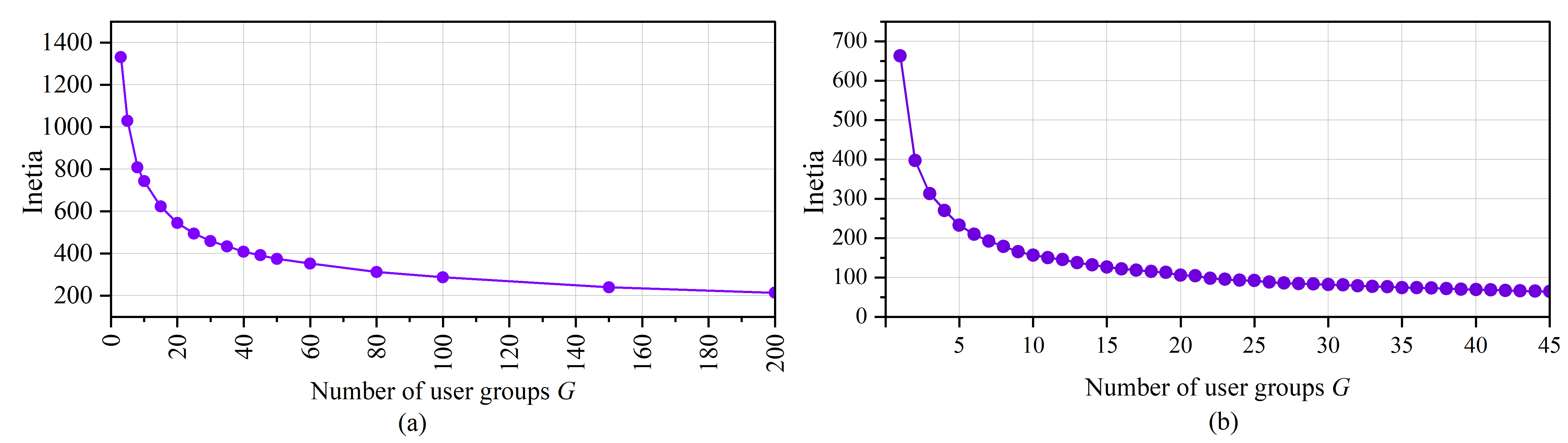}
  \caption{The relationship between Inertia and $G$ on EmoMusicLJ and EmoMusicLJ-small.}
  \Description{}
  \label{G}
\end{figure}

Inspired by Zhang et al. \cite{Zhang2014}, we cluster users based on the music genres they have listened to using K-means, rather than clustering users based on the specific music tracks they have listened to. Each user is represented as a genre vector, where the elements of the vector indicate the proportion of each genre in the user’s listening history. The optimal G is determined using the Elbow method \cite{Liu2020}. Denote the sum of the distances of all users to the center of their respective groups as “Inertia”. The Elbow method suggests that as $G$ increases, Inertia decreases. As $G$ approaches the optimal value, the rate of decrease in Inertia diminishes and tends to plateau. Consequently, the relationship between Inertia and $G$ exhibits an elbow-shaped curve, with the optimal value of $G$ corresponding to this inflection point. Figure \ref{G}(a) and Figure \ref{G}(b) show the relationship between Inertia and $G$ on the EmoMusicLJ and EmoMusicLJ-small datasets, respectively. We set $G$ to 50 and 10 on EmoMusicLJ and EmoMusicLJ-small datasets, respectively.

\subsubsection{The number of layers and $\alpha$ for music mood preference prediction model}
\begin{figure}[h]
  \centering
  \includegraphics[width=\linewidth]{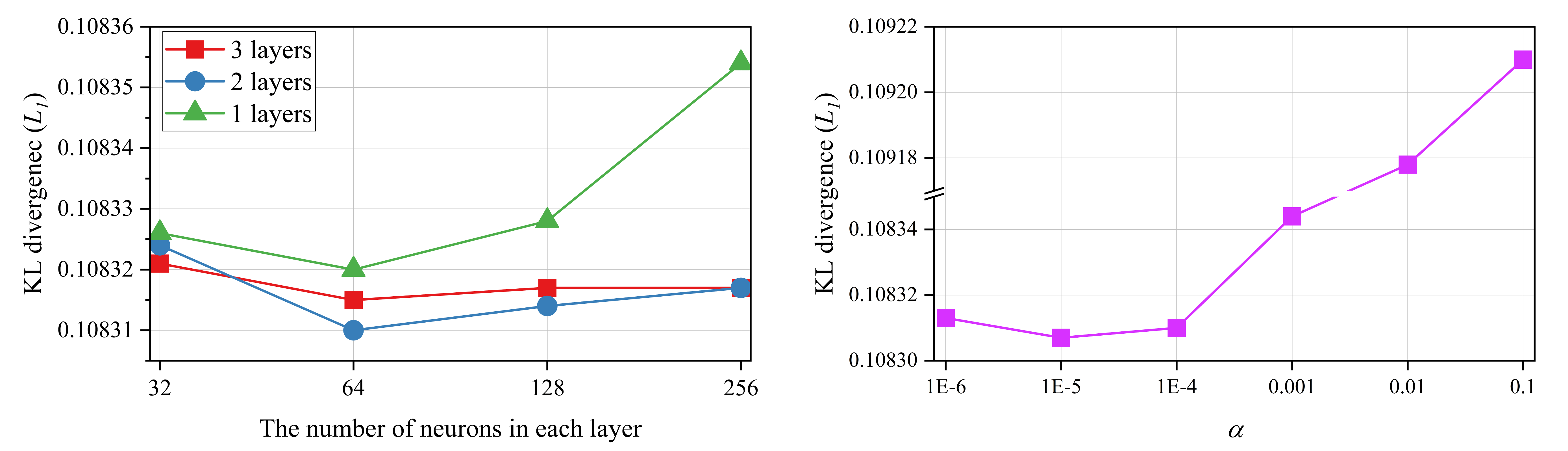}
  \caption{The influence of the number of layers and neurons in each layer.}
  \Description{}
  \label{layers and neurons D1}
\end{figure}

\begin{figure}[h]
  \centering
  \includegraphics[width=\linewidth]{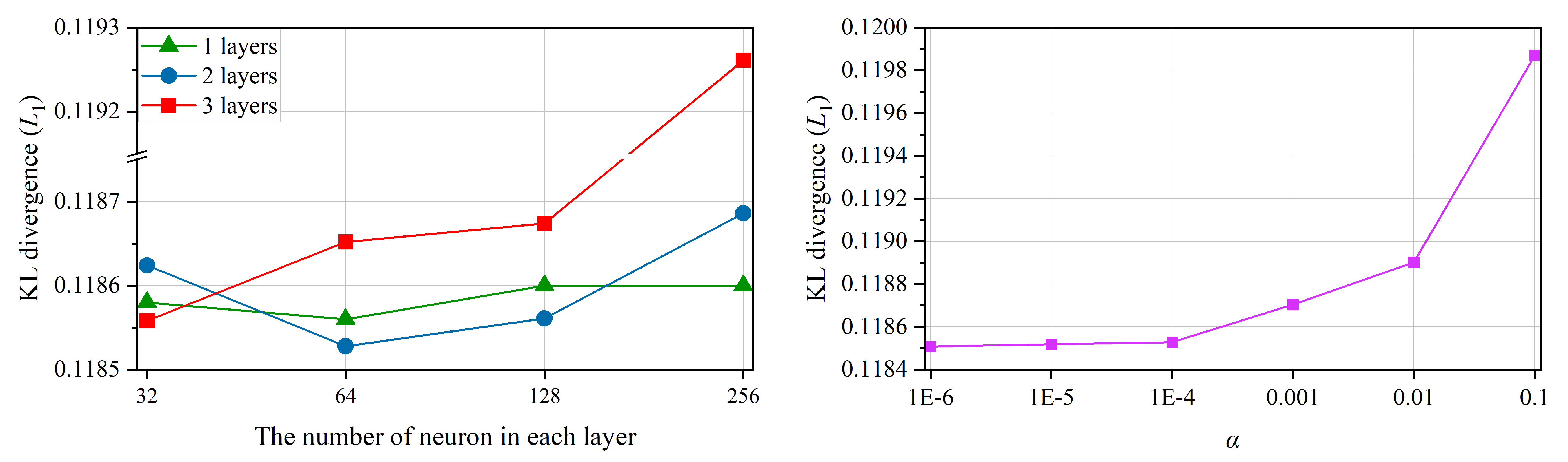}
  \caption{The influence of the number of layers and neurons in each layer.}
  \Description{}
  \label{layers and neurons D2}
\end{figure}

BNNs with parameters of $\boldsymbol{\psi}$ and $\{\boldsymbol{\psi}_g\}$ were trained to infer user-preferred music mood $\boldsymbol{l}_{u,v}$. We searched the layer numbers of the BNN within {1, 2, 3}, the number of neurons of each layer within {32, 64, 128, 256}, and the value of $\alpha$ in Equation (15) within {1e-6, 1e-5, 1e-4, 0.001, 0.01, 0.1} to identify the optimal parameter configuration. Figure \ref{layers and neurons D1} and Figure \ref{layers and neurons D2} present the prediction performance of user-preferred music mood under different configurations on EmoMusicLJ and EmoMusicLJ-small, respectively. ultimately, we set the number of layers to 2, the neuron number of each layer to 64, and $\alpha$ to 1e-5 yields the best performance on EmuMusicLJ dataset, the number of layers to 2, the neuron number of each layer to 64, and $\alpha$ to 1e-6 on EmoMusicLJ-small dataset.

With $\boldsymbol{\psi}$ determined, we initialize $\{\boldsymbol{\psi}_g\}$ with $\boldsymbol{\psi}$ and fine-tune it on data of each user group. In comparison to the BNN $\boldsymbol{\psi}$ with $\mathrm{KL}(\boldsymbol{o}_v\parallel \boldsymbol{l}_{u,v})=0.1083$ and $\mathrm{KL}(\boldsymbol{o}_v\parallel \boldsymbol{l}_{u,v})=0.1185$ on EmoMusicLJ and EmoMusicLJ-small datasets, the average value of $\mathrm{KL}(\boldsymbol{o}_v\parallel \boldsymbol{l}_{u,v})$ with the fine-tuned BNNs $\{\boldsymbol{\psi}_g\}$ is 0.0972 and 0.1076, respectively. This indicates that user grouping can enhance the prediction performance of music mood preferences.

\subsubsection{The dimensions of user and music embeddings}
\begin{figure}[h]
  \centering
  \includegraphics[width=\linewidth]{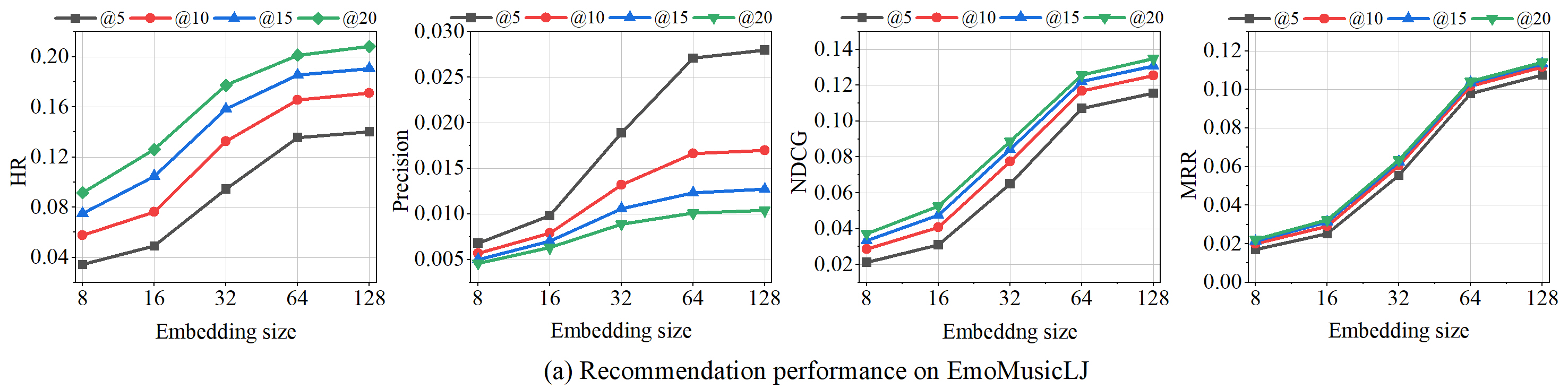}
  \includegraphics[width=\linewidth]{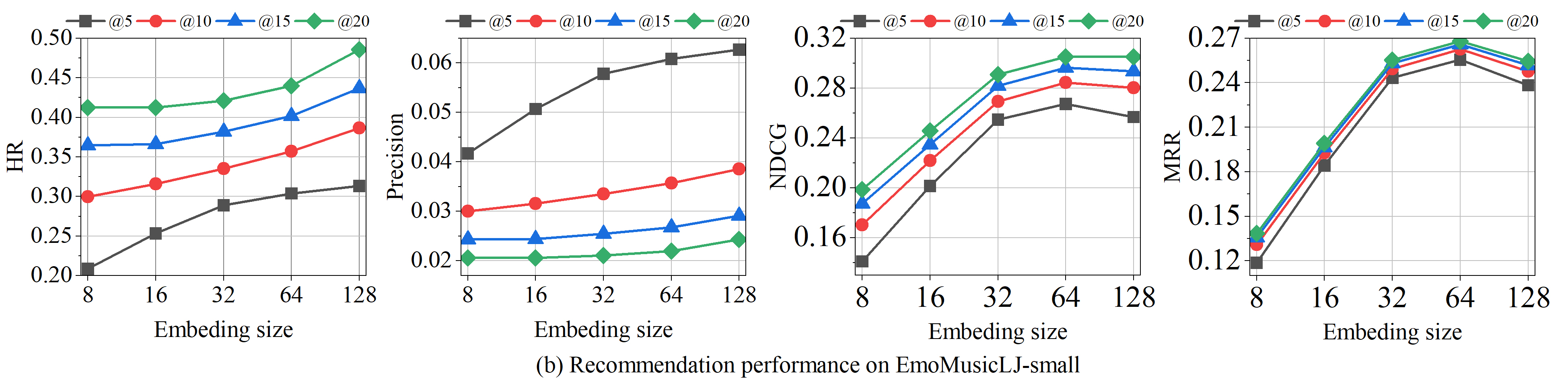}
  \caption{Recommendation performances with different embedding dimensions on both datasets.}
  \Description{}
  \label{embedding_size}
\end{figure}

The dimension of embeddings is an important factor affecting recommendation performance, while it also impacts the training efficiency \cite{Wang2023}. A large embedding dimension can adequately capture the features of users and music, but it also requires more computational resources and time. We searched for the dimension value within {8, 16, 32, 64, 128}. Figures \ref{embedding_size}(a) and \ref{embedding_size}(b) illustrate the recommendation performance across various embedding dimensions on both datasets, respectively.

It can be observed that smaller dimensions (e.g., 8 and 16) lead to poorer recommendation performance. The recommendation performance improves as the embedding dimension increases. However, beyond a certain threshold (e.g., 64 and 128), the enhancement in recommendation performance becomes marginal. Considering both the validity and computational complexity, we opt for an embedding dimension of 64.

\subsubsection{Learning rate and training batch size}
\begin{figure}[h]
  \centering
  \includegraphics[width=\linewidth]{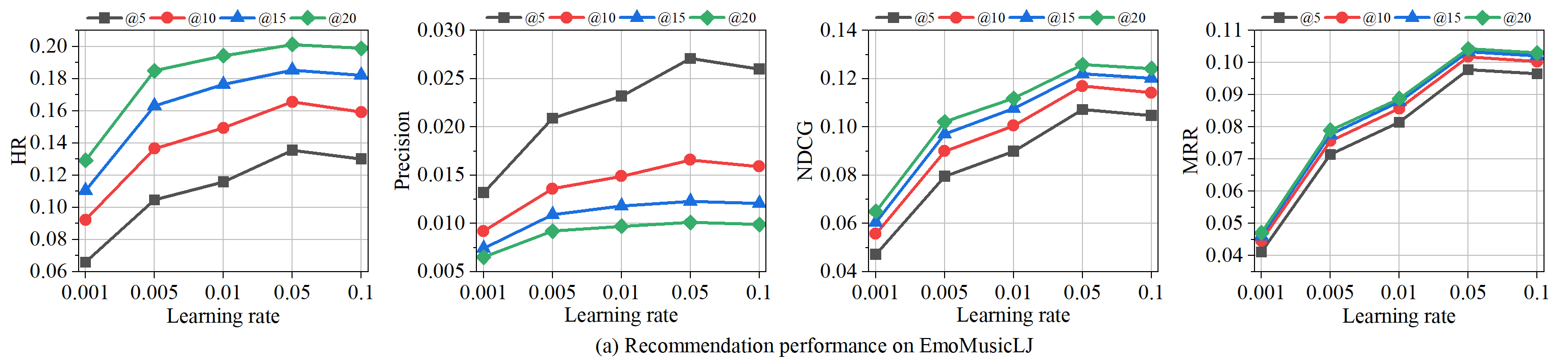}
  \includegraphics[width=\linewidth]{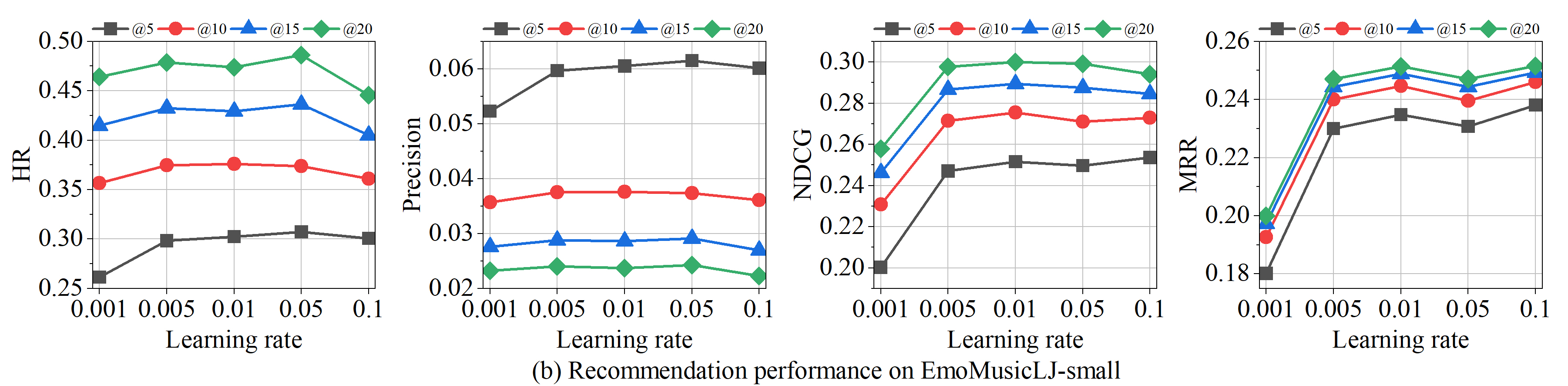}
  \caption{Recommendation performances with different learning rates on both datasets.}
  \Description{}
  \label{learning_rate}
\end{figure}

We searched for the influence of learning rate within {0.001, 0.005, 0.01, 0.05, 0.1}. Figures \ref{learning_rate}(a) and \ref{learning_rate}(b) display the results on both datasets, respectively. We can observe that recommendation performance initially improves and then declines with increasing learning rates. We determine the optimal learning rate to be 0.05.

\begin{figure}[h]
  \centering
  \includegraphics[width=\linewidth]{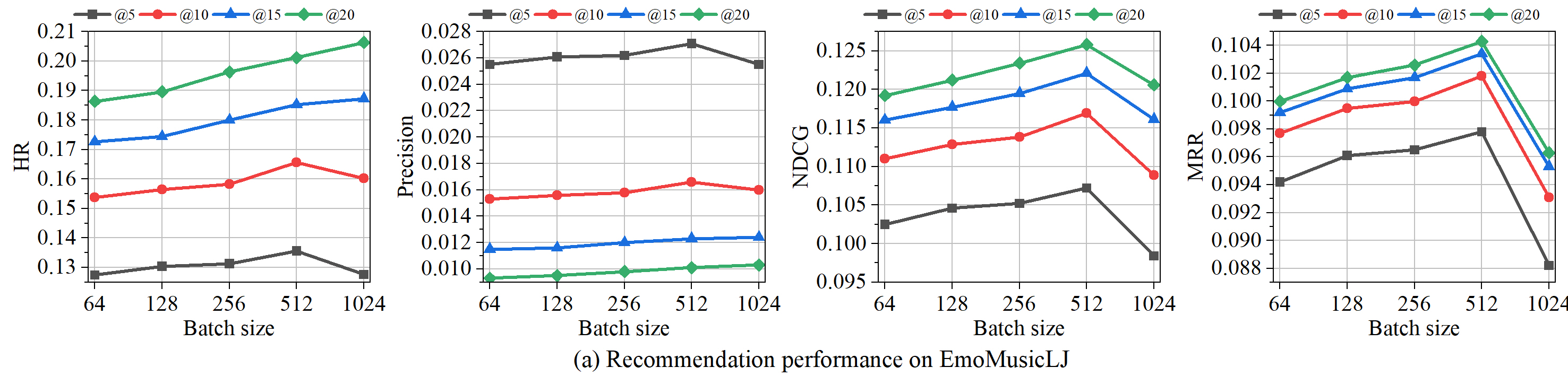}
  \includegraphics[width=\linewidth]{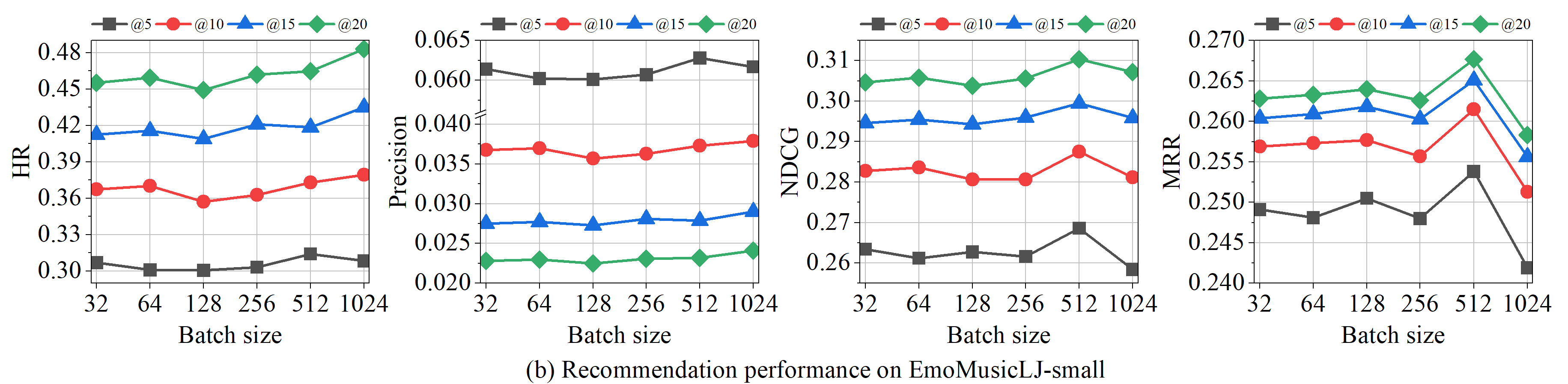}
  \caption{Recommendation performances with different training batch sizes on both datasets.}
  \Description{}
  \label{batch}
\end{figure}

Figure \ref{batch} shows the influence of batch size on recommendation performance. We searched for the batch size in {64, 128, 256, 512, 1204}. It can be observed that with the increase in batch size, the recommendation performance initially improves and then declines. This may be due to a small batch size causing instability in model training, while a large batch size hinder model generalization \cite{He2019}. Finally, we set the batch size to be 512 on both two datasets.

\subsubsection{Number of negative samples}
\begin{figure}[h]
  \centering
  \includegraphics[width=\linewidth]{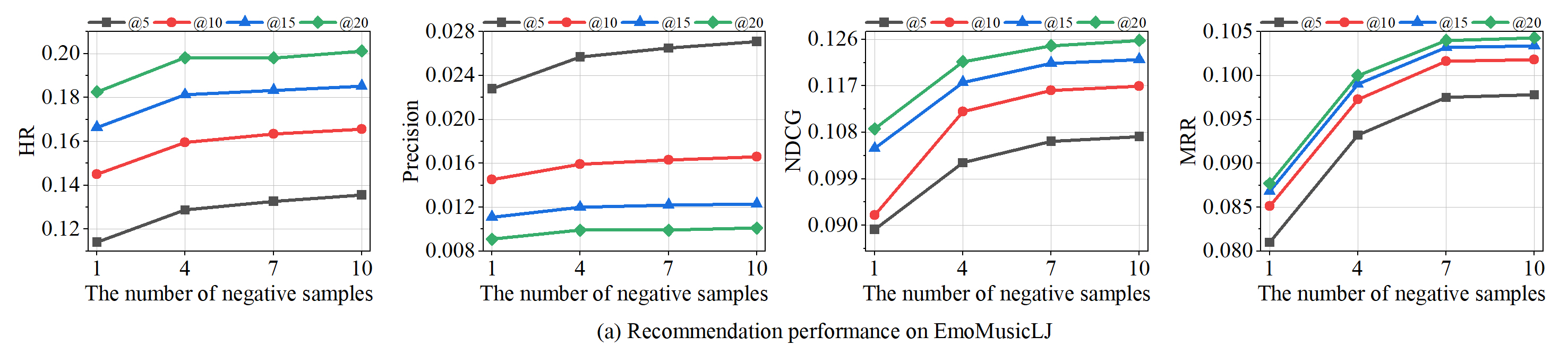}
  \includegraphics[width=\linewidth]{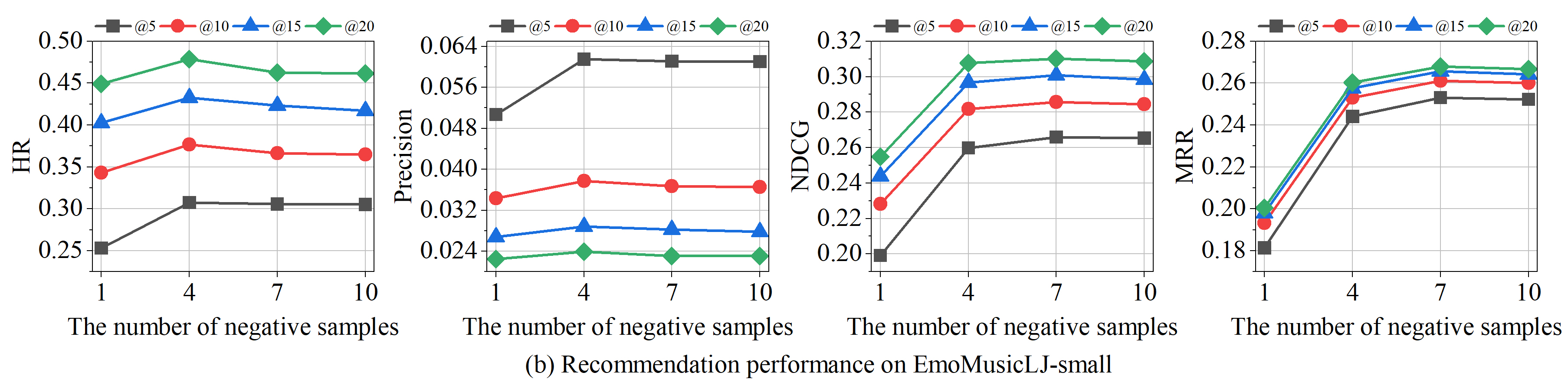}
  \caption{Recommendation performances with different negative sample sizes on both datasets.}
  \Description{}
  \label{neg_num}
\end{figure}

Sampling-based BPR pairwise loss learning strategy was used to train our model, making the number of negative samples a key factor influencing recommendation performance. We searched for the number of negative samples within {1, 4, 7, 10}. Figures \ref{neg_num}(a) and \ref{neg_num}(b) show the influence of the number of negative samples on recommendation performance on both datasets, respectively. It can be observed that consistent with previous findings \cite{Chen2023}, more negative samples is more beneficial. However, the marginal benefit diminishes as the number of negative samples increases. Considering computational efficiency, we set the number of negative samples to 10 on EmoMusicLJ and 7 on EmoMusicLJ-small.

\subsubsection{Hyperparameters $\lambda_1$, $\lambda_2$, $\lambda_3$, and $\lambda_4$}
\begin{figure}[h]
  \centering
  \includegraphics[width=\linewidth]{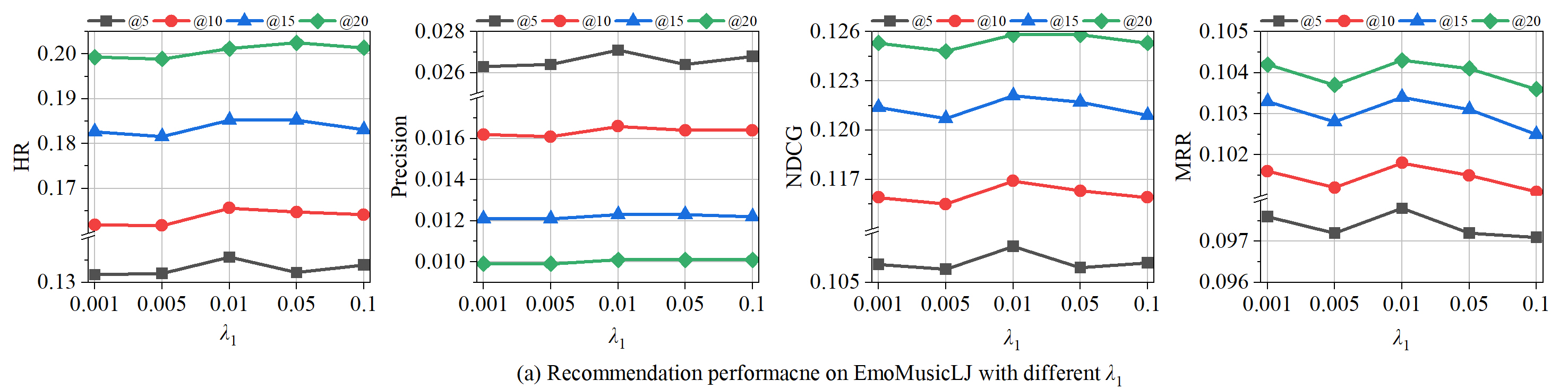}
  \includegraphics[width=\linewidth]{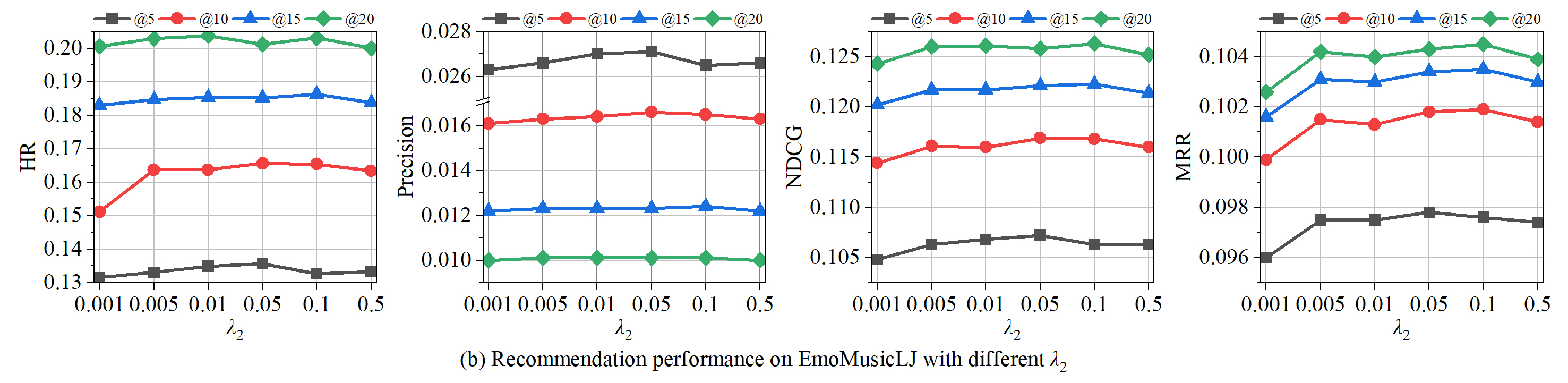}
  \includegraphics[width=\linewidth]{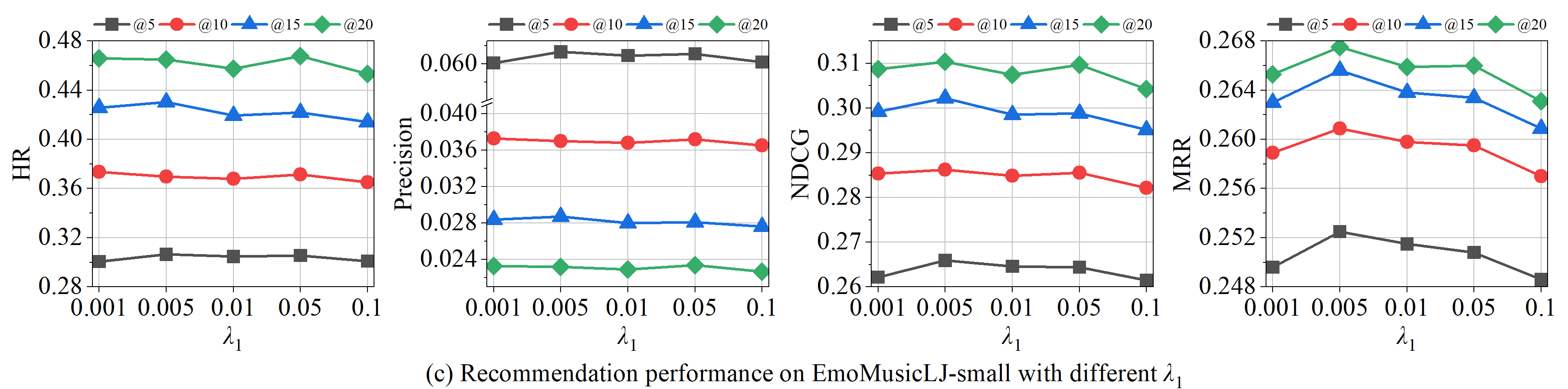}
  \includegraphics[width=\linewidth]{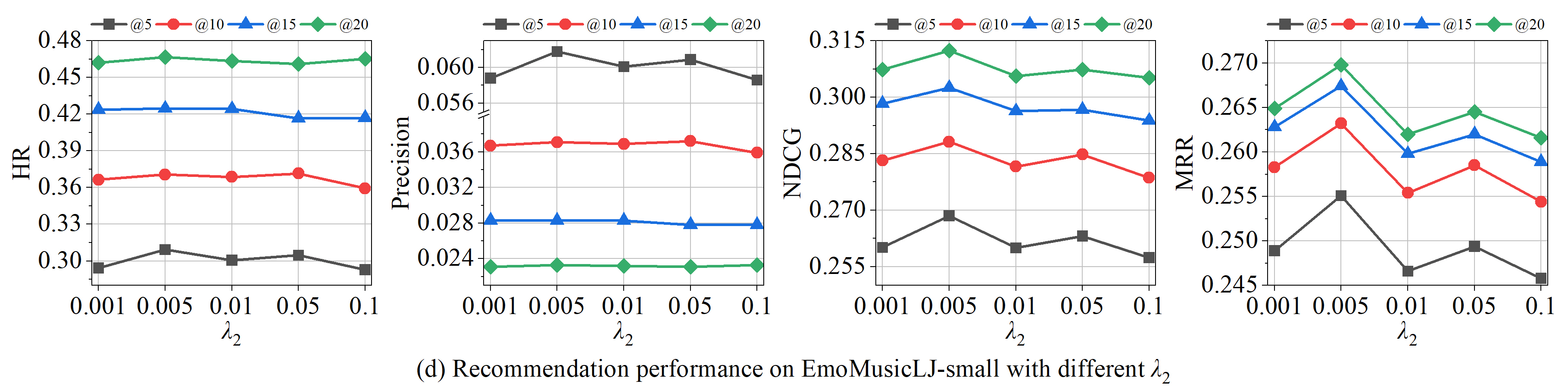}
  \caption{Recommendation performances with different $\lambda_1$ and $\lambda_2$ on both datasets.}
  \Description{}
  \label{lambda_1_2}
\end{figure}

$\lambda_1$ and $\lambda_2$ are the relative weights of $\mathcal{L}_{\mathrm{KL1}}$ and $\mathcal{L}_{\mathrm{KL2}}$, respectively. In Figure \ref{lambda_1_2}, we investigate the influence of $\lambda_1$ and $\lambda_2$ on recommendation performance. It can be observed that all the evaluation metrics first increase and then decrease with the increase of $\lambda_1$ and $\lambda_2$. We finally set $\lambda_1$=0.01 and $\lambda_2$=0.05 on EmoMusicLJ dataset, and $\lambda_1$=0.005 and $\lambda_2$=0.005 on EmoMusicLJ-small dataset. We also observe that $\lambda_2$ no less than $\lambda_1$. This is might because $\lambda_1$ corresponds to inferring the personalized prior LED from user representations, while $\lambda_2$ corresponds to inferring the posterior LED from user self-reported emotion tags. User self-reported emotion tags better reflect the posterior LED, so the constraint of $\lambda_2$ is stronger than that of $\lambda_1$.

\begin{figure}[h]
  \centering
  \includegraphics[width=\linewidth]{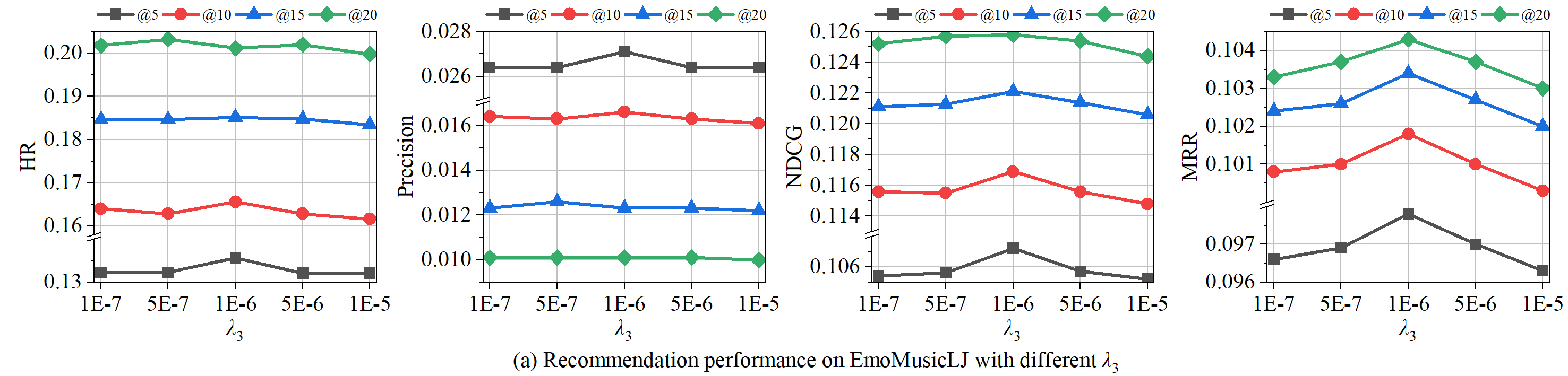}
  \includegraphics[width=\linewidth]{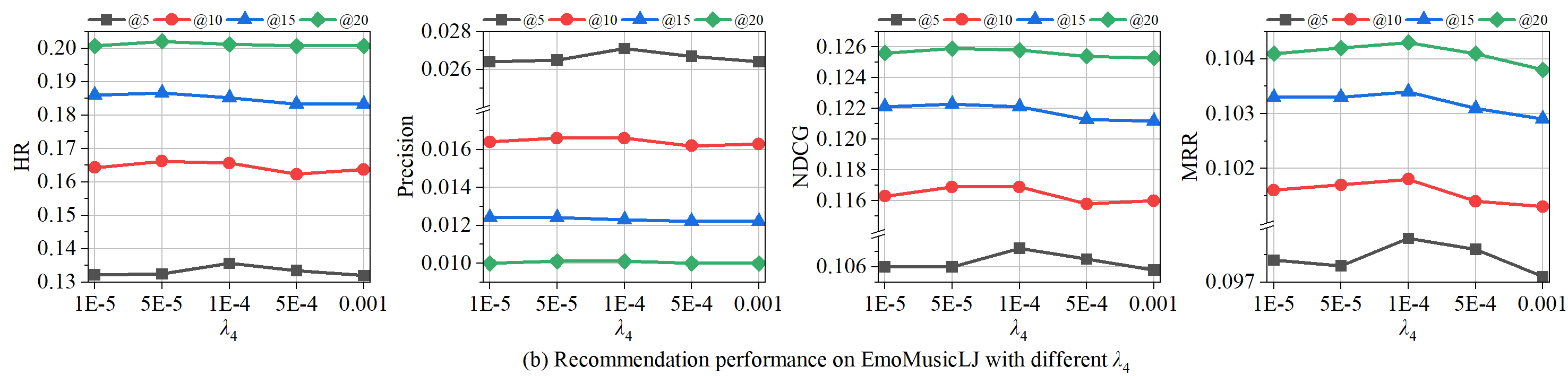}
  \includegraphics[width=\linewidth]{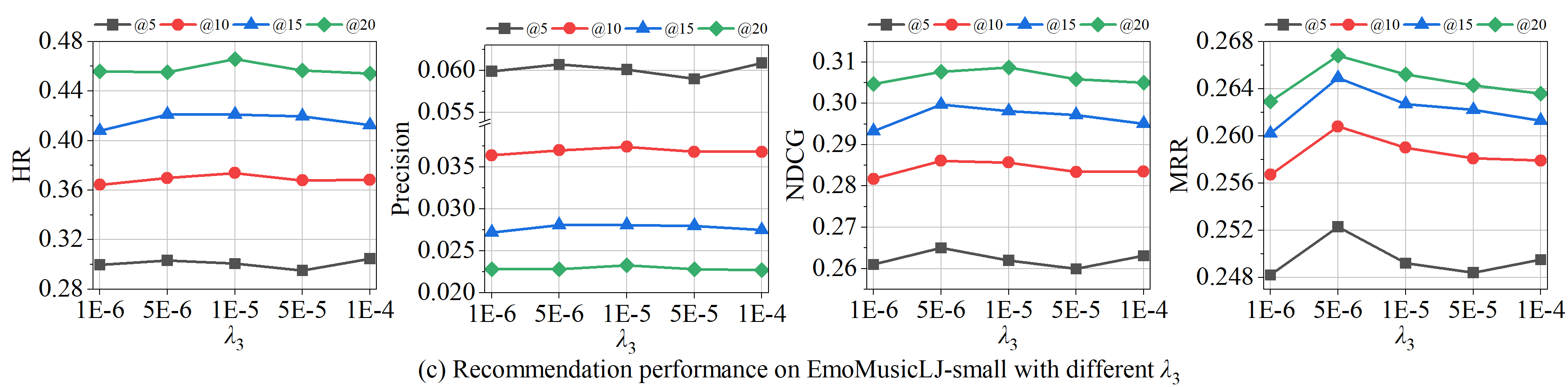}
  \includegraphics[width=\linewidth]{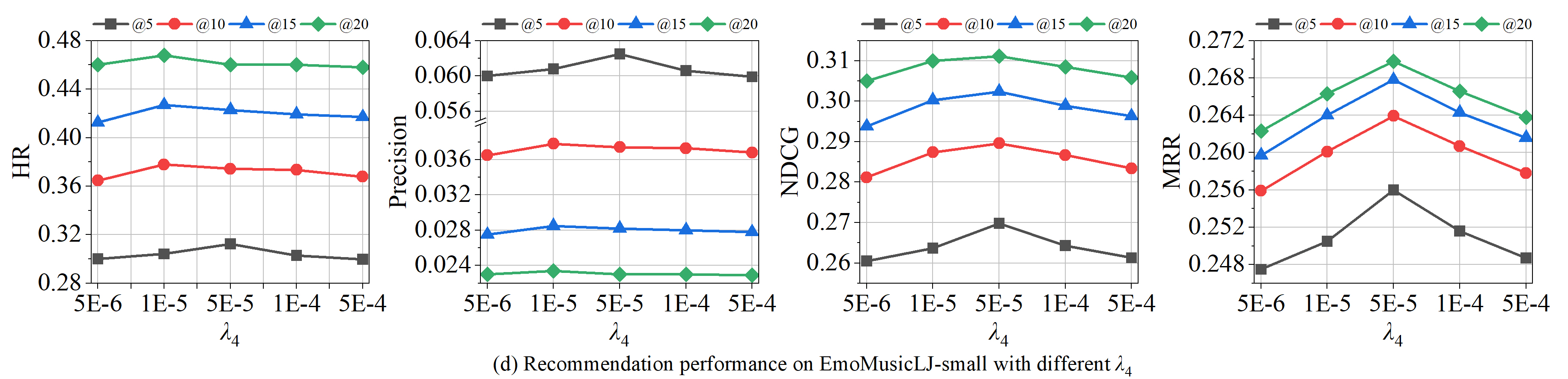}
  \caption{Recommendation performances with different $\lambda_3$ and $\lambda_4$ on both datasets.}
  \Description{}
  \label{lambda_3_4}
\end{figure}

$\lambda_3$ and $\lambda_4$ are the relative weights of $\mathcal{L}_{\mathrm{MSE2}}$ and $\mathcal{L}_{\mathrm{MSE1}}$, respectively. Figure \ref{lambda_3_4} shows the influences of $\lambda_3$ and $\lambda_4$ on recommendation performance. It can be observed that all evaluation metrics initially increase and then decrease with the increase of $\lambda_3$ and $\lambda_4$. We set $\lambda_3$ to 1e-6, and $\lambda_4$ to 1e-4 on EmoMusicLJ dataset, and set $\lambda_3$ to 5e-6, and $\lambda_4$ to 5e-5. Similarly, we can see that $\lambda_4$ is greater than $\lambda_3$. This might be due to $\lambda_4$ controls the constraint on reconstructing user self-reported emotion tags from the posterior LED, while $\lambda_3$ controls the constraint on reconstructing user representations from the personalized prior LED. It is relatively difficult to reconstruct user representation from the prior LED, so the value of $\lambda_3$ is relatively smaller than $\lambda_4$.

\subsection{Interpreting the meaning of LED}
\begin{figure}[h]
  \centering
  \includegraphics[width=0.9\linewidth]{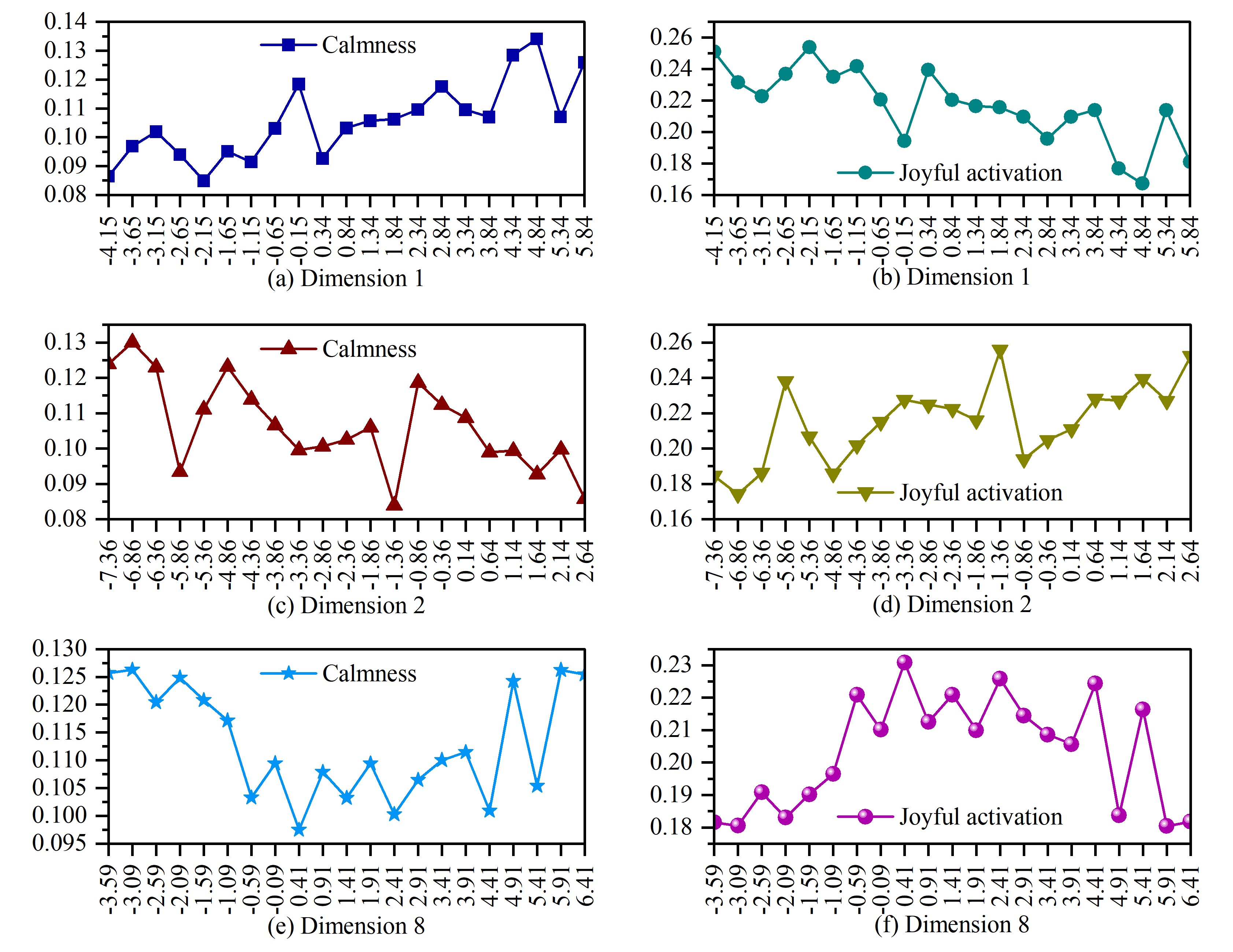}
  \caption{The influence of the dimensional value of the posterior LED on music mood preference prediction.}
  \Description{}
  \label{LED_interprete}
\end{figure}

Following Ma et al. \cite{Ma2019}, we interpret the dimensions of the learned posterior LED by gradually varying the value of a specific dimension and observing its impact on predicting the user’s music mood preference. As noted by Ma et al., not all dimensions are readily interpretable. In this context, we analyze a user from EmoMusicLJ dataset with ID 273,937 under the emotion of \textit{Sad}, and Figure \ref{LED_interprete} provides the influence of representative dimensions.

From Figure \ref{LED_interprete}, we have the following observations:
\begin{enumerate}
    \item [(1)] In Figure \ref{LED_interprete}(a) and Figure \ref{LED_interprete}(b), as the value of the first dimension in posterior LED increases, there is an increase in the predicted preference for \textit{Calmness} while \textit{Joyful Activation} decreases. This suggests that the first dimension likely represents the user’s inclination towards low-activity music moods.
    \item [(2)] In Figure \ref{LED_interprete}(c) and Figure \ref{LED_interprete}(d), the second dimension in the posterior LED appears to represent the user’s preference for high-activity music moods.
    \item [(3)] Notably, the 8th dimension in the posterior LED shows a U-shape impact on the user’s music mood preference (see Figure \ref{LED_interprete}(e) and Figure \ref{LED_interprete}(f)). Initially, as this dimension increases, the user’s preference for low-activity music mood decreases, while the preference for high-activity music mood increases. However, beyond a certain point, this trend reverses.
\end{enumerate}
These observations suggest that our method can learn a posterior LED with a degree of interpretability.

\subsection{Case study}
\begin{figure}[h]
  \centering
  \includegraphics[width=0.9\linewidth]{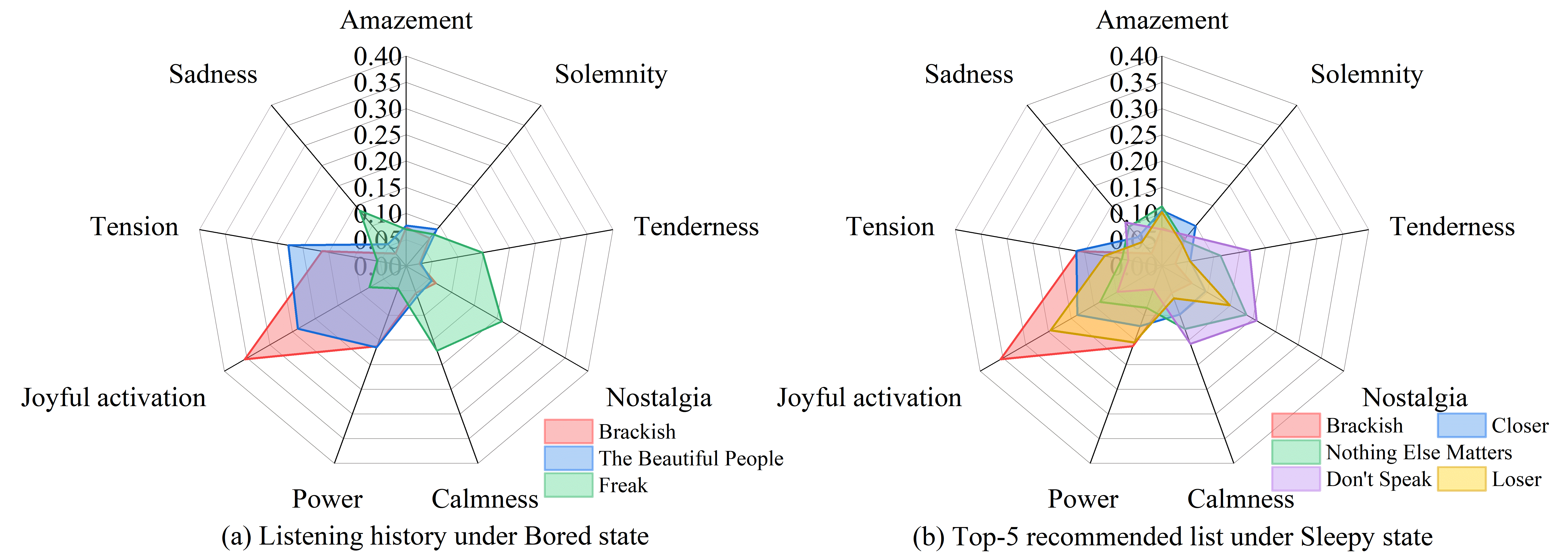}
  \caption{Listening history and top-5 recommendation of user \#274,637.}
  \Description{}
  \label{user274637}
\end{figure}

To assess whether our method can effectively capture user music mood preference heterogeneity within a user, we randomly select a user with ID 274,637 from the test set of EmoMusicLJ and analyze the user’s listening history and the top-5 recommendation list. From Figure \ref{user274637}(a), we can observe that user \#274,637 has listened to three music under the emotion of \textit{Bored}, which is akin to \textit{Sleepy}. Notably, two of these tracks share a similar mood distribution, while the third one exhibits a distinct mood distribution. In Figure \ref{user274637}(b), we present the music mood distributions of the recommended top-5 tracks. It can be observed that the recommended tracks cater to the heterogeneous music mood preferences of user \#274,637 under similar emotions.

\begin{figure}[h]
  \centering
  \includegraphics[width=0.9\linewidth]{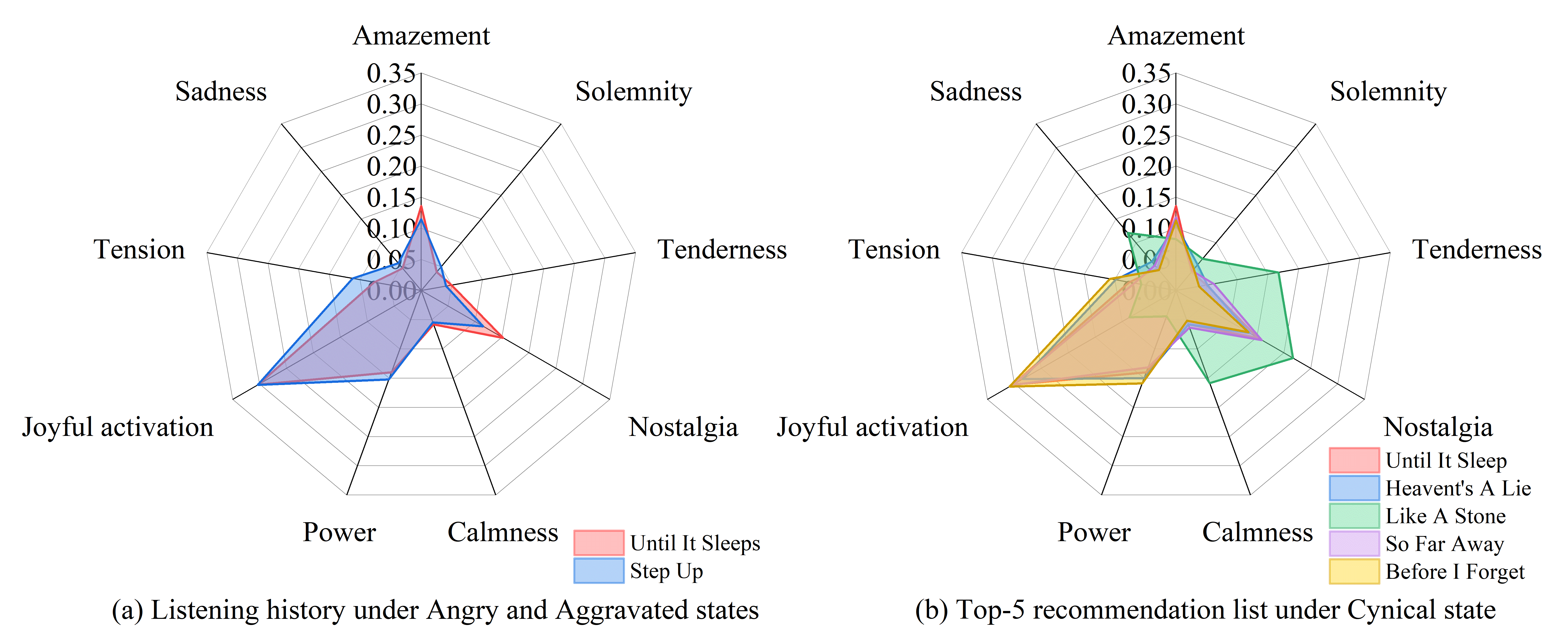}
  \caption{Listening history and top-5 recommendation of user \#269990.}
  \Description{}
  \label{user269990}
\end{figure}

In Figure \ref{user269990}(a), we observe another user with ID 269,990, who has listened to two tracks while experiencing emotions of \textit{Angry} and \textit{Aggravated}, both exhibiting similar mood distributions. Figure \ref{user269990}(b) shows the top-5 recommendation for user \#269,990 in the emotion of \textit{Cynical}, akin to \textit{Angry} and \textit{Aggravated}. It can be observed that our method can learn the user’s stable music mood preference. Meanwhile, our method demonstrates the capability to recommend music that deviates from the user’s historical music mood preference (illustrated by the green track in Figure \ref{user269990}(b)).

\begin{figure}[h]
  \centering
  \includegraphics[width=0.9\linewidth]{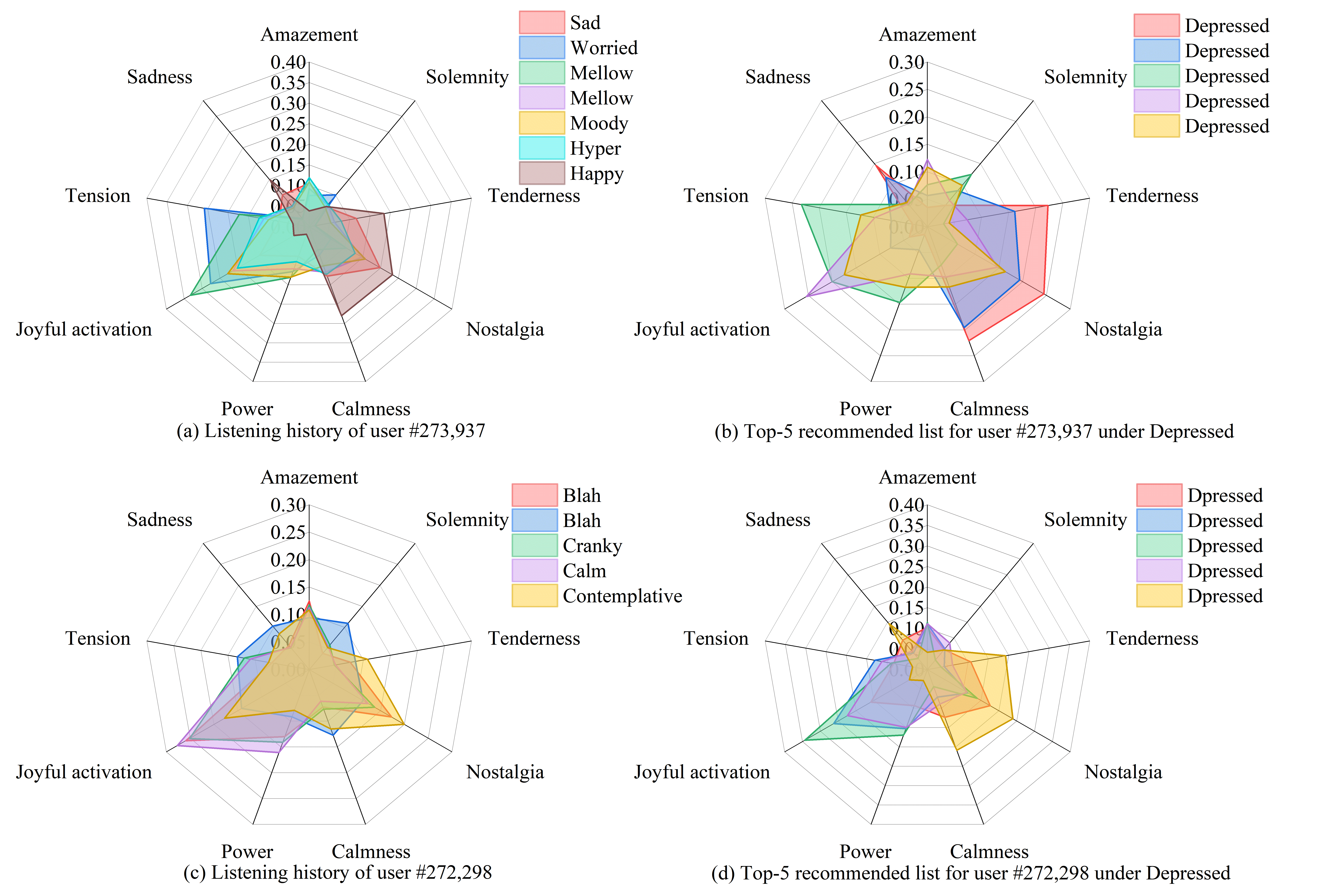}
  \caption{Listening histories and top-5 recommendations of user \#273,937 and user \#272,298.}
  \Description{}
  \label{depressed}
\end{figure}

We also conduct an experiment to evaluate whether our method could capture the music mood preference heterogeneity across users with the same self-reported emotion. We randomly select two users with IDs 273,937 and 272,298, both in the emotion of \textit{Depressed}. Figure \ref{depressed}(a) and Figure \ref{depressed}(c) illustrate their respective listening histories in different emotions. We can see that user \#273,937 shows more variation in the distribution of music mood listened to compared to user \#272,298. Similarly, Figure \ref{depressed}(b) and Figure \ref{depressed}(d) depict the mood distributions of the top-5 recommended tracks for each user. We can observe that the mood distributions for user \#273,937’s top-5 recommendations display more significant fluctuations compared to user \#272,298.

\subsection{User experiment (RQ5)}
To further validate the effectiveness and advantage of our method, we conducted a user experiment to evaluate the recommendation results through human evaluation. Specifically, we randomly drew 50 user listening records from the test set of dataset EmoMusicLJ-small, and generated top-5 recommendations based on our method and the best baseline method (i.e., MF-BPR) for each of the 50 records. We designed a web page to present the two recommendation lists (see Figure \ref{user study} in Appendix A.3 for details) and hired three participants to complete this experiment. Participants were asked to wear headphones in a quiet environment for the experiment. They first listened to a song chosen by the user under a specific emotion, then sequentially listened to songs from the recommended list. After listening to all the songs, participants were asked to rate the likelihood of the user liking each recommended list on a scale from 1 to 7, with higher scores indicating greater preference.
 We determined the winning recommendation list for each test record by majority vote. If at least two participants believed a recommendation list was more likely to be preferred by the user, we recorded that list as the winner.
 
 We propose the null hypothesis ($H_0$): there is no significant difference in the judgment of whether the user in the test case prefers the recommendation list generated by our method or the one generated by the best baseline method. We obtained 45 valid test records, excluding five cases where both lists were considered equally liked by the users. Of the 45 records, our method won 30 votes, while the baseline method won 15. We performed statistical analysis on the experimental data, and the result showed a \textit{p}-value of 0.0375, which is less than 0.05, indicating that we have sufficient evidence to reject the null hypothesis. Therefore, we can conclude that users prefer the recommendations provided by our model, demonstrating the superiority of our method.

\section{CONCLUSION}
In this study, we propose a novel Heterogeneity-aware Deep Bayesian Network (HDBN) for emotion-aware music recommendation. HDBN comprises four components to cope with the problems of emotion heterogeneity across users and within a user and music mood preference heterogeneity across users and within a user. Specifically, we design an inference network for personalized prior LED modeling for emotion heterogeneity across users; we design another inference network for posterior LED modeling for emotion heterogeneity within a user; we group users for music mood preference heterogeneity across users, and design BNN-based music mood preference prediction models for music mood preference heterogeneity within a user. To validate the effectiveness of our model, we constructed two datasets with different levels of sparsity for emotion-aware music recommendation, named EmoMusicLJ and EmoMusicLJ-small, and conducted extensive experiments. Experimental results demonstrate the effectiveness of our method. Case studies show our method’s ability to effectively utilize user emotion and music mood for recommendation, and interpretability analysis shows that our method can learn meaningful latent emotion space for humans to understand. Furthermore, the user study with human evaluation indicates that the recommendations generated by our method are more favored by users.

Our study has several limitations that merit consideration in future research. First, the current method represents the personalized prior LED using relatively simple Gaussian priors. Future studies could delve into more sophisticated priors, drawing inspiration from relevant theories in psychology. Second, due to data limitations, our study relies on learning the posterior LED from user self-report emotion tags. In future studies, more nuanced user emotion modeling could be explored if richer user behavior data, such as post text content, becomes accessible. Third, limited by the datasets, this study does not account for the influence of music on user emotion. Future research can explore how users’ emotions change after listening to music and how they further select music based on these changes.


\bibliographystyle{ACM-Reference-Format}
\bibliography{sample-base}

\appendix

\section{APPENDICES}

\subsection{Equation Derivation}

\subsubsection{The Derivation of $\log⁡ p(\{m_{u,v}\},\{\boldsymbol{s}_{u,v}\})$}\label{appA11}
\begin{equation}
    \begin{split}
        \log p(\{m_{u,v}\},\{\boldsymbol{s}_{u,v}\})&=\log p(\{m_{u,v}\},\{\boldsymbol{s}_{u,v}\})\int({q\{\boldsymbol{\psi}_g\},\boldsymbol{\psi},\{\boldsymbol{\mu}_{u,v}\},\{\boldsymbol{\mu}_u\},\{\boldsymbol{l}_{u,v}\})}d(\{\boldsymbol{\psi}_g\},\boldsymbol{\psi},\{\boldsymbol{\mu}_{u,v}\},\{\boldsymbol{\mu}_u\},\{\boldsymbol{l}_{u,v}\})\\
        &=\int{{q(\{\boldsymbol{\psi}_g\},\boldsymbol{\psi},\{\boldsymbol{\mu}_{u,v}\},\{\boldsymbol{\mu}_u\},\{\boldsymbol{l}_{u,v}\})}\log p(\{m_{u,v}\},\{\boldsymbol{s}_{u,v}\})}d\{\boldsymbol{\psi}_g\},\boldsymbol{\psi},\{\boldsymbol{\mu}_{u,v}\},\{\boldsymbol{\mu}_u\},\{\boldsymbol{l}_{u,v}\})\\
        &=\mathbb{E}_q\left[\log p(\{m_{u,v}\},\{\boldsymbol{s}_{u,v}\})\right]\\
        &=\mathbb{E}_q\left[\log \left(\frac{p\left(\{m_{u,v}\},\{\boldsymbol{s}_{u,v}\},\{\boldsymbol{\psi}_g\},\boldsymbol{\psi},\{\boldsymbol{\mu}_{u,v}\},\{\boldsymbol{\mu}_u\},\{\boldsymbol{l}_{u,v}\}\right)}{p\left(\{\boldsymbol{\psi}_g\},\boldsymbol{\psi},\{\boldsymbol{\mu}_{u,v}\},\{\boldsymbol{\mu}_u\},\{\boldsymbol{l}_{u,v}\}\mid \{m_{u,v}\},\{\boldsymbol{s}_{u,v}\}\right)}\right)\right]\\
        &=\mathbb{E}_q\left[\log \left(\frac{p\left(\{m_{u,v}\},\{\boldsymbol{s}_{u,v}\},\{\boldsymbol{\psi}_g\},\boldsymbol{\psi},\{\boldsymbol{\mu}_{u,v}\},\{\boldsymbol{\mu}_u\},\{\boldsymbol{l}_{u,v}\}\right)\cdot q\left(\{\boldsymbol{\psi}_g\},\boldsymbol{\psi},\{\boldsymbol{\mu}_{u,v}\},\{\boldsymbol{\mu}_u\},\{\boldsymbol{l}_{u,v}\}\right)}{p\left(\{\boldsymbol{\psi}_g\},\boldsymbol{\psi},\{\boldsymbol{\mu}_{u,v}\},\{\boldsymbol{\mu}_u\},\{\boldsymbol{l}_{u,v}\}\mid \{m_{u,v}\},\{\boldsymbol{s}_{u,v}\}\right)\cdot q\left(\{\boldsymbol{\psi}_g\},\boldsymbol{\psi},\{\boldsymbol{\mu}_{u,v}\},\{\boldsymbol{\mu}_u\},\{\boldsymbol{l}_{u,v}\}\right)}\right)\right]\\
        &=\mathbb{E}_q\left[p\left(\{m_{u,v}\},\{\boldsymbol{s}_{u,v}\},\{\boldsymbol{\psi}_g\},\boldsymbol{\psi},\{\boldsymbol{\mu}_{u,v}\},\{\boldsymbol{\mu}_u\},\{\boldsymbol{l}_{u,v}\}\right)-\log q\left(\{\boldsymbol{\psi}_g\},\boldsymbol{\psi},\{\boldsymbol{\mu}_{u,v}\},\{\boldsymbol{\mu}_u\},\{\boldsymbol{l}_{u,v}\}\right)\right]\\
        &\quad+\mathrm{KL}\left(\log q\left(\{\boldsymbol{\psi}_g\},\boldsymbol{\psi},\{\boldsymbol{\mu}_{u,v}\},\{\boldsymbol{\mu}_u\},\{\boldsymbol{l}_{u,v}\}\right)\parallel p\left(\{\boldsymbol{\psi}_g\},\boldsymbol{\psi},\{\boldsymbol{\mu}_{u,v}\},\{\boldsymbol{\mu}_u\},\{\boldsymbol{l}_{u,v}\}\mid \{m_{u,v}\},\{\boldsymbol{s}_{u,v}\}\right)\right)\nonumber
    \end{split}
\end{equation}

\subsubsection{The Derivation of $\mathrm{ELBO}(q)$}\label{appA12}
\begin{equation}
    \begin{split}
        \mathrm{ELBO}_q&=\mathbb{E}_q\left[\log p\left(\{m_{u,v}\},\{\boldsymbol{s}_{u,v}\},\{\boldsymbol{\psi}_g\},\boldsymbol{\psi},\{\boldsymbol{\mu}_{u,v}\},\{\boldsymbol{\mu}_u\},\{\boldsymbol{l}_{u,v}\}\right)-\log q\left(\{\boldsymbol{\psi}_g\},\boldsymbol{\psi},\{\boldsymbol{\mu}_{u,v}\},\{\boldsymbol{\mu}_u\},\{\boldsymbol{l}_{u,v}\}\right)\right]\\
        &=\mathbb{E}_q\left[\log p\left(\{m_{u,v}\},\{\boldsymbol{s}_{u,v}\},\{\boldsymbol{\psi}_g\},\boldsymbol{\psi},\{\boldsymbol{\mu}_{u,v}\},\{\boldsymbol{\mu}_u\},\{\boldsymbol{l}_{u,v}\}\right)\right]-\mathbb{E}_q\left[\log q\left(\{\boldsymbol{\psi}_g\},\boldsymbol{\psi},\{\boldsymbol{\mu}_{u,v}\},\{\boldsymbol{\mu}_u\},\{\boldsymbol{l}_{u,v}\}\right)\right]\\
        &=\mathbb{E}_q\left[\log p\left(\{\boldsymbol{\mu}_u\},\{\boldsymbol{l}_{u,v}\}\mid \{m_{u,v}\},\{\boldsymbol{s}_{u,v}\},\{\boldsymbol{\psi}_g\},\boldsymbol{\psi},\{\boldsymbol{\mu}_{u,v}\}\right)\cdot p\left(\{m_{u,v}\},\{\boldsymbol{s}_{u,v}\},\{\boldsymbol{\psi}_g\},\boldsymbol{\psi},\{\boldsymbol{\mu}_{u,v}\}\right)\right]\\
        &\quad-\mathbb{E}_q\left[\log q\left(\{\boldsymbol{\psi}_g\},\boldsymbol{\psi},\{\boldsymbol{\mu}_{u,v}\},\{\boldsymbol{\mu}_u\},\{\boldsymbol{l}_{u,v}\}\right)\right]\\
        &=\mathbb{E}_q\left[\log p\left(\{\boldsymbol{\mu}_u\},\{\boldsymbol{l}_{u,v}\}\mid \{m_{u,v}\},\{\boldsymbol{s}_{u,v}\},\{\boldsymbol{\psi}_g\},\boldsymbol{\psi},\{\boldsymbol{\mu}_{u,v}\}\right)\right]+\mathbb{E}_q\left[\log p\left(\{m_{u,v}\},\{\boldsymbol{s}_{u,v}\},\{\boldsymbol{\psi}_g\},\boldsymbol{\psi},\{\boldsymbol{\mu}_{u,v}\}\right)\right]\\
        &\quad-\mathbb{E}_q\left[\log q\left(\{\boldsymbol{\psi}_g\},\boldsymbol{\psi},\{\boldsymbol{\mu}_{u,v}\},\{\boldsymbol{\mu}_u\},\{\boldsymbol{l}_{u,v}\}\right)\right]\\
        &=\mathbb{E}_q\left[\log p\left(\{\boldsymbol{\mu}_u\},\{\boldsymbol{l}_{u,v}\}\mid \{m_{u,v}\},\{\boldsymbol{s}_{u,v}\},\{\boldsymbol{\psi}_g\},\boldsymbol{\psi},\{\boldsymbol{\mu}_{u,v}\}\right)\right]\\
        &\quad-\left(\mathbb{E}_q\left[\log q\left(\{\boldsymbol{\psi}_g\},\boldsymbol{\psi},\{\boldsymbol{\mu}_{u,v}\},\{\boldsymbol{\mu}_u\},\{\boldsymbol{l}_{u,v}\}\right)\right]-\mathbb{E}_q\left[\log p\left(\{m_{u,v}\},\{\boldsymbol{s}_{u,v}\},\{\boldsymbol{\psi}_g\},\boldsymbol{\psi},\{\boldsymbol{\mu}_{u,v}\}\right)\right]\right)\\
        &=\mathbb{E}_q\left[\log p\left(\{\boldsymbol{\mu}_u\},\{\boldsymbol{l}_{u,v}\}\mid \{m_{u,v}\},\{\boldsymbol{s}_{u,v}\},\{\boldsymbol{\psi}_g\},\boldsymbol{\psi},\{\boldsymbol{\mu}_{u,v}\}\right)\right]\\
        &\quad-\mathrm{KL}\left(q\left(\{\boldsymbol{\psi}_g\},\boldsymbol{\psi},\{\boldsymbol{\mu}_{u,v}\},\{\boldsymbol{\mu}_u\},\{\boldsymbol{l}_{u,v}\}\right)\parallel p\left(\{m_{u,v}\},\{\boldsymbol{s}_{u,v}\},\{\boldsymbol{\psi}_g\},\boldsymbol{\psi},\{\boldsymbol{\mu}_{u,v}\}\right)\right)\nonumber
    \end{split}
\end{equation}

\subsubsection{The derivation of $\mathrm{KL}\left(q(\{\boldsymbol{\mu}_{u,v}\},\{\boldsymbol{\mu}_u\})\parallel p(\{\boldsymbol{\mu}_{u,v}\},\{\boldsymbol{\mu}_u\})\right)$}\label{appA13}
\begin{equation}
    \begin{split}
        \mathrm{KL}\left(q(\{\boldsymbol{\mu}_{u,v}\},\{\boldsymbol{\mu}_u\})\parallel p(\{\boldsymbol{\mu}_{u,v}\},\{\boldsymbol{\mu}_u\})\right)&=\int{q(\{\boldsymbol{\mu}_{u,v}\},\{\boldsymbol{\mu}_u\})\log \frac{q(\{\boldsymbol{\mu}_{u,v}\},\{\boldsymbol{\mu}_u\})}{p(\{\boldsymbol{\mu}_{u,v}\},\{\boldsymbol{\mu}_u\})}d(\{\boldsymbol{\mu}_{u,v}\},\{\boldsymbol{\mu}_u\})}\\
        &=\int{q(\{\boldsymbol{\mu}_{u,v}\}\mid \{\boldsymbol{\mu}_u\})q(\{\boldsymbol{\mu}_u\})\log \frac{q(\{\boldsymbol{\mu}_{u,v}\}\mid\{\boldsymbol{\mu}_u\})q(\{\boldsymbol{\mu}_u\})}{p(\{\boldsymbol{\mu}_{u,v}\}\mid\{\boldsymbol{\mu}_u\})p(\{\boldsymbol{\mu}_u\})}d(\{\boldsymbol{\mu}_{u,v}\},\{\boldsymbol{\mu}_u\})}\\
        &=\int{q(\{\boldsymbol{\mu}_{u,v}\}\mid \{\boldsymbol{\mu}_u\})q(\{\boldsymbol{\mu}_u\})\left(\log \frac{q(\{\boldsymbol{\mu}_{u,v}\}\mid\{\boldsymbol{\mu}_u\})}{p(\{\boldsymbol{\mu}_{u,v}\}\mid\{\boldsymbol{\mu}_u\})}+\log \frac{q(\{\boldsymbol{\mu}_u\})}{p(\{\boldsymbol{\mu}_u\})}\right)d(\{\boldsymbol{\mu}_{u,v}\},\{\boldsymbol{\mu}_u\})}\\
        &=\int{q(\{\boldsymbol{\mu}_{u,v}\}\mid \{\boldsymbol{\mu}_u\})\left(\int{q(\{\boldsymbol{\mu}_u\})\log \frac{q(\{\boldsymbol{\mu}_u\})}{p(\{\boldsymbol{\mu}_u\})}d\{\boldsymbol{\mu}_u\}}\right)d\{\boldsymbol{\mu}_{u,v}\}}\\
        &\quad+\int{q(\{\boldsymbol{\mu}_u\})\left(\int{q(\{\boldsymbol{\mu}_{u,v}\}\mid \{\boldsymbol{\mu}_u\})\log \frac{q(\{\boldsymbol{\mu}_{u,v}\}\mid\{\boldsymbol{\mu}_u\})}{p(\{\boldsymbol{\mu}_{u,v}\}\mid\{\boldsymbol{\mu}_u\})}d\{\boldsymbol{\mu}_{u,v}\}}\right)d\{\boldsymbol{\mu}_u\}}\\
        &=\mathrm{KL}\left(q(\{\boldsymbol{\mu}_u\})\parallel p(\{\boldsymbol{\mu}_u\})\right)+\mathrm{KL}\left(q(\{\boldsymbol{\mu}_{u,v}\}\mid \{\boldsymbol{\mu}_u\})\parallel p(\{\boldsymbol{\mu}_{u,v}\}\mid \{\boldsymbol{\mu}_u\})\right)\nonumber
    \end{split}
\end{equation}

Since $\boldsymbol{\mu}_{u,v}$ is generated from $\boldsymbol{\mu}_u$, the posterior distribution of $\boldsymbol{\mu}_u$ is used as the prior of $\boldsymbol{\mu}_{u,v}$. Above Equation can be written as,
\begin{equation}
    \mathrm{KL}\left(q(\{\boldsymbol{\mu}_{u,v}\},\{\boldsymbol{\mu}_u\})\parallel p(\{\boldsymbol{\mu}_{u,v}\},\{\boldsymbol{\mu}_u\})\right)=\mathrm{KL}\left(q(\{\boldsymbol{\mu}_u\})\parallel p(\{\boldsymbol{\mu}_u\})\right)+\mathrm{KL}\left(q(\{\boldsymbol{\mu}_{u,v}\}\mid \{\boldsymbol{\mu}_u\})\parallel q(\{\boldsymbol{\mu}_u\})\right)\nonumber
\end{equation}

Since $q(\boldsymbol{\mu}_u)=\mathcal{N}(\boldsymbol{\mu}_u,\boldsymbol{\sigma}_1)$, $p(\boldsymbol{\mu}_u)=\mathcal{N}(\boldsymbol{0},\boldsymbol{1})$, and $q(\{\boldsymbol{\mu}_{u,v}\}\mid \{\boldsymbol{\mu}_u\})=\mathcal{N}(\boldsymbol{\mu}_{u,v},\boldsymbol{\sigma}_2)$, the first and the second term in above Equation can be calculated as,
\begin{equation}
    \begin{split}
        \mathrm{KL}\left(q(\{\boldsymbol{\mu}\})\parallel p(\{\boldsymbol{\mu}_u\})\right)&=\sum_{\boldsymbol{\mu}_u \in \{\boldsymbol{\mu}_u\}}\mathrm{KL}\left(q(\boldsymbol{\mu}_u)\parallel p(\boldsymbol{\mu}_u)\right)\\
        &=\frac{1}{2}\sum_{\boldsymbol{\mu}_u\in \{\boldsymbol{\mu}_u\}}\sum_{j=1}^J\left(\mu_{u,j}^2+\sigma_{1,j}^2-\log (\sigma_{1,j}^2)-1\right)\nonumber
    \end{split}
\end{equation}

\begin{equation}
    \begin{split}
        \mathrm{KL}\left(q(\{\boldsymbol{\mu}_{u,v}\}\mid \{\boldsymbol{\mu}_u\})\parallel q(\{\boldsymbol{\mu}_u\})\right)&=\sum_{\boldsymbol{\mu}_{u,v}\in \{\boldsymbol{\mu}_{u,v}\}}\mathrm{KL}\left(q(\boldsymbol{\mu}_{u,v}\mid \boldsymbol{\mu}_u)\parallel q(\boldsymbol{\mu}_u)\right)\\
        &=\frac{1}{2}\sum_{\boldsymbol{\mu}_{u,v}\in \{\boldsymbol{\mu}_{u,v}\}}\sum_{j=1}^J\left(\log \frac{\sigma_{1,j}^2}{\sigma_{2,j}^2}+\frac{\sigma_{2,j}^2}{\sigma_{2,j}^2}+\frac{(\mu_{u,v,j}-\mu_{u,v})^2}{\sigma_{1,j}^2}-1\right)\nonumber
    \end{split}
\end{equation}

where $J$ is the dimensional size of $\boldsymbol{\mu}_u$.

Thus, $\mathrm{KL}\left(q(\{\boldsymbol{\mu}_{u,v}\},\{\boldsymbol{\mu}_u\})\parallel p(\{\boldsymbol{\mu}_{u,v}\},\{\boldsymbol{\mu}_u\})\right)$ can be computed as,
\begin{equation}
    \begin{split}
        \mathrm{KL}\left(q(\{\boldsymbol{\mu}_{u,v}\},\{\boldsymbol{\mu}_u\})\parallel p(\{\boldsymbol{\mu}_{u,v}\},\{\boldsymbol{\mu}_u\})\right)&=\frac{1}{2}\sum_{\boldsymbol{\mu}_u\in \{\boldsymbol{\mu}_u\}}\sum_{j=1}^J\left(\mu_{u,j}^2+\sigma_{1,j}^2-\log (\sigma_{1,j}^2)-1\right)\\
        &\quad+\frac{1}{2}\sum_{\boldsymbol{\mu}_{u,v}\in \{\boldsymbol{\mu}_{u,v}\}}\sum_{j=1}^J\left(\log \frac{\sigma_{1,j}^2}{\sigma_{2,j}^2}+\frac{\sigma_{2,j}^2}{\sigma_{2,j}^2}+\frac{(\mu_{u,v,j}-\mu_{u,v})^2}{\sigma_{1,j}^2}-1\right)\nonumber
    \end{split}
\end{equation}

\subsection{Details of UCFE and ICFE}
In this part, we provide the detailed calculation process of UCFE and ICFE.
\subsubsection{UCFE}\label{appA21}\quad\\
For two users $u_1$ and $u_2$, the similarity is computed as,

\begin{equation}
    sim(u_1,u_2)=\frac{\sum_{v\in \mathbb{V}_{u_1}\bigcap\mathbb{V}_{u_2}}\mathrm{cos}(\boldsymbol{e}_{u_1,v},\boldsymbol{e}_{u_2,v})}{\sqrt{\mid\mathbb{V}_{u_1}\mid\times \mid\mathbb{V}_{u_2}\mid}}\nonumber
\end{equation}

where $\mathbb{V}_{u_1}$ and $\mathbb{V}_{u_2}$ are music set listened by $u_1$ and $u_2$ respectively. Then the first $m$ users $\mathbb{U}_{u,m}$ for the target user $u$ can be obtained based on the user similarity. The target user $u$’s interest in music $v$ can be computed as,
\begin{equation}
    r_{u,v}=\sum_{u^{'}\in \mathbb{U}_{u,m}\bigcap\mathbb{U}_v}sim(u,u^{'})\times \mathrm{cos}(\boldsymbol{e}_{u_1,v},\boldsymbol{e}_{u_2,v})\nonumber
\end{equation}
where $\mathrm{U}_v$ is the set of users who listened to music $v$. $\boldsymbol{e}_u$ is the current emotion of $u$, and $\boldsymbol{e}_{u^{'},v}$ is the emotion of user$ u^{'}$ when listening to music $v$.

\subsubsection{ICFE}\label{appA22}\quad\\
For two music $v_1$ and $v_2$, the similarity is computed as,
\begin{equation}
    sim(v_1,v_2)=\frac{\sum_{u\in \mathbb{U}_{v_1}\bigcap\mathbb{U}_{v_2}}\mathrm{cos}(\boldsymbol{e}_{u,v_1},\boldsymbol{e}_{u,v_2})}{\sqrt{\mid\mathbb{U}_{v_1}\mid\times \mid\mathbb{U}_{v_2}\mid}}\nonumber
\end{equation}
where $\mathbb{U}_{v_1}$ is the set of users who listened to music $v_1$. $\mathbb{U}_{v_2}$ is the set of users who listened to music $v_2$. $\boldsymbol{e}_{u,v_1}$ and $\boldsymbol{e}_{u,v_2})$ are emotion of user $u$ when listening to music $v_1$ and $v_2$, respectively. Then the first $m$ music pieces $\mathbb{V}_{v,m}$ for the target music $v$ can be obtained based on the similarity. The target user $u$’s interest in music $v$ can be computed as,
\begin{equation}
     r_{u,v}=\sum_{v^{'}\in \mathbb{V}_{v,m}\bigcap\mathbb{V}_u}sim(v,v^{'})\times \mathrm{cos}(\boldsymbol{e}_{u},\boldsymbol{e}_{u,v^{'}})\nonumber
\end{equation}
where $\mathbb{V}_u$ is the set of music that had been listened by $u$. $\boldsymbol{e}_{u,v^{'}}$ is the emotion of user $u$ when listening to music $v^{'}$.

\subsection{The Web Interface for User Experiment}
Here we provided a sample interface of the web page used in the user study in Section 4.10.
\begin{figure}[h]
  \centering
  \includegraphics[width=\linewidth]{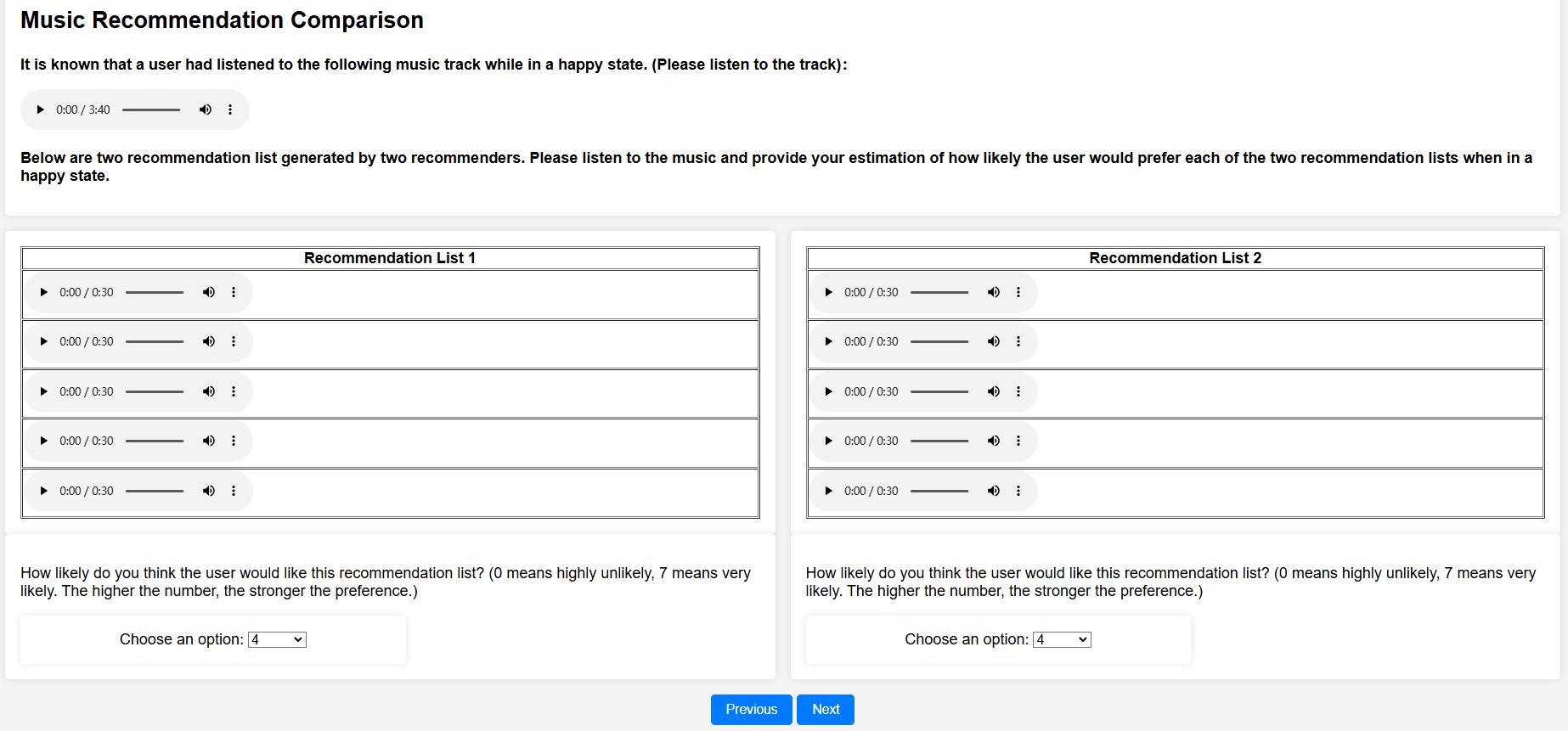}
  \caption{Example of the interface of user experiment web page.}
  \Description{}
  \label{user study}
\end{figure}

\end{document}